\documentclass[11pt]{article}

\usepackage[final]{acl}

\usepackage{times}
\usepackage{latexsym}

\usepackage[T1]{fontenc}
\usepackage[utf8]{inputenc}

\usepackage{microtype}

\usepackage{inconsolata}

\usepackage{graphicx}

\usepackage{microtype}
\usepackage{graphicx}
\usepackage{subfigure}
\usepackage{booktabs} % for professional tables

\usepackage{hyperref}

\usepackage{algorithm}  % Add this line
\usepackage{algorithmic}  % Add this line too for algorithm contents

\usepackage{amsmath}
\usepackage{amssymb}
\usepackage{mathtools}
\usepackage{amsthm}
\usepackage{multirow} 
\usepackage{pifont}
\usepackage{amssymb}
\usepackage{xspace}
\usepackage{siunitx}
\usepackage{colortbl}  % Add this before using \columncolor
\usepackage{tcolorbox}
\usepackage{titletoc} % table of contents
\usepackage{svg}
\usepackage{twemojis}
\usepackage{multicol}

\newcommand{\std}[1]{{\scriptsize\textcolor{gray}{~\raisebox{0.5pt}{$\pm$}#1}}}

\definecolor{ForestGreen}{RGB}{34,139,34}
\newcommand{\COMMENTT}[1]{\textcolor{ForestGreen}{\# \text{#1}}}

\newcommand{\command}[1]{\texttt{\textcolor{blue}{#1}}}
\newcommand{\red}[1]{\textcolor{red}{#1}}

\definecolor{darkgreen}{rgb}{0.0,0.5,0.0}
\newcommand{\cmark}{\textcolor{darkgreen}{\ding{51}}}

\newcommand{\vision}{\raisebox{-0.5ex}{\includegraphics[height=1em]{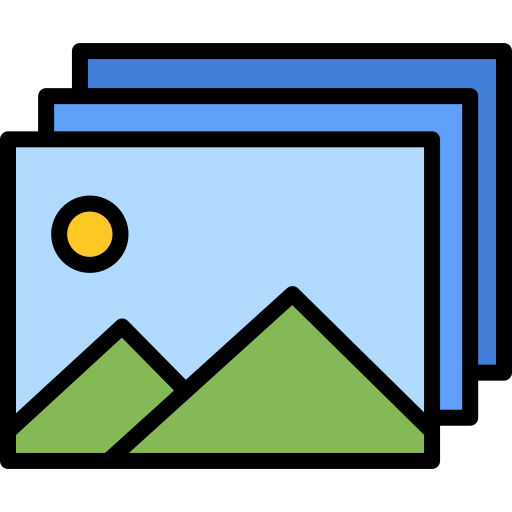}}}
\newcommand{\calculator}{\raisebox{-0.5ex}{\includegraphics[height=1em]{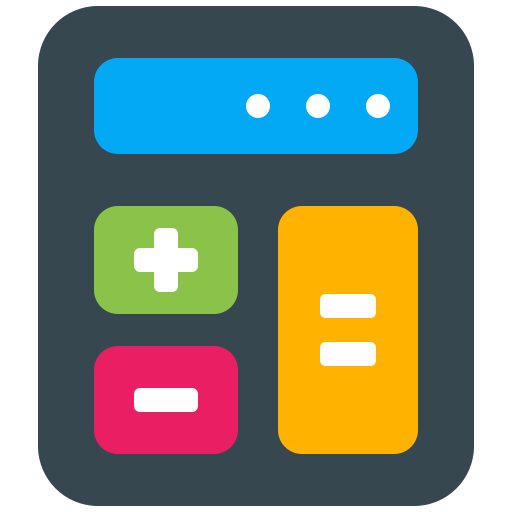}}}
\newcommand{\knowledge}{\raisebox{-0.5ex}{\includegraphics[height=1em]{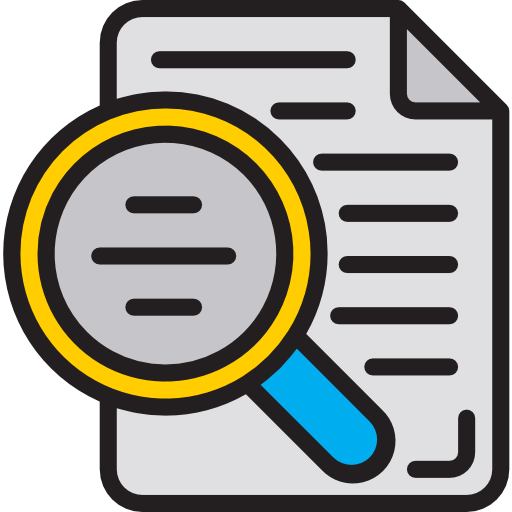}}}
\newcommand{\steps}{\raisebox{-0.5ex}{\includegraphics[height=1em]{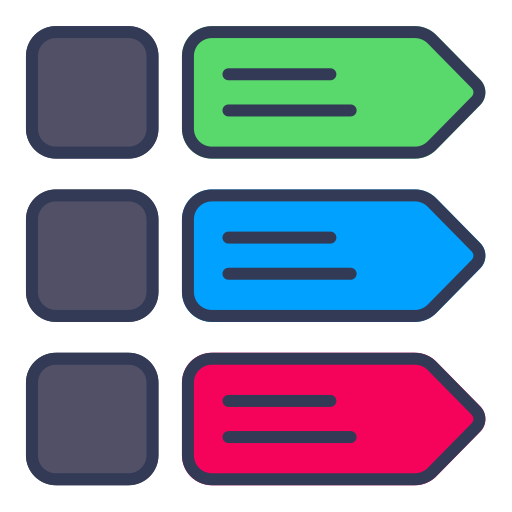}}}

\newcommand{\base}{\raisebox{-0.5ex}{\includegraphics[height=1em]{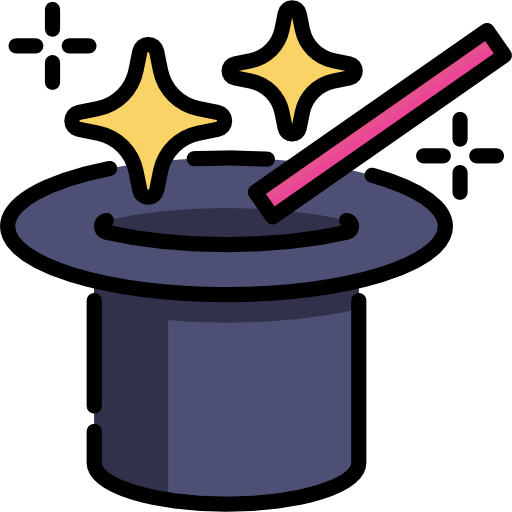}}}
\newcommand{\www}{\raisebox{-0.5ex}{\includegraphics[height=1em]{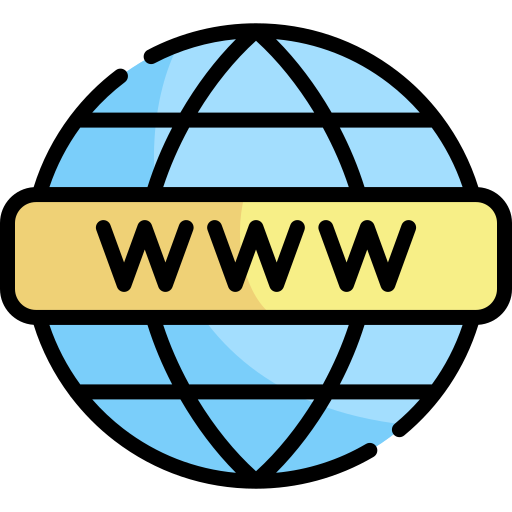}}}
\newcommand{\med}{\raisebox{-0.5ex}{\includegraphics[height=1em]{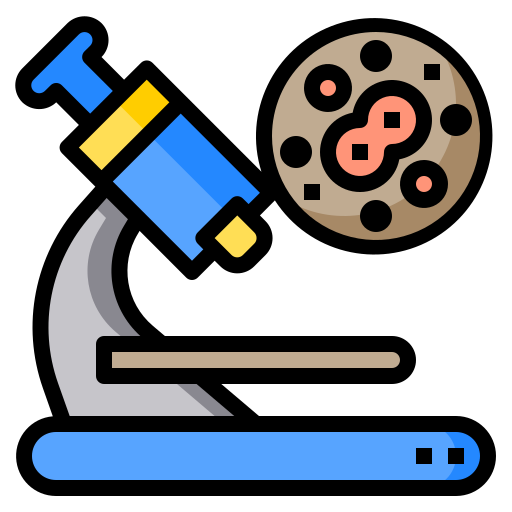}}}
\newcommand{\code}{\raisebox{-0.5ex}{\includegraphics[height=1em]{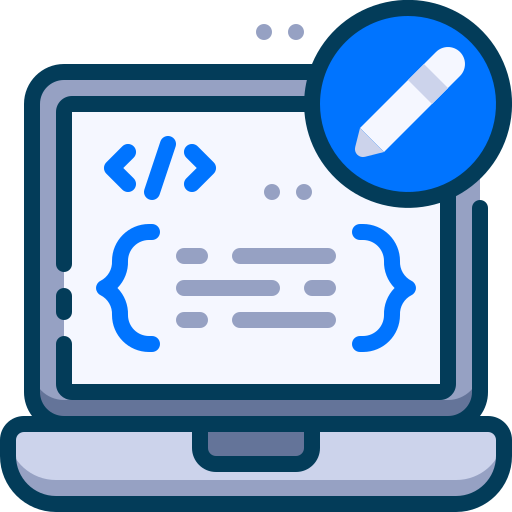}}}
\newcommand{\image}{\raisebox{-0.5ex}{\includegraphics[height=1em]{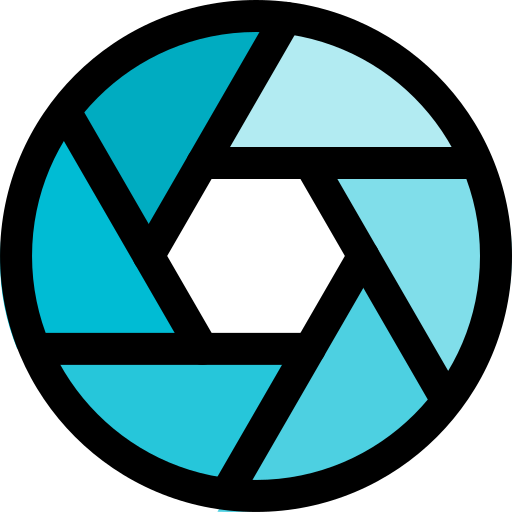}}}

\newcommand{\six}{\raisebox{-0.5ex}{\includegraphics[height=1em]{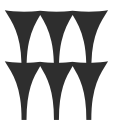}}\xspace}
\newcommand{\eight}{\raisebox{-0.5ex}{\includegraphics[height=1em]{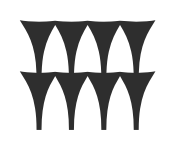}}\xspace}
\newcommand{\fifty}{\raisebox{-0.5ex}{\includegraphics[height=1em]{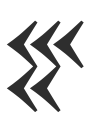}}\xspace}
\newcommand{\fiftysix}{\raisebox{-0.5ex}{\includegraphics[height=1em]{examples/mm_50.png}\includegraphics[height=1em]{examples/mm_6.png}}}

\newcommand{\model}{OctoTools\xspace}
\newcommand{\modelbase}{$\text{OctoTools}_\text{base}$\xspace}

\newcommand{\gpt}{GPT-4o\xspace}
\newcommand{\gptmini}{GPT-4o-mini\xspace}

\newcommand{\autogen}{AutoGen\xspace}
\newcommand{\langchain}{LangChain\xspace}
\newcommand{\gptplugin}{GPT-Functions\xspace}

\newcommand{\gptengine}{\texttt{gpt-4o-2024-08-06}\xspace}

\usepackage[utf8]{inputenc}
\usepackage{CJKutf8}

\definecolor{toolcardbox}{RGB}{240, 248, 255} % Light blue background
\definecolor{toolcardborder}{RGB}{52, 52, 173} % Purple border

\newtcolorbox{textbox}[1][]{
    colback=white,
    colframe=toolcardborder,
    arc=1pt,                % small corner rounding
    boxrule=1pt,          % thin rule
    title=#1,               % uses optional argument as title
    fonttitle=\bfseries,
    left=5pt,               % additional space on left
    right=5pt,              % additional space on right
    top=5pt,                % additional space at top
    bottom=5pt,             % additional space at bottom
    before skip=1em,        % vertical skip before the box
    after skip=1em          % vertical skip after the box
}

\definecolor{querybg}{RGB}{245, 245, 245}      % Light gray background
\definecolor{queryframe}{RGB}{64, 64, 64}      % Dark gray frame
\definecolor{querybadge}{RGB}{169, 169, 169}   % Medium gray badge

\newtcolorbox{querybox}[1][]{
    float=!ht,  % Add this line to force "here" placement
    colback=querybg,
    colframe=queryframe,
    arc=4pt,                % small corner rounding
    title=#1,               % uses optional argument as title
    fonttitle=\bfseries,
    left=5pt,               % additional space on left
    right=5pt,              % additional space on right
    top=5pt,                % additional space at top
    bottom=5pt,             % additional space at bottom
    before skip=1em,        % vertical skip before the box
    after skip=1em,          % vertical skip after the box
    colbacktitle=querybadge,  % background of the title (the 'badge')
    coltitle=black              % title text color, if desired
}

\definecolor{plannerbg}{RGB}{248, 230, 234}      % Light red-gray background
\definecolor{plannerframe}{RGB}{176, 36, 24}    % Steel red frame
\definecolor{plannerbadge}{RGB}{225, 151, 168}      % Dark red gray badge

\newtcolorbox{plannerbox}[1][]{
    colback=plannerbg,
    colframe=plannerframe,
    arc=4pt,                % small corner rounding
    title=#1,               % uses optional argument as title
    fonttitle=\bfseries,
    left=5pt,               % additional space on left
    right=5pt,              % additional space on right
    top=5pt,                % additional space at top
    bottom=5pt,             % additional space at bottom
    before skip=1em,        % vertical skip before the box
    after skip=1em,          % vertical skip after the box
    colbacktitle=plannerbadge,  % background of the title (the 'badge')
    coltitle=black              % title text color, if desired
}

\definecolor{executorbg}{RGB}{240, 248, 255} % Light blue background
\definecolor{executorframe}{RGB}{37, 82, 144}    % Steel blue frame
\definecolor{executorbadge}{RGB}{167, 196, 230}      % Dark blue gray badge

\newtcolorbox{executorbox}[1][]{
    colback=executorbg,
    colframe=executorframe,
    arc=4pt,                % small corner rounding
    title=#1,               % uses optional argument as title
    fonttitle=\bfseries,
    left=5pt,               % additional space on left
    right=5pt,              % additional space on right
    top=5pt,                % additional space at top
    bottom=5pt,             % additional space at bottom
    before skip=1em,        % vertical skip before the box
    after skip=1em,          % vertical skip after the box
    colbacktitle=executorbadge,  % background of the title (the 'badge')
    coltitle=black              % title text color, if desired
}

\definecolor{answerbg}{RGB}{239, 255, 229}      % Light green-tinted background
\definecolor{answerframe}{RGB}{34, 139, 34}     % Forest green frame
\definecolor{answerbadge}{RGB}{182, 200, 108}   % Light green badge

\newtcolorbox{answerbox}[1][]{
    colback=answerbg,
    colframe=answerframe,
    arc=4pt,                % small corner rounding
    title=#1,               % uses optional argument as title
    fonttitle=\bfseries,
    left=5pt,               % additional space on left
    right=5pt,              % additional space on right
    top=5pt,                % additional space at top
    bottom=5pt,             % additional space at bottom
    before skip=1em,        % vertical skip before the box
    after skip=1em,          % vertical skip after the box
    colbacktitle=answerbadge,  % background of the title (the 'badge')
    coltitle=black              % title text color, if desired
}

\newtcolorbox{textcolorbox}[1][]{
    float=!ht,  % Add this line to force "here" placement
    colback=toolcardbox,
    colframe=toolcardborder,
    arc=1pt,                % small corner rounding
    boxrule=1pt,          % thin rule
    title=#1,               % uses optional argument as title
    fonttitle=\bfseries,
    left=5pt,               % additional space on left
    right=5pt,              % additional space on right
    top=5pt,                % additional space at top
    bottom=5pt,             % additional space at bottom
    before skip=1em,        % vertical skip before the box
    after skip=1em          % vertical skip after the box
}

\tcbuselibrary{listings}
\usepackage{listings}

\newtcblisting{codecolorbox}[2][]{%
    float=!ht,  % Add this line to force "here" placement
    colback=toolcardbox,
    colframe=toolcardborder,
    arc=1pt,         % small corner rounding
    boxrule=1pt,     % thin rule
    title=#1,        % first optional argument becomes the box title
    fonttitle=\bfseries,
    listing only,
    listing options={
      language=#2,
      basicstyle=\ttfamily\footnotesize,
      keywordstyle=\color{blue!20!black}\bfseries,
      commentstyle=\itshape\color{green!40!black},
      stringstyle=\color{blue!20!black},
      showstringspaces=false,
      columns=flexible,
      breaklines=true,
      morekeywords={tool_name, tool_description, input_types, output_type, demo_commands, user_metadata}, % Add keywords you want in bold
    },
    left=5pt,
    right=5pt,
    top=-5pt,
    bottom=-5pt,
}

\newcommand{\toolinput}[1]{\textbf{Input (#1):}}
\newcommand{\tooloutput}[1]{\textbf{Output (#1):}}

\newtcblisting{answercodebox}{
    colback=answerbg,
    colframe=white,
    arc=0pt,
    boxrule=0pt,
    listing only,
    listing options={
        language=Python,
        basicstyle=\ttfamily\footnotesize,
        keywordstyle=\color{blue!70!black}\bfseries,
        commentstyle=\itshape\color{green!40!black},
        stringstyle=\color{red!60!black},
        showstringspaces=false,
        columns=flexible,
        breaklines=true,
        inputencoding=utf8,
        extendedchars=true,
        texcl=true,
        escapeinside={(*@}{@*)},  % Define escape characters
    },
    left=0pt,
    right=0pt,
    top=-10pt,
    bottom=-10pt,
}

\newtcblisting{codebox}{
    colback=toolcardbox,
    colframe=white,
    arc=0pt,
    boxrule=0pt,
    listing only,
    listing options={
        language=Python,
        basicstyle=\ttfamily\footnotesize,
        keywordstyle=\color{blue!70!black}\bfseries,
        commentstyle=\itshape\color{green!40!black},
        stringstyle=\color{red!60!black},
        showstringspaces=false,
        columns=flexible,
        breaklines=true,
        inputencoding=utf8,
        extendedchars=true,
        texcl=true,
        escapeinside={(*@}{@*)},  % Define escape characters
    },
    left=0pt,
    right=0pt,
    top=-10pt,
    bottom=-10pt,
}

\title{
\raisebox{-0.5ex}{\includegraphics[height=1.1em]{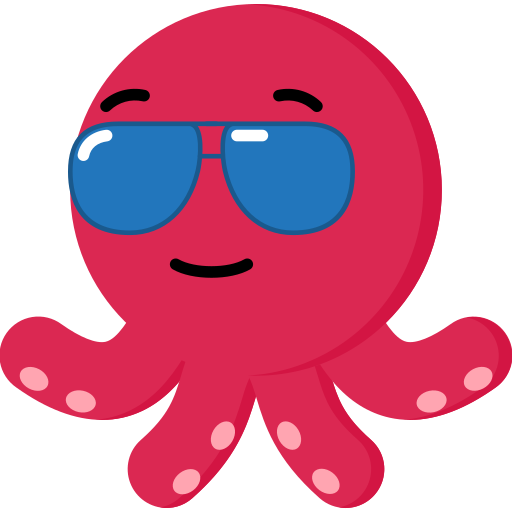}} 
OctoTools: A Multi-Agent Framework with Extensible Tools for \\
Complex Reasoning}

\newcommand{\stanford}{\twemoji{evergreen tree}}

\author{Pan Lu$^{* \stanford}$, Bowen Chen$^{* \stanford}$, Sheng Liu$^{* \stanford}$, Rahul Thapa$^{\stanford}$, Joseph Boen$^{\stanford}$, James Zou$^{\stanford}$
\\
* Equal contribution \hspace{0.3em}
\stanford Stanford University \vspace{0.3em}
\\
\texttt{Website:} \text{\url{https://octotools.github.io}} \vspace{0.3em}
\\
\raisebox{-0.4ex}{\includegraphics[height=1em]{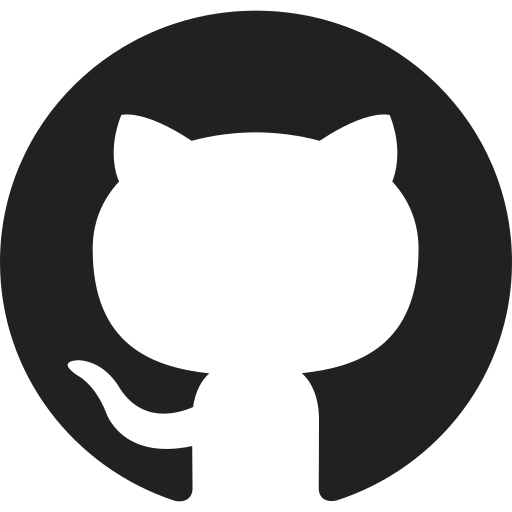}}\hspace{0.3em}\href{https://github.com/octotools/octotools}{Code}\hspace{0.8em}
\raisebox{-0.4ex}{\includegraphics[height=1em]{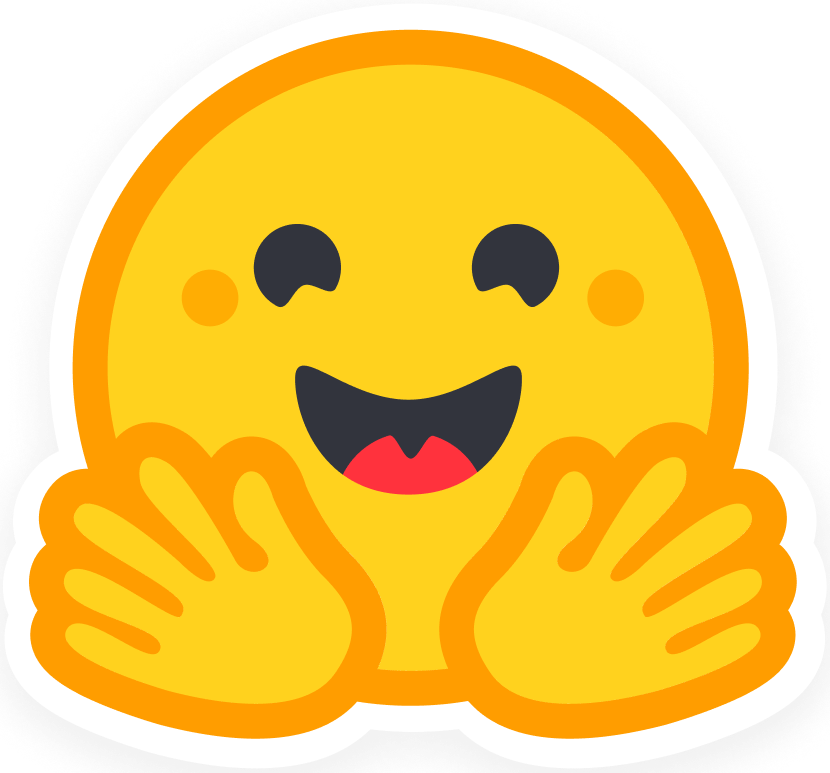}}\hspace{0.3em}\href{https://huggingface.co/spaces/OctoTools/octotools}{Demo}\hspace{0.8em}
\raisebox{-0.4ex}{\includegraphics[height=1em]{logos/octotools.png}}\hspace{0.3em}\href{https://octotools.github.io/\#tool-cards}{Tool Cards}\hspace{0.8em} 
\raisebox{-0.4ex}{\includegraphics[height=1em]{logos/octotools.png}}\hspace{0.3em}\href{https://octotools.github.io/\#visualization}{Visualization}\hspace{0.8em}
\raisebox{-0.4ex}{\includegraphics[height=1em]{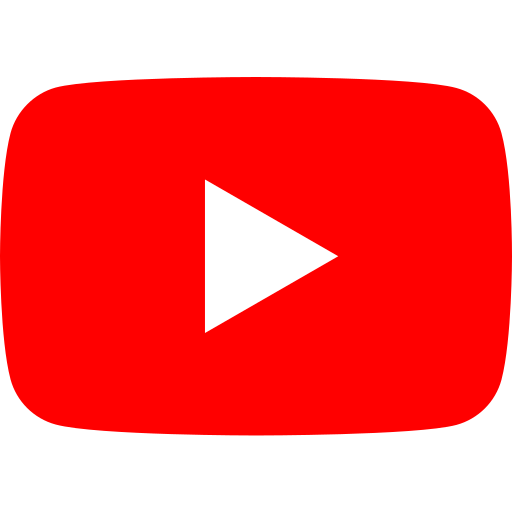}}\hspace{0.3em}\href{https://www.youtube.com/watch?v=4828sGfx7dk}{Video}
}

\begin{document}
\maketitle

\begin{abstract}
Solving complex reasoning tasks may involve visual understanding, domain knowledge retrieval, numerical calculation, and multi-step reasoning. Existing methods augment large language models (LLMs) with external tools but are restricted to specialized domains, limited tool types, or require additional training data. In this paper, we introduce OctoTools, a training-free, user-friendly, and easily extensible multi-agent framework designed to tackle complex reasoning across diverse domains. OctoTools introduces standardized tool cards to encapsulate tool functionality, a planner for both high-level and low-level planning, and an executor to carry out tool usage. We validate OctoTools’ generality across 16 diverse tasks (including MathVista, MMLU-Pro, MedQA, and GAIA-Text), achieving substantial average accuracy gains of 9.3\% over GPT-4o. Furthermore, OctoTools also outperforms AutoGen, GPT-Functions, and LangChain by up to 10.6\% when given the same set of tools. Through comprehensive analysi, ablations, and robustness tests with compact backbones and noisy tool environments, OctoTools demonstrates advantages in task planning, effective tool usage, and multi-step problem solving. \textit{Code, demos, and visualization are publicly available at \url{https://octotools.github.io/}.}
\end{abstract}

\section{Introduction}
\label{sec:intro}
Large language models (LLMs) \citep{brown2020language, chowdhery2022palm, openai2023gpt4} have made rapid progress on tasks such as summarization, translation \citep{thoppilan2022lamda}, code generation \citep{nakano2021webgpt}, and math problem solving \citep{shuster2022blenderbot}. However, complex reasoning tasks that involve multiple steps, logical decomposition, or specialized domain knowledge remain challenging. For example, solving a visual riddle may require fine-grained image understanding and text-based reasoning, while a math or chemistry question can require thorough computations or domain expertise. Existing prompting methods often fail to orchestrate these varied processes into a coherent chain of reasoning \citep{yao2022react}. 

\begin{figure}[t!]
    \centering
    \includegraphics[width=0.38\textwidth]{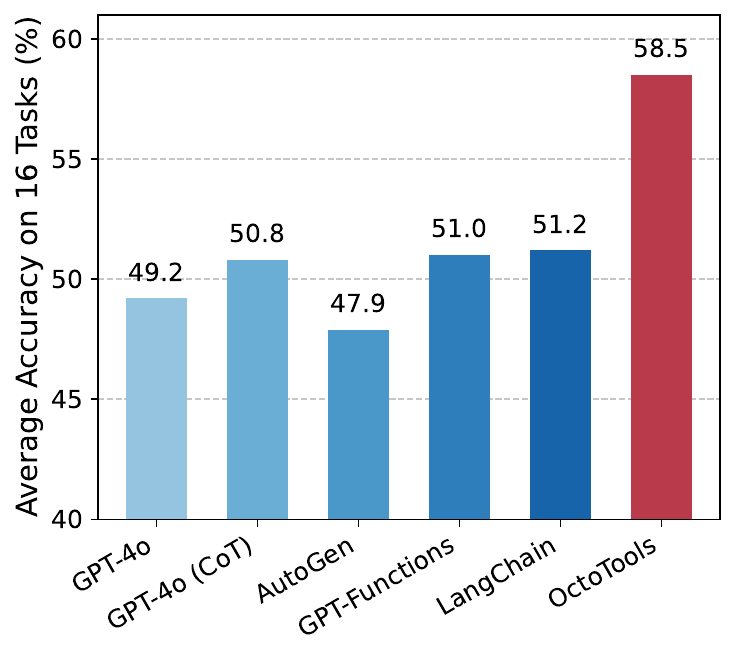}
    \caption{Performance comparison across 16 benchmarks. Our \model framework achieves an average accuracy gain of 9.3\% over GPT-4o without function plugins and 7.3\% over LangChain, under the same configuration.}
    \vspace{-2mm}
\label{fig:main_scores_bar_chart}
\end{figure}

A promising direction to address these challenges is to \emph{augment} LLMs with \emph{external tools}. By offloading specialized subtasks (e.g., web queries, Python-based calculations, and specialized scientific tools) to dedicated modules, LLMs can focus on higher-level planning and synthesis. Several frameworks have explored such tool usage, from those relying on extensive supervised data and fine-tuning \citep{schick2023toolformer, liu2023llava}, to static solutions without refinement \citep{lu2023chameleon}, and those limited to one specialized domain of tools \citep{nakano2021webgpt,tao2023webwise,hu2024visual}. Although these methods perform well on specific tasks, they still face challenges that hinder general widespread use. Many require substantial training with curated data, which limits their adaptability to new domains. Others are designed for a particular domain \citep{bran2023chemcrow,kang2024chatmof,li2024mmedagent,schmidgall2024agentclinic} or cannot easily support multi-step problem-solving \citep{lu2023chameleon}, restricting their generality.

\begin{figure*}[th!]
    \centering
    \includegraphics[width=1.0\textwidth]{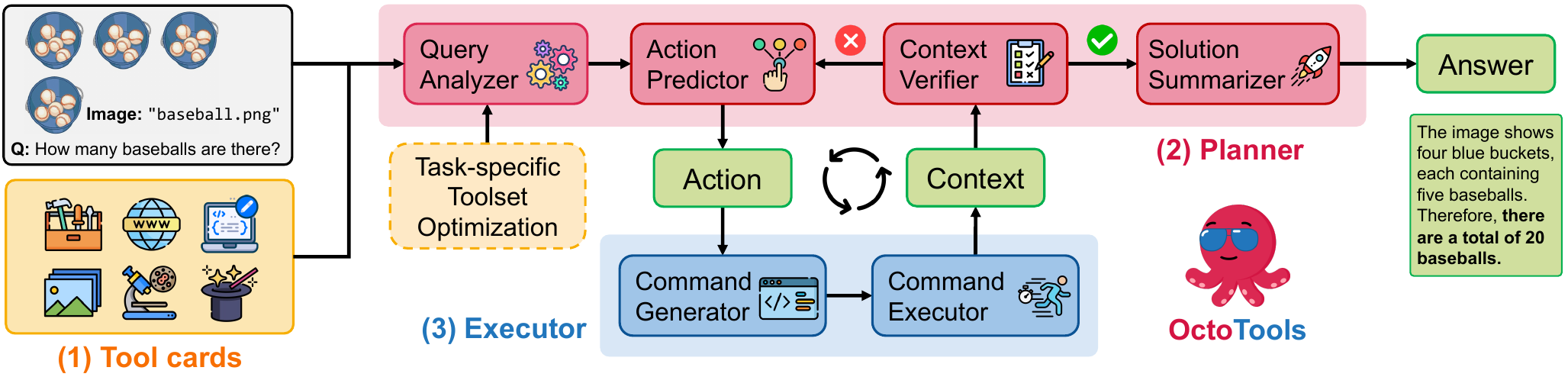}
    \vspace{-4mm}
    \caption{The framework of \model. (1) \emph{Tool cards} define tool metadata and encapsulate tools, enabling integration of new tools without retraining or framework changes. (2) The \emph{planner} agent handles both high-level and low-level planning to achieve the global objective and refine actions step by step. (3) The \emph{executor} agent generates executable commands for tool calls and stores structured results in the context, from which the final answer is summarized. In addition, the \textit{task-specific toolset optimization} algorithm learns to select a beneficial subset of tools for downstream tasks.}
\label{fig:model_framework}
\end{figure*}

\begin{figure*}[th!]
    \centering
    \includegraphics[width=1.0\textwidth]{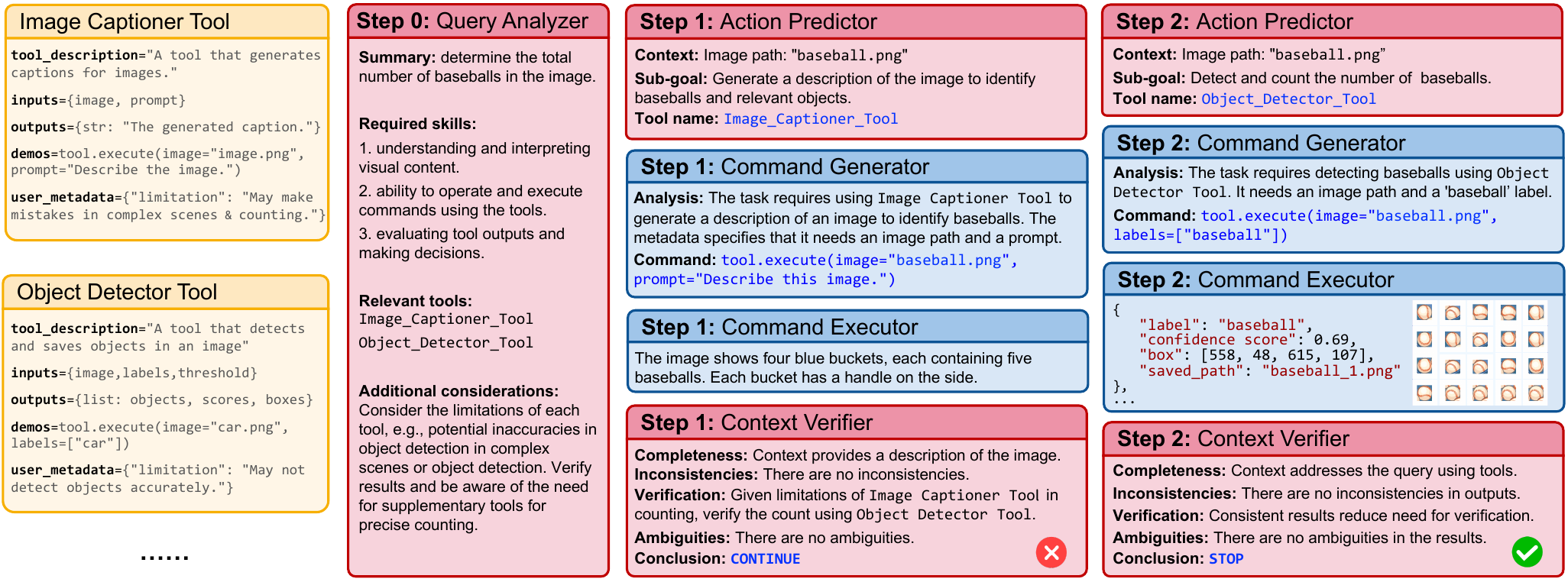}
    \vspace{-4mm}
    \caption{A self-contained example. We show tool cards for selected tools, the initial plan generated by the planner, and two steps where the planner and the executor orchestrate low-level planing and tool usage before producing the final answer. See \S\ref{app:demo_example} for details, \S\ref{app:exp_examples} for additional cases, and the interactive visualization on our project website at \url{https://octotools.github.io/}.}
\label{fig:model_example}
\end{figure*}

In this paper, we propose \model, a \emph{training-free} (i.e., it does not require updating model weights), \emph{user-friendly}, and \emph{extensible} multi-agent framework for tackling \emph{complex reasoning} tasks across diverse domains (Figure~\ref{fig:model_framework}). A key feature of \model is the concept of \emph{tool cards}, standardized wrappers that encapsulate heterogeneous tools (e.g., Python calculators, web search APIs, and domain-specific modules), along with metadata such as input-output formats, usage constraints, and best practices that delineate ideal use cases. This standardized design enables easy integration, replacement, or expansion of tools, unlike approaches that require painstaking re-engineering for each new tool \citep{lu2023chameleon, hu2024visual}.

Building on these tool cards, \model employs a dedicated \emph{planner} agent that governs both high-level and low-level planning. Given a user query, the planner agent proposes a global plan for how tools will be used. At each step, it generates a text-based \emph{action} (including sub-goals and tool selection) conditioned on the evolving \emph{context}. A separate \emph{executor} agent then converts this action into an executable command, runs the tool, and updates the context with the results. This separation of strategic planning from command execution reduces errors and improves transparency, making \model more reliable and easier to maintain.

An additional challenge in agentic systems is determining which subset of tools to enable for a given domain. Although providing many tools can be beneficial, enabling them all may introduce noise or slow performance \citep{lumer2024toolshed, fore2024geckopt, paramanayakam2024less}. To address this, we propose a lightweight \emph{toolset optimization} algorithm that identifies a more useful subset of tools for each task based on validation performance, ultimately improving both accuracy and efficiency.

While recent general agent frameworks also allow LLMs to use external tools autonomously, they often focus on high-level abstractions \citep{langchain}, limited observability of intermediate decisions \citep{gpt4oplugin}, or multi-agent collaboration features  \citep{autogen}, with less emphasis on enhancing complex reasoning and \emph{quantitatively} benchmarking downstream task performance. In contrast, we systematically evaluate the entire agentic workflow of \model across diverse tasks, including controlled comparisons in which competing frameworks share the same backbone, tool implementations, and reasoning budgets. This isolates architectural benefits from tool coverage and provides in-depth analyses of when and how tool-based reasoning succeeds or fails in complex reasoning scenarios.

We conduct large-scale experiments across 16 reasoning benchmarks spanning vision, mathematical, scientific, medical, and agentic domains. As shown in Figure~\ref{fig:main_scores_bar_chart}, \model substantially outperforms baselines, with an average accuracy gain of 9.3\% over zero-shot prompting by \gpt and 7.7\% over chain-of-thought (CoT), as well as up to 10.6\% improvement over existing agentic frameworks with the same tools \citep{autogen, gpt4oplugin, langchain}. Analyses show that \model effectively combines multi-step planning and specialized tool usage, with each providing distinct improvements. Tool usage is particularly valuable for intricate calculations or specialized knowledge, while multi-step planning offers significant gains for reasoning decomposition.

Furthermore, our comprehensive studies provide insights into \model's performance and robustness under different conditions. Accuracy increases with higher step budgets, highlighting the value of longer multi-step chains. Even without toolset optimization, enabling the full toolset reaches 57.4\% accuracy, 3.5\% above the base tool alone, suggesting benefits from broader tool availability. 
A lightweight, task-specific toolset search further lifts accuracy to 58.9\%. Robustness holds under weaker and smaller backbones: with \gptmini, \model still delivers an average gain of 7.1\% across 16 tasks, and on compact Qwen2.5 3B/7B models it improves average accuracy by 6.3--13.6\% on representative tasks. It also degrades gracefully under injected random tool failures, indicating resilience to noisy tool environments.
Our in-depth analysis further underscores the framework's robustness, with a failure case study showing that errors typically stem from tool limitations rather than the core system. A human study on our interactive demo complements this, where users praised the system's transparency and interpretability.

\textbf{Our contributions} are as follows:
(1) We propose \model, a \textbf{training-free}, modular, and \textbf{extensible} multi-agent framework that enables LLMs to solve complex reasoning tasks by planning and executing multi-step tool calls without fine-tuning.
(2) We introduce a new \textbf{planner-executor paradigm} with standardized \textbf{tool cards} to simplify tool integration and improve reliability. We complement this with a lightweight \textbf{toolset optimization algorithm} that automatically selects effective tools for a given task.
(3) We conduct \textbf{large-scale experiments on 16 diverse benchmarks}, providing one of the most systematic evaluations of agentic systems to date. These experiments show that \model achieves substantial gains over strong baselines, including leading agentic frameworks.
(4) We provide \textbf{in-depth analyses and ablations} that disentangle the contributions of multi-step planning and tool usage. To promote future research, we release our codebase, tool-card templates, and interactive demos, offering practical guidance for the community.

\section{The \model Framework}
\label{sec:system}

\subsection{Preliminary}
We propose \model, an open-source, versatile, and user-friendly multi-agent framework for complex reasoning tasks, as illustrated in Figure~\ref{fig:model_framework}. 
Given a user query $q \in \mathcal{Q}$ and a pretrained language model $\text{LLM}_\theta(\cdot)$, a naive approach would generate an output directly as $y \sim \text{LLM}_\theta(q)$, providing a single-step response. In contrast, our \model framework introduces a structured, multi-step process that leverages external tools to tackle queries effectively.

Specifically, \model contains a set of \textit{tools} $\mathcal{D} = \{d_i\}_{i=1}^n$ and associated metadata $\mathcal{M} = \{m_i\}_{i=1}^n$, where $n$ is the number of available tools. Given a query, a \textit{planner} agent (based on a language model) first generates a \textit{tentative plan} from a high-level perspective, indicating how these tools can be used to address the query, which forms the initial \textit{context} $s_0$. From this plan, the planner determines the initial \textit{action} $a_1$ for tool usage, specifying which tool $d_1$ to use, the relevant context, and a sub-goal. An \textit{executor} agent (also powered by a language model) then converts the planner's text-based action $a_1$ into a machine-executable \textit{command} $o_t$, which is run to obtain intermediate results $r_1$. These results, along with the original action, update the context to $s_1 := (a_1, o_1, r_1)$. This process constitutes one step in our framework.

This process repeats, with the planner iteratively refining its actions based on the evolving context until it either finds a complete solution or inference limits (\textit{e.g.}, time or steps) are reached. After $T$ steps, the framework produces a full \textit{trajectory} $(s_0, s_1, \dots, s_T)$, which is stored in a structured manner in the context. The planner then uses this trajectory to generate the final solution to the original query.

\model provides a robust and effective framework for solving complex tasks through sub-goal decomposition and systematic tool usage. Standardized \textit{tool cards} encapsulate functionality (\S\ref{sec:toolcards}), the \emph{planner} agent orchestrates both high-level and low-level task planning (\S\ref{sec:agent_planning}), and the \emph{executor} agent instantiates tool calls for each sub-goal (\S\ref{sec:agent_execution}).

\subsection{Tool Cards} 
\label{sec:toolcards}

To enable seamless interaction among tools, the planner, and the executor, the toolbox serves as a fundamental building block of \model. In contrast to previous work~\cite{lu2023chameleon,hu2024visual} that rely on careful adaptations and tuning to support new tools, \model leverages \textit{tool cards}, which encapsulate the specifications of tools in a structured manner. 

Each \textit{tool card} represents a tool $d$ alongside its essential metadata $m$ (as illustrated in Figure \ref{fig:model_example}). This metadata includes the tool's name, version, input and output types, and command demonstrations. It may also contain additional usage constraints (\textit{e.g.}, user-defined ``\textit{limitations}'' and ``\textit{best practices}''), which provide developers’ insights to guide the planner and executor. For example, \texttt{Image\_Captioner\_Tool} includes ``\textit{it may make mistakes in complex scenes}'' and ``\textit{consider using other tools for context/verification}'' in the user metadata (\S\ref{app:image_captioner_tool}). See \S\ref{app:toolcards} for details on featured tools.
The design of each tool card is modular relative to the framework, enabling users to integrate diverse tools without modifying the underlying framework or agent logic. 
New tool cards can be added, replaced, or updated with low effort.
In practice, adding a tool requires only lightweight natural-language metadata and a few demonstration commands, making new tool cards easy to add, replace, or update with low effort. Because the schema is standardized, this authoring process can also be partially automated with LLM assistance, which we view as a promising direction for future work.

\subsection{Planner Agent}
\label{sec:agent_planning}

\paragraph{Planner initialization.}
The planner agent inspects the \emph{toolbox} and loads each \emph{tool card} as defined in \S\ref{sec:toolcards}. Formally, it constructs a set $\{d_i\}_{i=1}^n$ of $n$ available tools, each with metadata $m$ that describes its input-output schema and usage constraints. Rather than enabling the full toolset, one might enable a subset $\mathcal{D^\text{*}} = \{d_i\}_{i=1}^k$ chosen based on expert insights or optimized with a small set of examples.

\paragraph{Query analysis and action prediction.}
Given a query $q$, the planner agent formulates a \emph{high-level}, tentative plan for tool usage based on its initialization. As shown in Figure \ref{fig:model_framework}, the high-level plan summarizes the query objective, analyzes the required skills, identifies relevant tools, and includes additional considerations to highlight the need for verification. The \emph{high-level} task plan provides a global perspective on the final objective, ensuring that each subsequent sub-goal remains aligned with the original query.

Subsequently, at step $t$, an \emph{action predictor} produces an action $a_t$ that combines the planner-suggested sub-goal (e.g., \emph{``detect baseballs in the image''}), the selected tool $d$ (e.g. \texttt{Object\_Detector}), and the relevant context. This \emph{low-level} plan refines and executes each sub-goal in real time, adapting to new information or feedback at each step.

\paragraph{Context verification and solution summarization.}
After each command is executed, a \emph{context verifier} checks whether the problem can be solved given the current context. It verifies completeness (e.g., whether all sub-goals are satisfied) and identifies any ambiguities. If the problem remains incomplete, the planner continues the next iteration of the cycle by predicting the next action $a_{t+1}$.
Once the verifier concludes that the query has been solved, a separate \emph{solution summarizer} compiles the final answer from the trajectory $(s_0,s_1,\dots, s_T)$. This stage integrates intermediate tool outputs, traces reasoning steps, and presents a concise, user-friendly summary as the final solution.

\subsection{Executor Agent}
\label{sec:agent_execution}

\paragraph{Command prediction.} Prior work~\cite{lu2023chameleon, hu2024visual} often expects a single language model both for planning each step (i.e., which tool to use) and for generating the corresponding executable command. This dual responsibility can overload the model and lead to errors, especially when dealing with complex or environment-specific code \cite{bran2023chemcrow,li2024formal,ji2024testing}. To mitigate these issues, \model introduces a \emph{command generator} that interprets the planner's text-based actions and produces executable code.
Given the action $a_t$ predicted by the planner, the \emph{command generator} (powered by a language model) creates a low-level command $o_t$ in the form of an executable Python script, which calls the tool $d_t$ with necessary inputs and performs any required data preparation. This step bridges the abstract action specified in $a_t$ and the concrete tool call. 

\paragraph{Command execution.}
Once an executable command is generated, it must be run in an environment that may involve dependencies, external libraries, or resource access (e.g., file systems). Directly coupling execution with planning poses security and maintainability challenges, especially if the planner is not capable of managing code execution details.
In \model, an \emph{command executor} runs the generated command $o_t$ in a Python environment, obtaining a result $r_t$. This may include tool outputs, logs, or error messages. The executor then adds the current context of this step $s_t:=(a_t, o_t, r_t)$ to the agent's current trajectory $(s_0, s_1, \ldots, s_{t-1})$. The trajectory preserves a clear history of the actions taken, the code produced, and the results obtained.

\subsection{Task-specific Toolset Optimization}
\label{sec:toolset_optimization}

The \model toolbox contains a diverse set of tools covering different modalities and skills. By leveraging structured tool cards and robust planning capabilities, \model demonstrates strong generality when all available tools are enabled across different tasks (see \S\ref{sec:toolset_optimization}). However, when a small set of validation examples are available for a task, configuring a \emph{task-specific} subset of tools can further enhance efficiency and effectiveness.
To this end, we propose an automated algorithm to optimize the toolset configuration for each task. Given $n$ available tools in the toolbox, the total number of possible subsets is $O(2^n)$, which is prohibitively large. To make this tractable, we employ a greedy search strategy that reduces the complexity to $O(n)$. Our algorithm proceeds in three stages.

\begin{algorithm}[!th]
    \small
    \caption{Task-specific Toolset Optimization}
    \label{alg:tool_selection_optimization}
    \begin{algorithmic}[1]
    \STATE \textbf{Input:} Toolbox $\mathcal{D} = \{d_i\}_{i=1}^n$, base toolset $\mathcal{D}_\text{base}$
    \STATE \textbf{Output:} Optimized toolset $\mathcal{D}^*$
    
    \STATE \COMMENTT{Stage 1: Baseline setup}
    % \STATE $\mathcal{D}_\text{base} \gets \{ d_\text{base} \}$
    \STATE $\text{Acc}_\text{baseline} \gets \text{Acc}(\mathcal{D}_\text{base})$
    
    \STATE \COMMENTT{Stage 2: Individual tool evaluation}
    \FOR {each $d_i$ in $D$ \textbf{such that} $d_i \notin \mathcal{D}_\text{base}$}
        \STATE $\mathcal{D}_i \gets \mathcal{D}_\text{base} \cup \{ d_i \}$
        \STATE $\text{Acc}_i \gets \text{Acc}(\mathcal{D}_i)$
        \STATE $\Delta_{d_i} \gets \text{Acc}_i - \text{Acc}_\text{baseline}$
        \IF{$\Delta_{d_i} > 0$}
            \STATE $\mathcal{D}_\text{beneficial} \gets \mathcal{D}_\text{beneficial} \cup \{ d_i \}$
        \ENDIF
    \ENDFOR
    
    \STATE \COMMENTT{Stage 3: Select optimized toolset}
    \STATE $\mathcal{D}^* \gets \mathcal{D}_\text{beneficial} \cup \mathcal{D}_\text{base}$
    
    \STATE \textbf{Return} $\mathcal{D}^*$
    \end{algorithmic}
\end{algorithm}

\paragraph{Stage 1: Baseline setup.} 
We establish a baseline performance by enabling the base toolset in the toolbox, denoted $\mathcal{D}_\text{base}$. This base set represents a minimal starting toolset, which can be pre-defined by the user. 

\paragraph{Stage 2: Individual tool evaluation.}
Next, we measure each candidate tool $d_i$ by enabling it alongside the base toolset. For each $d_i$, we form an augmented subset 
$\mathcal{D}_i = \mathcal{D}_\text{base} \cup \{d_i\}$ and compute the accuracy difference on the set of validation examples
\vspace{-3mm}
\begin{equation}
\Delta_{d_i} \;=\; \text{Acc}\bigl(\mathcal{D}_i\bigr) \;-\; \text{Acc}\bigl(\mathcal{D}_\text{base}\bigr).
\vspace{-3mm}
\end{equation}
If $\Delta_{d_i} > 0$, we consider the tool $d_i$ beneficial.

\paragraph{Stage 3: Optimized toolset selection.}
Finally, we aggregate all tools that yield positive improvements and combine them with the default set to form the optimized toolset:
\vspace{-3mm}
\begin{equation}
\mathcal{D}^* = \mathcal{D}_\text{base}\;\cup\;\{\, d_i \mid \Delta_{d_i} > 0 \}.
\vspace{-3mm}
\end{equation}
This set contains all tools that individually demonstrate a performance gain over the baseline, ensuring a customized yet efficient configuration for the downstream task. The optimized subset is selected once per task to define the candidate toolbox, while actual tool usage remains query-dependent through the planner-executor-verifier loop. While this selection does not guarantee global optima, we observe overall improvements over simply using all tools (see \S\ref{sec:ablation_study}).

\section{Experiments}
\label{sec:exp}

\subsection{Experimental Setups}
\label{sec:exp_setups}

To demonstrate the generality of \model, we evaluate it on 16 benchmarks spanning two modalities, five domains, and four reasoning types (Table~\ref{tab:main_results_across_benchmarks}). These benchmarks cover diverse reasoning tasks, including visual understanding, numerical calculation, knowledge retrieval, and multi-step reasoning.

For general visual reasoning, we include AlgoPuzzleVQA~\cite{ghosal2024language}, Hallusion-VD~\cite{guan2024hallusionbench}, PuzzleVQA~\cite{chia2024puzzlevqa}, and  VQA 2.0~\cite{goyal2017making}. For mathematical reasoning, we use Game of 24~\cite{24game}, Omni-MATH~\cite{gao2024omni}, CLEVR-Math~\cite{lindstrom2022clevr}, and MathVista~\cite{lu2024mathvista}. For scientific reasoning, we adopt GPQA~\cite{rein2023gpqa}, MMLU-Pro~\cite{wang2024mmlu}, and SciFIBench~\cite{roberts2024scifibench}. To evaluate models in specialized medical domains, we test on MedQA~\cite{jin2021disease}, PathCLS~\cite{sun2025pathmmu}, PathVQA~\cite{he2020pathvqa}, and SLAKE~\cite{liu2021slake} Additionally, we incorporate GAIA-Text, a textual subset of the GAIA~\cite{mialon2023gaia} benchmark designed to evaluate agentic frameworks with tool-calling capabilities. See \S\ref{app:benchmark_datasets} for additional details.

For each benchmark, we sampled 100 examples to construct a validation set for toolset optimization (\S\ref{sec:toolset_optimization}) and ablation studies (\S\ref{sec:ablation_study}), and set aside a held-out test set of 200 examples for final evaluation. For benchmarks with fewer than 300 total samples, the test set consisted of all remaining samples not used for validation. To mitigate randomness, we report the average accuracy with standard deviation across three trials for all experiments, including ablation studies. Details on experimental setups are provided in \S\ref{app:exp_details}. 

% Table

\begin{table*}[th!]
% \vspace{-6mm}
\centering
 \small
 \resizebox{1.0\linewidth}{!}{
\begin{tabular}{lcccccc|cccc|>{\columncolor{gray!20}}r>{\columncolor{gray!20}}r}
\toprule
\textbf{Datasets} & \textbf{Modality}  & \textbf{Domain} & \vision & \calculator & \knowledge & \steps & \textbf{0-shot} & \textbf{CoT} & \textbf{\modelbase} & \textbf{\model} & \textbf{$\Delta$ (0-shot)} & \textbf{$\Delta$ (CoT)} \\
\midrule
AlgoPuzzleVQA & Vision & General & \cmark & & & \cmark & 41.3\std{0.3} & 42.7\std{1.0} & 44.0\std{0.9} & \textbf{48.7}\std{0.3} & +7.4 & +6.0 \\
Hallusion-VD & Vision & General & \cmark & & &  & 52.0\std{1.0} & 53.3\std{2.1} & 59.0\std{0.0} & \textbf{63.3}\std{2.9} & +11.3 & +10.0 \\
PuzzleVQA & Vision & General & \cmark & & & \cmark & 52.2\std{1.0} & 54.0\std{1.3} & 59.3\std{0.8} & \textbf{61.0}\std{0.5} & +8.8 & +7.0 \\
VQA 2.0 & Vision & General & \cmark & & & \cmark & 50.3\std{1.0} & 48.7\std{0.3} & 47.2\std{0.8} & \textbf{54.5}\std{0.0} & +4.2 & +5.8 \\
\midrule
Game of 24 & Text & Mathematical & & \cmark & & \cmark & 22.2\std{2.5} & 33.3\std{1.5} & 37.8\std{3.3} & \textbf{44.7}\std{2.8} & +22.5 & +11.4 \\
Omni-MATH & Text & Mathematical & & \cmark & \cmark & \cmark & 27.0\std{0.0} & 29.3\std{1.3} & 30.2\std{0.6} & \textbf{32.2}\std{0.8} & +5.2 & +2.9 \\
CLEVR-Math & Vision & Mathematical & \cmark & \cmark & &  & 64.5\std{3.0} & 75.2\std{1.5} & 68.8\std{0.8} & \textbf{79.0}\std{0.9} & +14.5 & +3.8 \\
MathVista & Vision & Mathematical & \cmark & \cmark & \cmark & \cmark & 59.3\std{0.8} & 59.5\std{1.5} & 63.0\std{1.3} & \textbf{64.3}\std{1.0} & +5.0 & +4.8 \\
\midrule
GPQA & Text & Scientific & & \cmark & \cmark & \cmark & 53.7\std{1.9} & 52.3\std{2.0} & 53.7\std{2.5} & \textbf{54.7}\std{1.3} & +1.0 & +2.4 \\
MMLU-Pro & Text & Scientific &  & & \cmark & \cmark & 71.7\std{0.3} & 70.3\std{0.6} & 71.5\std{1.3} & \textbf{73.7}\std{1.3} & +2.0 & +3.4 \\
SciFIBench & Vision & Scientific & \cmark &  & \cmark & & 72.5\std{0.0} & 75.0\std{0.9} & 77.3\std{0.8} & \textbf{78.3}\std{0.6} & +5.8 & +3.3 \\
\midrule
MedQA & Text & Medical & & & \cmark &  & 84.5\std{1.0} & 84.8\std{0.6} & \textbf{92.8}\std{0.6} & 91.5\std{1.8} & +7.0 & +6.7 \\
PathCLS & Vision & Medical & \cmark & & \cmark & & 36.0\std{0.9} & 37.5\std{1.8} & 37.0\std{1.8} & \textbf{58.2}\std{1.3} & +22.2 & +20.7 \\
PathVQA & Vision & Medical & \cmark & & \cmark & \cmark & 32.0\std{1.8} & 27.8\std{1.8} & 43.5\std{2.6} & \textbf{49.2}\std{1.2} & +17.2 & +21.4 \\
SLAKE & Vision & Medical & \cmark & & \cmark & \cmark & 59.3\std{1.0} & 60.3\std{0.6} & 59.2\std{1.8} & \textbf{63.8}\std{1.4} & +4.5 & +3.5 \\
\midrule
GAIA-Text & Text & Agentic & & \cmark & \cmark & \cmark & ~~8.7\std{0.8} & ~~8.4\std{0.5} & ~~9.7\std{0.9} & \textbf{18.4}\std{1.2} & +9.7 & +10.0 \\
\midrule
\rowcolor{red!30} \textbf{Average (\%)} & - & - & - & - & - & - & \multicolumn{1}{l}{49.2} & 50.8~~~~~~~ & 53.4~~~~~~~ & \textbf{58.5}~~~~~~~ & \textbf{+9.3} & \textbf{+7.7} \\
\bottomrule
\end{tabular}
}
\vspace{-1mm}
\caption{Main results across 16 benchmarks spanning different modalities, domains, and required reasoning skills: visual understanding (\vision), numerical calculation (\calculator), knowledge retrieval (\knowledge), and multi-step reasoning (\steps). \modelbase uses only the base tool (\texttt{Generalist\_Solution\_Generator}), while \model uses the optimal toolset. Performance gains ($\Delta$) are computed for OctoTools relative to both zero-shot (0-shot) and chain-of-thought (CoT) baselines, with \model achieving 58.5\% average accuracy and improvements of 9.3\% and 7.7\%, respectively.}
 % All results show average accuracy and standard deviation (gray) over three trials.
\vspace{-1mm}
\label{tab:main_results_across_benchmarks}
\end{table*}

We created a diverse array of tools in the toolbox for our experiments. The base toolset consists of \texttt{Generalist\_Solution\_Generator}, a tool built on top of the base LLM that takes as input a specialized prompt generated by the executor, such as ``\textit{Describe the image in detail}'' or ``\textit{Generate stepwise solutions for the math problem}''. This tool allows for general step-by-step reasoning without requiring external tools but lacks domain specificity. The toolbox also includes image perception tools such as \texttt{Image\_Captioner}, web search APIs like \texttt{Google\_Search}, a code generator \texttt{Python\_Code\_Generator}, and specialized tools like \texttt{Path\_Generalist\_Classifier} for pathology image classification. Among them, \texttt{Relevant\_Patch\_Zoomer} is a unique tool that takes a textual query and returns zoomed-in quarter patches, which provide localized details for fine-grained visual reasoning scenarios. See \S\ref{app:tool_descriptions} and \S\ref{app:toolcards} for more details.

\subsection{Main Results}
\label{sec:main_results}

We compare the performance of our framework after toolset optimization (denoted as \model) against three baselines: (1) zero-shot, where the base LLM directly answers queries without additional prompting, (2) chain-of-thought (CoT), where the base LLM is prompted to ``Think step by step'' to generate step-by-step reasoning, and (3) \modelbase, which uses only the base tool without external integrations. Unless otherwise specified, all results presented below use \gptengine as the base model.

Table~\ref{tab:main_results_across_benchmarks} and Figure~\ref{fig:main_scores_bar_chart} summarize performance across all 16 benchmarks. \model achieves consistent gains, outperforming zero-shot and CoT baselines by 9.3\% and 7.7\% on average, respectively. \modelbase also demonstrates improvements over zero-shot (4.2\%) and CoT (2.6\%), indicating that our framework's step-by-step reasoning contributes significantly to performance, independent of external tool integration. Figure~\ref{fig:benchmark_gains_radar_barplot} visualizes gains over zero-shot. Detailed analyses of these gains are presented in \S \ref{sec:analysis_of_gains}. 

% Table

\begin{table}[t!]
\centering
\small
\renewcommand\tabcolsep{3.0pt}
\resizebox{1.0\linewidth}{!}{
    \begin{tabular}{lcccc}
    \toprule
    \textbf{Datasets} & \textbf{AutoGen} & \textbf{\gptplugin} & \textbf{LangChain} & \textbf{\model} \\
    \midrule
    AlgoPuzzleVQA & 44.0\std{1.0} & 44.5\std{0.5} & 42.7\std{2.8} & \textbf{48.7}\std{0.3} \\
    Hallusion-VD & 52.7\std{4.7} & 57.0\std{1.7} & 53.7\std{3.1} & \textbf{63.3}\std{2.9} \\
    Puzzle VQA & 40.0\std{2.3} & 52.5\std{2.8} & 53.5\std{7.8} & \textbf{61.0}\std{0.5} \\
    VQA 2.0 & 46.0\std{1.0} & 45.5\std{0.9} & 54.0\std{1.0} & \textbf{54.5}\std{0.0} \\
    \midrule
    Game of 24 & 24.2\std{2.4} & 34.5\std{2.3} & 18.3\std{4.1} & \textbf{44.7}\std{2.8} \\
    Omni-MATH & 28.5\std{1.3} & 22.8\std{1.8} & 29.7\std{0.6} & \textbf{32.2}\std{0.8} \\
    CLEVR-Math & 69.5\std{3.9} & 71.2\std{1.0} & 69.2\std{4.6} & \textbf{79.0}\std{0.9} \\
    MathVista & 24.7\std{2.5} & 54.5\std{2.0} & 55.7\std{0.3} & \textbf{64.3}\std{1.0} \\
    \midrule
    GPQA & 48.7\std{2.9} & 45.8\std{2.6} & 52.2\std{1.2} & \textbf{54.7}\std{1.3} \\
    MMLU-Pro & 65.0\std{2.5} & 65.8\std{2.4} & 70.3\std{1.2} & \textbf{73.7}\std{1.3} \\
    SciFIBench & 70.0\std{2.2} & 68.8\std{3.2} & 77.0\std{0.5} & \textbf{78.3}\std{0.6} \\
    \midrule
    MedQA & 83.7\std{2.8} & 84.8\std{0.3} & 73.7\std{0.6} & \textbf{91.5}\std{1.8} \\
    PathCLS & 58.0\std{1.3} & 58.2\std{0.6} & 56.3\std{1.3} & \textbf{58.2}\std{1.3} \\
    PathVQA & 42.7\std{0.8} & 42.8\std{2.3} & 45.7\std{4.4} & \textbf{49.2}\std{1.2} \\
    SLAKE & 62.2\std{1.8} & 59.7\std{1.9} & 59.3\std{0.8} & \textbf{63.8}\std{1.4} \\
    \midrule
    GAIA-Text & ~~6.3\std{0.8} & ~~7.9\std{0.8} & ~~7.6\std{1.2} & \textbf{18.4}\std{1.2} \\
    \midrule
    \rowcolor{red!30} \textbf{Average (\%)} & 47.9~~~~~~~ & 51.0~~~~~~~ & 51.2~~~~~~~ & \textbf{58.5}~~~~~~~ \\
    \bottomrule
    \end{tabular}
}
% \vspace{-3mm}
\caption{Comparison of agent frameworks with the same toolbox. \model outperforms the next best by 7.3\%.}
% Results are averaged over three trials.
\vspace{-5mm}
\label{tab:main_results_across_baselines}
\end{table}

\subsection{Comparisons with Other Agent Frameworks}
\label{sec:comparisons_with_other_agent_frameworks}

In addition, we compare three commonly used general AI agent frameworks: \gptplugin~\cite{gpt4oplugin}, \langchain~\cite{langchain}, and \autogen~\cite{autogen}. \gptplugin enables \gpt to call user-specified tools via function calling. \langchain is a framework providing multi-agent collaboration, long-term memory, and tool usage. \autogen is a recent agentic framework that creates multiple autonomous agents with tool usage.

For a standardized comparison of each system's ability to plan and use tools over multiple steps, we configure all agent frameworks, including \model, to use the same underlying model (\gpt) and hyperparameters. They share the same toolset (with the same implementations of tools), a maximum reasoning budget of 10 steps, and a time budget of 300 seconds. Under this controlled setup, performance differences are attributable to the agent architecture and workflow rather than differences in tool coverage. See \S \ref{app:exp_details} for further details.

\begin{figure*}[th!]
    \centering
    \vspace{-1mm}
    \includegraphics[width=0.95\textwidth]{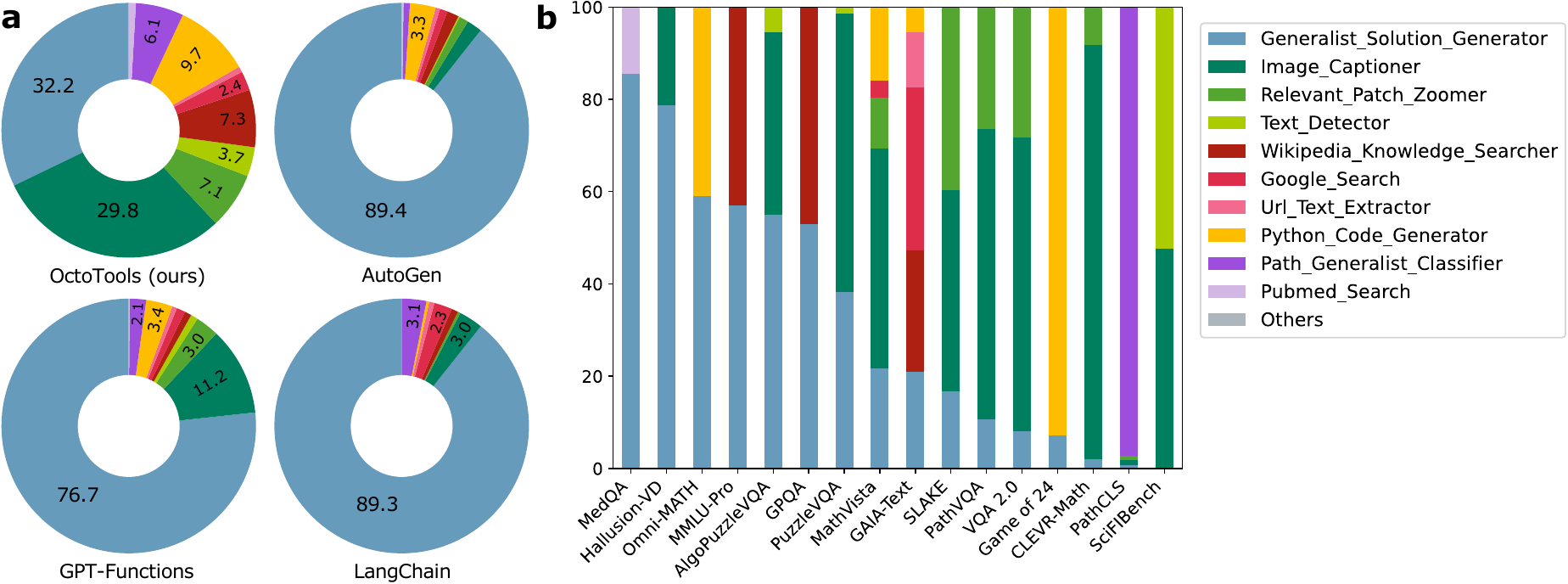}
    \vspace{-1mm}
    \caption{\textbf{a.} Tool usage distribution in \model and agent baselines averaged from 16 tasks. \textbf{b.} Tool usage distribution across 16 tasks in \model. \model takes advantage of different external tools to address task-specific challenges.}
    \vspace{-1mm}
    \label{fig:tool_usage_ours_baselines}
\end{figure*}

Table~\ref{tab:main_results_across_baselines} and Figure~\ref{fig:agent_baselines_bar_plot} show the performance of \model compared to other agent frameworks across 16 tasks. Overall, \model outperforms other agent frameworks, achieving an average accuracy gain of 10.6\% over \autogen, 7.5\% over \gptplugin, and 7.3\% over \langchain. See \S \ref{sec:analysis_of_gains} for detailed analysis.

\subsection{Analysis of Performance Gains}
\label{sec:analysis_of_gains}

Our agent framework includes three aspects for improvement in complex reasoning: \textit{task planning}, \textit{external tool calling}, and \textit{multi-step problem solving}. 

\paragraph{Tool usage distribution.} Figure~\ref{fig:tool_usage_ours_baselines} (a) shows the tool usage distribution for \model and the agent baselines, averaged over 16 tasks. \model takes advantage of \textit{task planning} by calling both a base tool to decompose the query into subtasks (32.2\%) and external specialized tools (67.8\%) for reasoning skills such as fine-grained image understanding, domain knowledge access, precise calculation, and other domain-specific tasks. In contrast, the other agent baselines exhibit limitations in task planning for external tool usage, with external tool usage rates of 10.6\% for \autogen, 23.3\% for \gptplugin, and 10.7\% for \langchain.

Figure~\ref{fig:tool_usage_ours_baselines}~(b) illustrates the tool usage distribution of \model across the 16 tasks. \model adapts to each task by selecting the most suitable tools. For instance, \model employs five different tools to address the diverse challenges in GAIA-Text. In MathVista, a benchmark featuring multiple diagram-based and mathematical reasoning problems, it uses \texttt{Relevant\_Patch\_Zoomer} for local vision perception, \texttt{Google\_Search} for web queries, and \texttt{Python\_Code\_Generator} for precise calculations.

\begin{figure}[th!]
    \centering
    \includegraphics[width=0.45\textwidth]{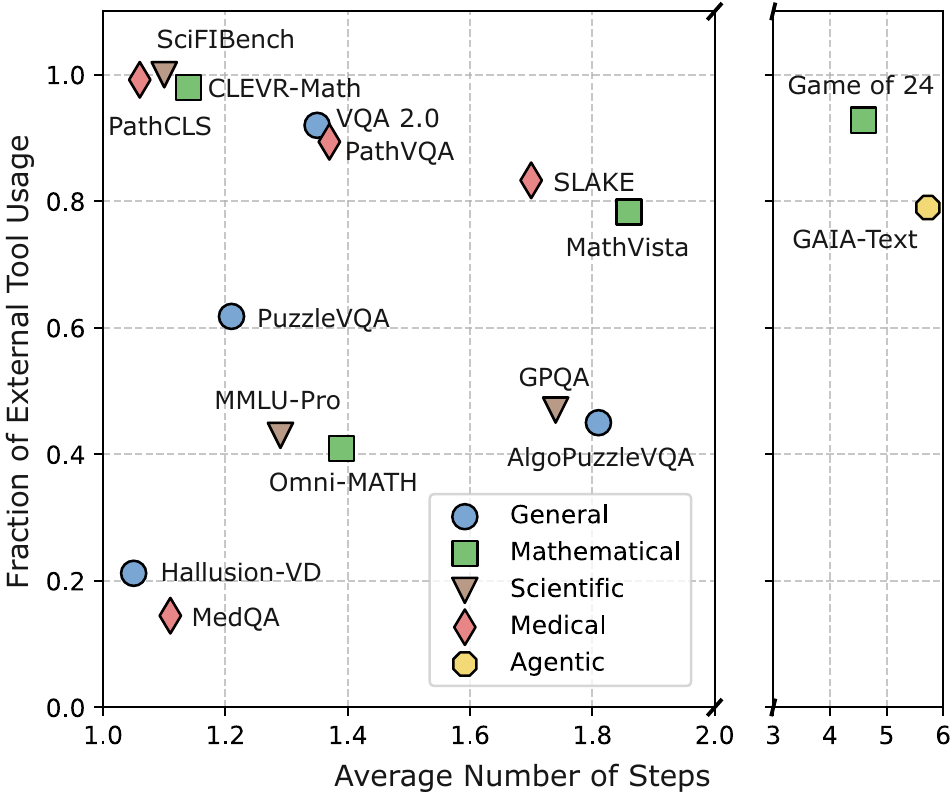}
    \vspace{-1mm}
    \caption{Benchmark distribution across average number of steps and fraction of external tool usage (excluding the base tool \texttt{Generalist\_Solution\_Generator}) in \model.}
    \vspace{-2mm}
    \label{fig:benchmark_distribution_x_steps_y_tools}
\end{figure}

\paragraph{External tool usage vs. multiple steps.} 
Figure~\ref{fig:benchmark_distribution_x_steps_y_tools} shows how tasks are distributed based on the average number of steps and the fraction of external tool usage in \model. Tasks with a high average number of steps indicate that \textit{multi-step problem solving} is valuable, while a high proportion of external tool usage highlights the benefits of \textit{external tool calling}. Notably, two tasks score highly on both dimensions: Game of 24, which frequently calls \texttt{Python\_Code\_Generator} to explore and verify arithmetic expressions that evaluate to 24, and GAIA-Text, which requires multiple tools to address a variety of underlying skills.

\section{Ablation Study}
\label{sec:ablation_study}

This section explores several factors that affect \model's performance, using 100 validation samples.

\begin{figure}[th!]
    \centering
    \vspace{-1mm}
    \includegraphics[width=0.4\textwidth]{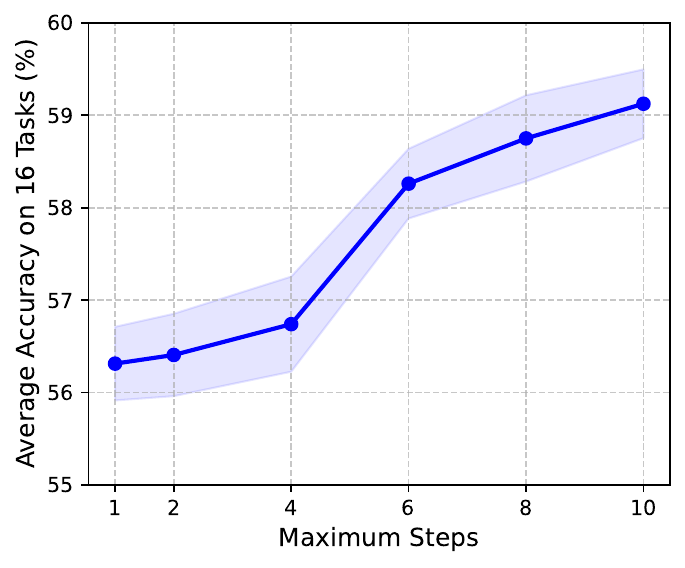}
    \vspace{-1mm}
    \caption{Average accuracy across 16 benchmarks with respect to maximum allowed reasoning steps in \model.}
    \vspace{-1mm}
\label{fig:average_accuracy_by_max_steps}
\end{figure}

\paragraph{Number of maximum allowed steps.}
We explore how the behavior and performance of \model change under different maximum step limits. We report the average over three trials, with results summarized in Figure~\ref{fig:average_accuracy_by_max_steps}.
Overall, performance tends to improve as the maximum number of steps increases, highlighting the benefit of longer chains of multi-step reasoning. Detailed results and analysis for individual tasks are provided in \S\ref{app:exp_results}. 

\begin{figure}[th!]
    \centering
    \vspace{-1mm}
    \includegraphics[width=0.45\textwidth]{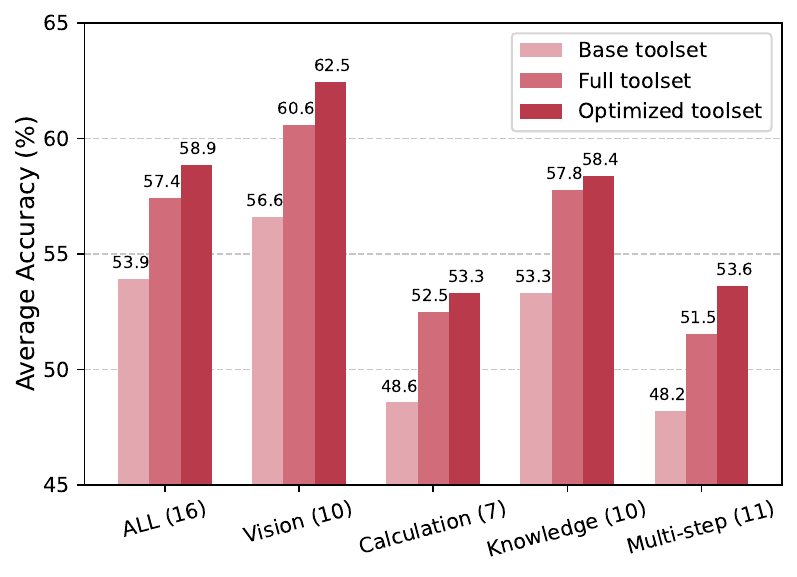}
    \vspace{-2mm}
    \caption{Performance of \model under three toolset strategies across different task categories (parentheses indicate the number of tasks per category).}
    \vspace{-1mm}
\label{fig:ours_tool_selection_category_scores_bar_chart}
\end{figure}

\paragraph{Toolset optimization.}
To investigate the benefits of our toolset optimization algorithm (presented in \S\ref{sec:toolset_optimization} and Algorithm~\ref{alg:tool_selection_optimization}), we evaluate \model under three toolset strategies: (1) \modelbase, i.e., the setup with only the base \texttt{Generalist\_Solution\_Generator} tool, (2) \model with the \textit{full} toolset, where all possible tools are enabled, and (3) \model with the \textit{optimized} toolset. Figure~\ref{fig:ours_tool_selection_category_scores_bar_chart} presents results across all 16 tasks and different reasoning categories.

On average, the optimized toolset achieves 58.9\% accuracy, outperforming the \modelbase (53.9\%) by 5.0\%. Meanwhile, simply enabling all tools yields 57.4\% accuracy, which still outperforms \modelbase by 3.5\%. Although toolset optimization provides the higher overall performance, our framework is sufficiently versatile that enabling all tools also offers a substantial improvement, helping ensure generalization as we scale up the toolset.

\begin{figure}[th!]
    \centering
    \vspace{-1mm}
    \includegraphics[width=0.45\textwidth]{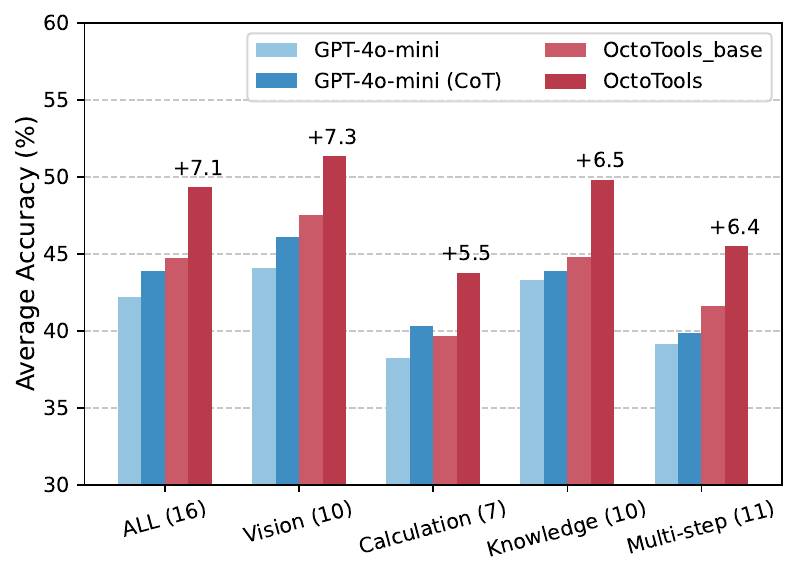}
    \vspace{-1mm}
    \caption{Performance of \model across task categories using a weaker LLM, \gptmini, as the backbone LLM.}
    \vspace{-1mm}
\label{fig:gpt4omini_ours_category_scores_bar_chart}
\end{figure}

\paragraph{Using a weaker LLM.}
Additionally, we evaluate our framework using a weaker LLM (\gptmini) as the base engine. Similar to the analysis with \gpt, we compare \model against the zero-shot and CoT baselines on the validation sets of 16 benchmarks. As shown in Figure~\ref{fig:gpt4omini_ours_category_scores_bar_chart}, \model generally outperforms the baselines across all 16 tasks and various categories. Task-specific analyses are provided in \S\ref{app:exp_results}.

\begin{figure}[th!]
    \centering
    \includegraphics[width=0.4\textwidth]{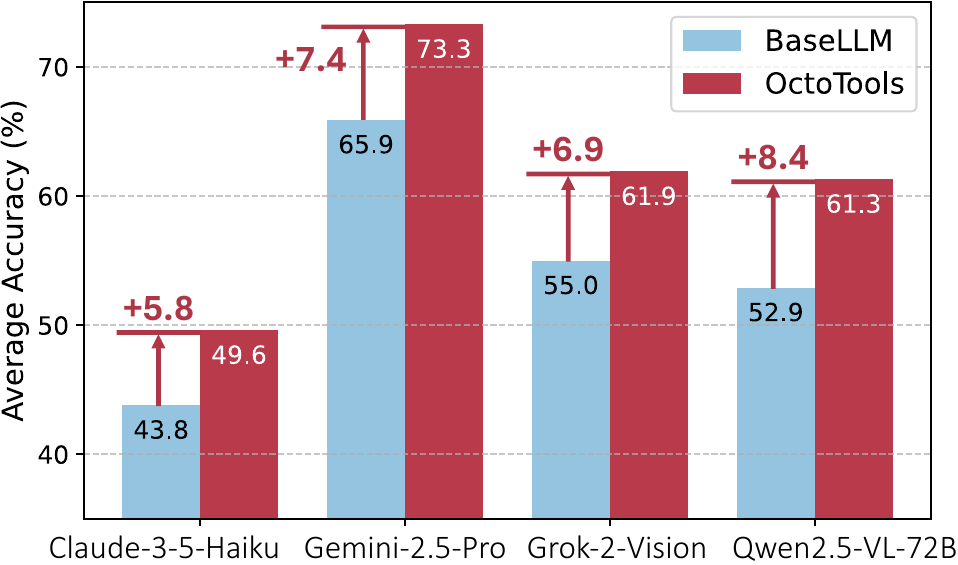}
    \vspace{-1mm}
    \caption{Performance gains of \model across different backbone LLMs, averaged over 16 benchmarks. }
    \vspace{-1mm}
\label{fig:main_generalization_bar_chart}
\end{figure}

\paragraph{Generalization to backbone LLMs.}
To highlight the generality of \model, we evaluated \model on four additional backbone LLMs: Claude 3.5 Haiku~\cite{anthropic2024claude35haiku}, Gemini 2.5 Pro~\cite{comanici2025gemini}, Grok-2 Vision~\cite{xai2024grok2vision}, and the open-source Qwen2.5-VL-72B~\cite{bai2025qwen2}. As shown in Figure~\ref{fig:main_generalization_bar_chart}, \model yields consistent accuracy gains of 5.8–8.4\% across all models, confirming that improvements stem from its planner-executor design and modular tool integration rather than any specific LLM.

\paragraph{Compact backbones and noisy tool environments.}
To probe capability boundaries more directly, we evaluate \model with compact open-source backbones on four representative tasks using Qwen2.5-3B and Qwen2.5-7B \citep{yang2024qwen2_5}, and stress-test it under noisy tool environments. As shown in Tables~\ref{tab:compact_backbones} and~\ref{tab:noisy_tools}, \model remains helpful even with small backbones, improving average accuracy by 13.6\% with Qwen2.5-3B and 6.3\% with Qwen2.5-7B. We further inject independent random tool failures during inference with the Qwen2.5-7B backbone, where each tool call returns an explicit error message with probability $p$. Performance degrades gracefully: up to $p=0.4$, the average drop on 10 tasks is at most 1.6\%, suggesting that the planner--executor loop can tolerate substantial tool noise.
\begin{table}[t]
    \centering
    \small
    \setlength{\tabcolsep}{4pt}
    \resizebox{1.0\linewidth}{!}{
    \begin{tabular}{llccccc}
        \toprule
        Backbone & Method & GAIA & G24 & GPQA & MedQA & Avg \\
        \midrule
        Qwen2.5-3B & Base LLM & 0.0 & 5.0 & 24.1 & 65.0 & 23.5 \\
        Qwen2.5-3B & \model   & 14.3 & 15.0 & 39.4 & 80.0 & \textbf{37.2} (+13.6) \\
        \midrule
        Qwen2.5-7B & Base LLM & 3.2 & 33.0 & 34.0 & 66.0 & 34.1 \\
        Qwen2.5-7B & \model   & 17.2 & 31.0 & 37.0 & 76.0 & \textbf{40.3} (+6.3) \\
        \bottomrule
    \end{tabular}
    }
    \vspace{-1mm}
    \caption{\model on compact open-source backbones across four representative tasks.}
    \vspace{-1mm}
    \label{tab:compact_backbones}
\end{table}

\begin{table}[t]
    \centering
    \small
    \setlength{\tabcolsep}{4pt}
    \begin{tabular}{lccccc}
        \toprule
        Setting & GAIA & GPQA & MedQA & Avg (10) & Drop \\
        \midrule
        No tool error & 17.2 & 37.0 & 76.0 & \textbf{41.4} & / \\
        $p=0.1$ & 22.8 & 30.0 & 67.0 & 39.4 & $-2.0$ \\
        $p=0.2$ & 22.1 & 31.0 & 66.0 & 40.8 & $-0.6$ \\
        $p=0.4$ & 19.7 & 28.0 & 66.7 & 39.8 & $-1.6$ \\
        $p=0.6$ & 14.2 & 22.0 & 63.0 & 35.3 & $-6.1$ \\
        \bottomrule
    \end{tabular}
    \vspace{-1mm}
    \caption{Robustness under injected tool failures with Qwen2.5-7B. With probability $p$, a tool call returns an error message instead of a valid result.}
    \vspace{-1mm}
    \label{tab:noisy_tools}
\end{table}

\section{In-depth Analysis}
\label{sec:in_depth_analysis}

\paragraph{Cost analysis.}
We measured per-query resource consumption of \model across 16 benchmarks with a 10-step limit. As shown in Table~\ref{tab:efficiency_cost}, GPT-4o costs about $0.05$ per query, while GPT-4o-mini reduces this to under $0.01$. Using open-source models like Qwen2.5-VL-72B on local hardware eliminates cost entirely. 
Notably, despite a maximum budget of 10 steps, the average number of executed steps is only 2.56 with GPT-4o and 3.49 with GPT-4o-mini, indicating that the verifier often stops early on simpler queries instead of always consuming the full budget. In large-scale deployments, users can further reduce cost by lowering the step budget, disabling expensive tools via task-specific toolset optimization, or routing queries to smaller backbones. These results show that \model offers a flexible trade-off between performance, latency, and cost depending on the chosen backbone.

% Table
\begin{table}[ht!]
\centering
\small
\renewcommand\tabcolsep{3pt}
\resizebox{1.0\linewidth}{!}{
\begin{tabular}{lccccc}
\toprule
\textbf{LLMs in \model} & \textbf{Cost} & \textbf{Avg. steps} & \textbf{Time (s)} & \textbf{Tokens} \\
\midrule
GPT-4o-mini & $0.01$ & 3.49 & 63.4 & 39.2K \\
GPT-4o & $0.05$ & 2.56 & 58.8 & 38.0K \\
Claude-3-5-Haiku & $0.04$ & 2.57 & 69.2 & 31.4K \\
Gemini-2.5-Pro & $0.15$ & 3.67 & 216.2 & -- \\
Grok-2-Vision & $0.13$ & 4.10 & 152.7 & 38.5K \\
Qwen2.5-VL-72B & $0.0$ & 2.29 & 111.9 & 27.7K \\
\bottomrule
\end{tabular}}
\vspace{-1mm}
\caption{Averaged resource usage per query across.}
\vspace{-2mm}
\label{tab:efficiency_cost}
\end{table}

\paragraph{Human study.}
To assess transparency and user experience, we launched an interactive online demo of \model that visualizes the full reasoning trajectory, including the initial plan, each tool call, and intermediate results. To date, the demo has attracted over 22.9K visits and processed 3.6K user queries. Of 100 sampled feedback entries, 69\% were positive upvotes, with users praising the system's interpretability. One user commented, \textit{``The step-by-step breakdown is incredibly helpful for understanding how the final answer was derived.''} This feedback suggests our transparent design enhances user experience and trust.

\paragraph{Failure case study.}
An analysis of 200 failure cases reveals that most errors stem from tool limitations and imperfect LLM planning, rather than the framework itself. As shown in Table~\ref{tab:failure_modes}, weak tools (65.0\%) and suboptimal planning (77.5\% combined) were the primary causes. In contrast, core framework errors like invalid command generation were rare (1.5\%), underscoring the robustness of our planner-executor design. This suggests that improving tool quality and the planner's reasoning are key directions for future work.

% Table: failure_modes
\begin{table}[h!]
\centering
\small
\resizebox{1.0\linewidth}{!}{%
\begin{tabular}{lc}
\toprule
\textbf{Failure Mode} & \textbf{Percentage (\%)}  \\
\midrule
Weak or incapable tools & 65.0 \\
Imperfect low-level plan & 45.0 \\
Imperfect high-level plan & 32.5\\
Tool execution errors & 18.0 \\
Incorrect solution summarization & 6.0 \\
Invalid tool command generation & 1.5 \\
\bottomrule
\end{tabular}%
}
\vspace{-1mm}
\caption{Distribution of primary failure modes.}
\label{tab:failure_modes}
\end{table}

\section{Conclusion}
\label{sec:conclusion}
In this paper, we introduced \model, a training-free, extensible multi-agent framework for complex reasoning. \model employs standardized \emph{tool cards} to facilitate seamless integration of diverse tools and a dedicated planner-executor workflow that separates high-level planning over multiple steps from low-level planning and command generation within each step. Through extensive experiments on 16 diverse benchmarks, \model consistently outperforms baselines, achieving average accuracy gains of up to 9.3\% over \gpt and up to 10.6\% over strong agentic frameworks. Our in-depth analysis shows that \model's improvements stem from dynamic task planning, effective tool usage, and multi-step problem decomposition.
Ultimately, \model provides a robust and practical foundation for building more capable, reliable, and transparent AI agents, paving the way for future work in automated tool discovery and advanced long-horizon planning.

% \clearpage
\section{Limitations}

\paragraph{Base LLM capabilities.}
As a training-free framework, \model's performance is inherently tied to the underlying LLM (e.g., GPT-4o, GPT-4o-mini). The quality of planning, command prediction, and summarization depends on the LLM's reasoning and code generation abilities. When the base LLM hallucinates or misinterprets tool metadata, downstream performance may degrade. \S\ref{sec:ablation_study} shows that \model remains robust with weaker models, but stronger or more specialized LLMs further mitigate this issue.

\paragraph{Tool quality and availability.}
The framework assumes that external tools are accurate and accessible. In restricted environments (e.g., no internet, strict API quotas, or deprecated services), tool benefits may diminish. \model's modular design allows tools to be replaced or upgraded, (\S\ref{sec:toolcards}) yet ensuring robustness to degraded or missing tools remains an open challenge.

\paragraph{Task-specific toolset optimization.}
Our greedy search strategy (\S\ref{sec:toolset_optimization}) is efficient but not globally optimal and relies on validation data. With scarce or biased validation examples, the chosen subset may be suboptimal. Current optimization is static per task; future work may explore dynamic, query-level adaptation.

\paragraph{Computational cost and scalability.}
Multi-step planning improves accuracy but adds latency and costs that grow with the step budget and number of tools (\S\ref{sec:ablation_study}, \S\ref{sec:in_depth_analysis}). Some queries can exceed 90 seconds, and long trajectories increase the memory footprint. Large-scale or latency-sensitive applications may require batching, pruning, or model distillation.

\paragraph{Fairness, privacy, and security.}
By calling APIs or running generated code, \model may expose user data to third-party services or execute unsafe code. Although sandboxed execution is possible, tools like the \texttt{Python\_Code\_Generator} (\S\ref{sec:agent_execution}) still carry inherent risks. External tools (e.g., web search) may also propagate biases. Addressing these concerns requires systematic auditing beyond the present study.

\paragraph{Reproducibility and environment drift.}
Some tools rely on third-party services whose interfaces, quotas, or model versions may evolve. While we release our code and tool cards, exact reproduction requires snapshotting tool versions or containerized environments, which we view as necessary future work.

\paragraph{Ethical deployment.}
Finally, \model's modularity lowers barriers to building agentic systems but also raises risks of misuse, such as large-scale disinformation generation. Governance and safeguards are critical, though outside the scope of this work.

\section*{Broader Impacts and Ethical Considerations}

\model enhances LLMs with extensible tools for complex reasoning, holding significant promise for advancing research and development. Applications include accelerating scientific discovery through integrated literature search, database access, and computational tools, and improving efficiency in healthcare by assisting in clinical report generation. Its operation also produces valuable data, including detailed trajectories of planning, actions, and tool calls, that can be leveraged to fine-tune more capable and aligned AI agents.

At the same time, these capabilities present risks. Automated complex reasoning and tool use could be misused for large-scale disinformation, creation of deceptive profiles, or unauthorized surveillance through the combination of web-access and analytic tools. Privacy concerns may arise if sensitive user data or external services are accessed without safeguards. In addition, biases from the base LLM or external tools (e.g., search engines) may be propagated or amplified, potentially resulting in unfair or discriminatory outcomes. Harms could stem from malicious use, unintended consequences of otherwise correct tool use, or incorrect outputs in critical domains.

Mitigation strategies are essential. We plan to release the code with clear guidelines on sandboxing executable tools and emphasize the need for human oversight. The modular tool card design facilitates auditing and replacement of biased or insecure tools. Moreover, the generated reasoning trajectories can support research on AI safety, interpretability, and robustness. We encourage the community to establish best practices for responsible deployment, including gated access to sensitive tools and active monitoring for misuse.

Finally, in preparing this manuscript, we used LLMs solely for writing assistance, limited to phrasing, clarity, and grammar. All outputs from the LLM were critically assessed, revised, and approved by the authors to maintain accuracy and integrity. Importantly, the LLMs were not involved in formulating research ideas, designing experiments, analyzing data, or generating the core scientific arguments. 

\section*{Acknowledgments}
\label{sec:ack}
This work was partially supported by the Hoffman--Yee Research Grants Program at Stanford HAI, the AI for Math Fund by Renaissance Philanthrop, and the Chan Zuckerberg Initiative. We thank members of the Zou group for helpful discussions and feedback.

\section*{Contribution Statement}
\label{sec:contribution}
PL and RT initiated the project. PL developed the early framework. PL and BC refined the framework. PL, BC, and SL contributed to the experiments and paper writing. Correspondence to: \href{mailto:panlu@stanford.edu}{panlu@stanford.edu}.

%%%%%%%%%%%%%%%%%%%%%%%%%%%%%%%%%%%%%%%%%%%%%%%%%%%%%%%%%%%%%%%%%%%%%%%%%%%%%%%
%%%%%%%%%%%%%%%%%%%%%%%%%%%%%%%%%%%%%%%%%%%%%%%%%%%%%%%%%%%%%%%%%%%%%%%%%%%%%%%
% APPENDIX
%%%%%%%%%%%%%%%%%%%%%%%%%%%%%%%%%%%%%%%%%%%%%%%%%%%%%%%%%%%%%%%%%%%%%%%%%%%%%%%
%%%%%%%%%%%%%%%%%%%%%%%%%%%%%%%%%%%%%%%%%%%%%%%%%%%%%%%%%%%%%%%%%%%%%%%%%%%%%%%
% Bibliography entries for the entire Anthology, followed by custom entries
%\bibliography{anthology,custom}
% Custom bibliography entries only
\bibliography{custom}

\appendix
\clearpage

\onecolumn
%%%%%%%%%%%%%%%%%%%%%%%%%%%%%%%%%%%%%%%%
\renewcommand*\footnoterule{} % remove the separator line

\newcommand\blfootnote[1]{%
  \begingroup
  \renewcommand\thefootnote{}\footnote{#1}%
  \addtocounter{footnote}{-1}%
  \endgroup
}
%%%%%%%%%%%%%%%%%%%%%%%%%%%%%%%%%%%%%%%%%

%%%%%%%%%%%%%%%%%%%%%%%%%%%%%%%%%%%%%%%%%
% Create a new ToC for appendix only
\section*{Table of Contents}
\setcounter{tocdepth}{2}
\renewcommand{\contentsname}{Appendix Contents}
\startcontents[appendix]  % Requires the 'titletoc' package
\printcontents[appendix]{}{1}{}
%%%%%%%%%%%%%%%%%%%%%%%%%%%%%%%%%%%%%%%%%

\twocolumn

\section{Related Work}
\label{sec:related}

\paragraph{Tool-Augmented LLMs.}
A promising direction for enhancing large language models (LLMs) involves offloading specialized subtasks to external tools such as search engines \citep{komeili-etal-2022-internet, thoppilan2022lamda, lazaridou2022internet, shuster2022blenderbot, yao2022react}, web browsers \citep{nakano2021webgpt}, calculators \citep{cobbe2021training, thoppilan2022lamda}, translation systems \citep{thoppilan2022lamda}, or Python interpreters \citep{gao2023pal}. 
Early approaches relied on large-scale fine-tuning to teach LLMs how to invoke tools \citep{schick2023toolformer, thoppilan2022lamda}. More recently, reinforcement learning (RL) from outcome-based rewards has emerged as a popular paradigm for training monolithic policies to use tools \citep{qian2025toolrl, zhang2025nemotron}. However, these training-based methods face significant challenges in long-horizon credit assignment, often leading to training instability and brittle generalization \citep{zeng2025reinforcing, wang2025stepsearch}.
In contrast, \model is a \emph{training-free} framework that bypasses these complexities by integrating diverse tools through standardized \emph{tool cards} and a deterministic planner-executor paradigm, offering a more \emph{extensible} and \emph{modular} approach.

\paragraph{LLM Agents and Multi-Agent Systems} A growing body of work leverages LLMs as autonomous agents. These can be broadly categorized into single-agent and multi-agent systems. Many specialized single-agent frameworks have shown strong performance in narrow domains like chemistry \citep{bran2023chemcrow} or vision \citep{li2024mmedagent, hu2024visual}, but often lack generality. Concurrently, \emph{multi-agent systems} like AutoGen \citep{autogen} and MetaGPT \citep{hong2024metagpt} have gained prominence by decomposing complex tasks across several collaborating agents. While many of these frameworks initially rely on handcrafted logic and prompting, their static design can limit adaptability. To address this, recent research explores methods for dynamically optimizing agent coordination. For instance, some approaches refine collaborative strategies by leveraging static datasets of successful interactions or preferences \citep{deng2025pe, liao2025marft}. However, enabling agents to adapt their strategies in real-time based on the unfolding interaction remains a significant challenge, especially when feedback is sparse or delayed \citep{wang2025ragen}. \model contributes a distinct multi-agent paradigm built on a highly structured planner-executor architecture. Unlike systems with more flexible, peer-to-peer conversational models, \model enforces a clear division of labor: a planner agent handles high-level strategy and decomposition, while an executor agent translates those plans into concrete actions. This fixed, hierarchical interaction simplifies coordination, enhances transparency, and ensures reliability.

\paragraph{Complex Task Reasoning.}
When faced with multi-step problems, a common strategy is to break down a question into simpler sub-questions and solve them step by step. Early work approached decomposition with unsupervised or weakly supervised models \citep{perez2020unsupervised, khot2022decomposed}, and more recent research has explored prompting techniques for step-by-step reasoning, including Chain-of-Thought \citep{wei2022chain}, Least-to-Most \citep{zhou2022least}, ReAct \citep{yao2022react}, Pearl \citep{sun2023pearl}, Forest-of-Thought \citep{bi2024forest}, and rStar-Math\citep{guan2025rstar}. While these methods significantly improve LLMs’ single-model reasoning capabilities, they primarily rely on the latent capacity of an LLM without external validation or targeted tool usage. In contrast, \model systematically combines multi-step decomposition (via an iterative planner) with specialized tools (encapsulated by \emph{tool cards}) and an executor for reliable, context-aware function calling. This design makes it easy to incorporate domain-specific functionalities and check intermediate steps with external modules, thus improving both correctness and versatility in tackling challenging tasks.

\section{Experimental Details}
\label{app:exp_details}

\subsection{Benchmark Datasets}
\label{app:benchmark_datasets}
Here, we report further details of each of the 16 benchmark we used in this study. Unless specified otherwise, a validation set of 100 examples and a test set of 200 examples were sampled from each dataset. No additional preprocessing was performed for open-ended questions. For multiple choice questions, choices were enumerated using capital letters and each concatenated to the question following a new line character.

\subsubsection{General Domain Benchmarks}

\textbf{AlgoPuzzleVQA}~\cite{ghosal2024language} is a collection of geometric puzzles that require visual perception, language understanding, and complex algorithmic reasoning—tasks that are challenging for base VLMs. 

\textbf{Hallusion-VD} is a subset of HallusionBench~\cite{guan2024hallusionbench}, a benchmark designed to evaluate visual understanding through optical and geometric illusions. For our experiments, we use the Visual Dependent subset, which consists of questions that require the visual context to determine a definitive answer.

\textbf{PuzzleVQA}~\cite{chia2024puzzlevqa} is a dataset of puzzle instances based on abstract geometric patterns, testing reasoning and understanding across attributes such as color, number, size, and shape.

\textbf{VQA 2.0}~\cite{goyal2017making} contains open-ended questions about images that require fine-grained visual understanding.

\subsubsection{Mathematical Benchmarks}

\textbf{Game of 24}~\cite{24game} is based on the classic arithmetic puzzle of 24 (also known as the 24 numbers game). The task involves using four given numbers and basic arithmetic operations (addition, subtraction, multiplication, division) to construct an expression that evaluates to 24. For example, given the numbers 4, 9, 10, and 13, one valid solution is ``(10 - 4) × (13 - 9) = 24.'' Solving the puzzle requires numerical calculation skills as well as iterative attempts to verify proposed solutions. Each puzzle is presented as an open-ended question.

\textbf{Omni-MATH}~\cite{gao2024omni} is a text-only mathematical benchmark consisting of open-ended, competition-level problems at the Olympiad level, requiring advanced mathematical knowledge and reasoning.

\textbf{CLEVR-Math}~\cite{lindstrom2022clevr} is a dataset of multimodal math word problems involving addition and subtraction. Each problem combines a textual description with an accompanying image illustrating the scenario. Solving these problems requires a blend of linguistic, visual, and mathematical reasoning.

\textbf{MathVista}~\cite{lu2024mathvista} is a benchmark designed to integrate challenges from diverse mathematical and visual tasks. The queries include both multiple-choice and open-ended questions, requiring numerical computation, fine-grained visual understanding, and multi-step reasoning.

\subsubsection{Scientific Benchmarks}

\textbf{GPQA} (Graduate-Level Google-Proof Q\&A Benchmark)~\cite{rein2023gpqa} is a collection of challenging, text-only multiple-choice questions written by domain experts in biology, physics, and chemistry, designed to be ``extremely difficult.''

\textbf{MMLU-Pro}~\cite{wang2024mmlu} is a text-only benchmark consisting of reasoning-focused multiple-choice questions that require both general scientific knowledge and complex reasoning.

\textbf{SciFIBench}~\cite{roberts2024scifibench} is a benchmark of multiple-choice questions for scientific figure interpretation. Queries require understanding and extracting information from scientific figures.

\subsubsection{Medical Benchmarks}

\textbf{MedQA}~\cite{jin2021disease} consists of text-only multiple-choice questions curated from professional medical board exams. The questions cover general medical knowledge, clinical reasoning, and diagnosis.

\textbf{PathCLS}, a subset of PathMMU~\cite{sun2025pathmmu}, consists of multiple-choice questions reformulated from well-known pathology classification datasets. Each question is based on hematoxylin and eosin (H\&E)-stained histopathology images and typically involves disease diagnosis.

\textbf{PathVQA}~\cite{he2020pathvqa} is a visual question-answering dataset curated from pathology-related image–caption pairs sourced from textbooks, spanning multiple tissue types and stains. All questions were treated as open-ended in our evaluation. 

\textbf{SLAKE} (Semantically-Labeled Knowledge-Enhanced Dataset)~\cite{liu2021slake} is a radiology visual question-answering dataset. The associated images include X-rays, computed tomography (CT) scans, and magnetic resonance imaging (MRI). All questions were treated as open-ended in our evaluation. Although the dataset also provides object detection labels and segmentation masks for each image, we excluded these from our evaluation to increase difficulty.

\subsubsection{Agentic Benchmark}

\textbf{GAIA-Text}~\cite{mialon2023gaia} is a benchmark specifically designed to evaluate general AI assistants and agents, requiring abilities such as multi-step reasoning, web browsing, and generally tool-use proficiency. The questions are designed to be difficult for base LLMs. We use the text-only subset of this dataset in our experiments.

\subsection{Tools Used in Our Experiments}
\label{app:tool_descriptions}

We implemented 11 tools in the toolbox for our experiments. Here, we provide detailed descriptions of each tool. See \S\ref{app:toolcards} for the complete tool cards of each tool and usage examples. 

\textbf{ArXiv Paper Searcher} (\S\ref{app:arxiv_paper_searcher_tool}) searches arXiv\footnote{\url{https://arxiv.org/}}, an open-access pre-print repository, for abstracts and links that match a given query.

\textbf{Generalist Solution Generator} (\S\ref{app:generalist_solution_generator_tool}) is an instance of the \model base LLM and acts as the default reasoning engine if the agent decides not to use an external tool.

\textbf{Google Search} (\S\ref{app:google_search_tool}) uses the Google custom search API\footnote{\url{https://developers.google.com/custom-search/v1/introduction}} to search the web and return links and a summary of each result. 

\textbf{Image Captioner} (\S\ref{app:image_captioner_tool}) is an instance of the base LLM prompted for generating text descriptions of input images.

\textbf{Path Generalist Classifier} (\S\ref{app:path_generalist_classifier_tool}) is a tool for performing general classification of H\&E-stained pathology microscopy images. The tool relies on CONCH~\cite{lu2024avisionlanguage}, a pretrained vision-language foundation model for pathology, for performing zero-shot classification of pathology image patches.

\textbf{Pubmed Search} (\S\ref{app:pubmed_search_tool}) retrieves relevant article abstracts from PubMed based on a text query. The retrieval is performed using the PubMed and NCBI APIs\footnote{\url{https://www.ncbi.nlm.nih.gov/books/NBK25501/}}. 

\textbf{Python Code Generator} (\S\ref{app:python_code_generator_tool}) generates and executes Python code given a query and returns the execution result. The code generation is performed by an instance of the base LLM prompted for Python code generation. 

\textbf{Relevant Patch Zoomer} (\S\ref{app:relevant_patch_zoomer_tool}) is an instance of the base LLM that, given a query, decides which regions of the image to zoom into (among the four quadrants and the center patch) and saves the zoomed patches.

\textbf{Text Detector} (\S\ref{app:text_detector_tool}) detects multilingual text within an image by calling the EasyOCR tool for text detection\footnote{\url{https://github.com/JaidedAI/EasyOCR}}.

\textbf{URL Text Extractor} (\S\ref{app:url_text_extractor_tool}) visits web pages given the URL and returns the text content of the page.

\textbf{Wikipedia Knowledge Searcher} (\S\ref{app:wikipedia_knowledge_searcher_tool}) searches Wikipedia using the MediaWiki API\footnote{\url{https://www.mediawiki.org/wiki/API}} and returns articles matching a given query.

\subsection{Additional Tools for Exploration Study}

We also provide several additional tools for exploration, as follows:

\textbf{Object Detector} (\S\ref{app:object_detector_tool}) performs object detection on an image given a list of object labels to detect, using the Grounding DINO model \cite{caron2021emerging}. Due to the standardized design of tool cards, this tool can be upgraded to the \textbf{Advanced Object Detector} (\S\ref{app:advanced_object_detector_tool}), which uses DINO-X \cite{ren2024dinoxunifiedvisionmodel}, a more recent version powered by API calls.\footnote{\url{https://github.com/IDEA-Research/DINO-X-API}}

\textbf{Nature News Fetcher} (\S\ref{app:nature_news_fetcher_tool}) retrieves the latest news articles from the science journal \textit{Nature}.\footnote{\url{https://www.nature.com/latest-news}} An example in \S\ref{app:example_exploration} demonstrates how this tool can be used to obtain the latest research trends for a given topic.

% \clearpage
\section{More Experimental Results}
\label{app:exp_results}

\paragraph{Optimized tool sets.}  Table \ref{tab:optimized_toolsets} shows the optimized tool sets across 16 tasks from Algorithm \ref{alg:tool_selection_optimization}. The toolset optimization method in \model is able to find diverse optimal tool sets for different tasks. In general, the \texttt{Image\_Captioner} and \texttt{Relevant\_Patch\_Zoomer} tools are very commonly used in vision benchmarks, with the former being used in all vision benchmarks and the latter being used in 6 out of the 10. The \texttt{Python\_Code\_Generator} is represented in 3 out of the 4 mathematical domain benchmarks. The \texttt{Wikipedia\_Knowledge\_Searcher} is represented in all the scientific domain benchmarks. We also see that highly domain-specific tools are represented in their corresponding use cases, such as \texttt{Pubmed\_Search} in a general medical benchmark \cite{jin2021disease} and \texttt{Path\_Generalist\_Classifier} in a pathology classification benchmark \cite{sun2025pathmmu}. 

\begin{figure*}[th!]
    \centering
    \includegraphics[width=0.95\textwidth]{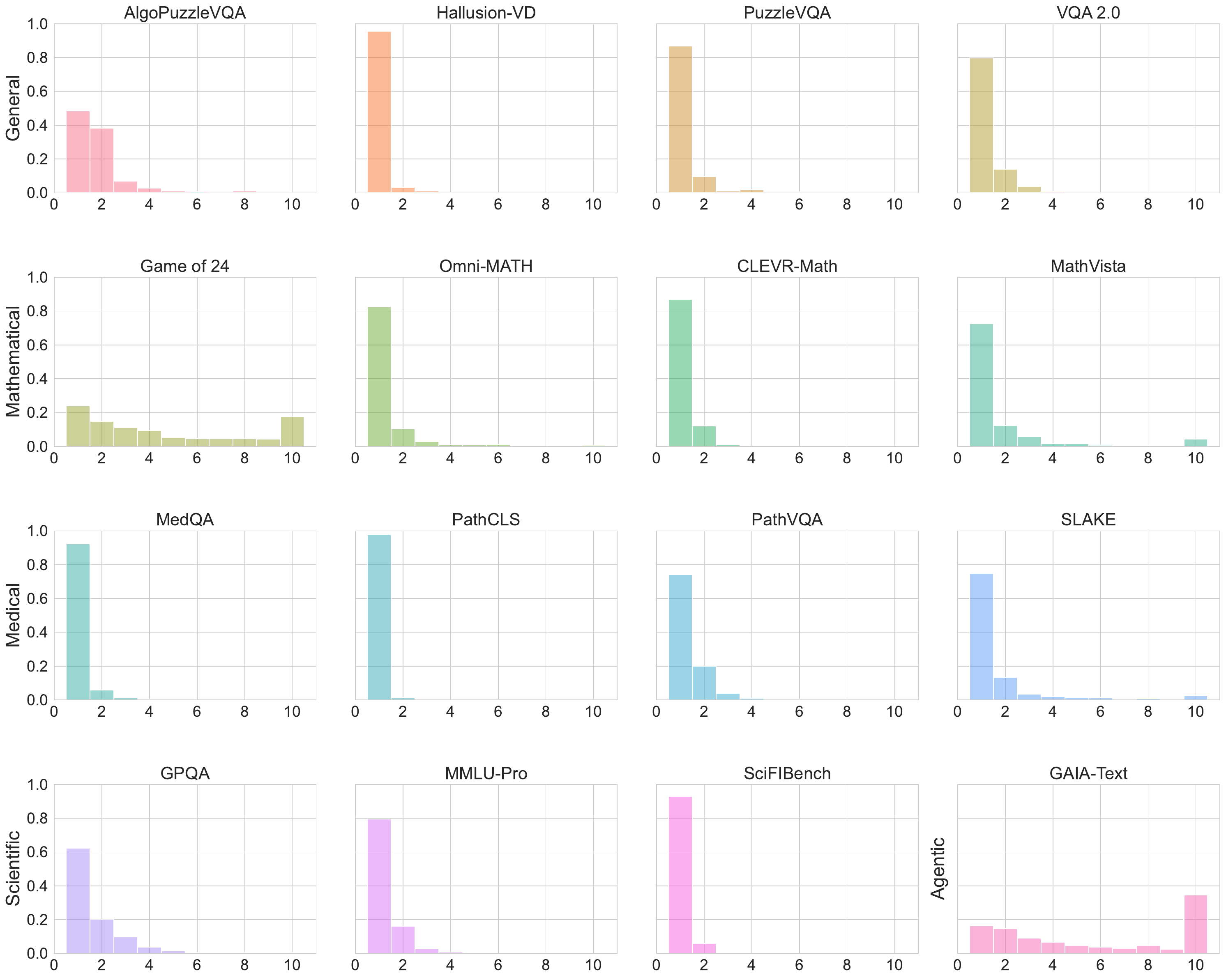}
    \caption{\textbf{Distribution of number of steps used.}}
\label{fig:distribution_of_number_of_steps}
\end{figure*}

\paragraph{Steps taken for different tasks.}
Figure~\ref{fig:distribution_of_number_of_steps} shows the distribution of the number of steps taken. \model is able to adapt to each task by applying different sets of tools and constructing chains of reasoning as needed.

\onecolumn

% Table: optimized_toolsets
\begin{table}[bh!]
    \centering
    \small
    \begin{tabular}{l|c|c|c|c|c|c|c|c|c|c|c|c|c}
        \toprule
        \textbf{Benchmarks} & \textbf{Modality} & \textbf{Domain} & 
        \rotatebox{90}{\base~\texttt{Generalist\_Solution\_Generator}} &
        \rotatebox{90}{\image~\texttt{Image\_Captioner}} & 
        \rotatebox{90}{\image~\texttt{Relevant\_Patch\_Zoomer}} & 
        \rotatebox{90}{\image~\texttt{Text\_Detector}} & 
        \rotatebox{90}{\www~\texttt{Wikipedia\_Knowledge\_Searcher}} & 
        \rotatebox{90}{\www~\texttt{Google\_Search}} & 
        \rotatebox{90}{\www~\texttt{URL\_Text\_Txtractor}} & 
        \rotatebox{90}{\www~\texttt{ArXiv\_Paper\_Searcher}} & 
        \rotatebox{90}{\code~\texttt{Python\_Code\_Generator}} & 
        \rotatebox{90}{\med~\texttt{Path\_Generalist\_Classifier}} & 
        \rotatebox{90}{\med~\texttt{Pubmed\_Search}} \\
        \midrule
        AlgoPuzzleVQA & Vision & General & \cmark & \cmark & & \cmark & & & & & & & \\
        Hallusion-VD & Vision & General & \cmark & \cmark & & & & & & & & & \\
        PuzzleVQA & Vision & General & \cmark & \cmark & & & & & & & & & \\
        VQA 2.0 & Vision & General & \cmark & \cmark & \cmark & & & & & & & & \\
        \midrule    
        Game of 24 & Text & Mathematical & \cmark & & & & & & & & \cmark & & \\
        Omni-MATH & Text & Mathematical & \cmark & & & & & & & & \cmark & & \\
        CLEVR-Math & Vision & Mathematical & \cmark & \cmark & \cmark & & & & & & & & \\
        MathVista & Vision & Mathematical & \cmark & \cmark & \cmark & & & \cmark & & & \cmark & & \\
        \midrule
        GPQA & Text & Scientific & \cmark & & & & \cmark & & & & & & \\
        MMLU-Pro & Text & Scientific & \cmark & & & & \cmark & & & & & & \\
        SciFIBench & Vision & Scientific & \cmark & \cmark & & \cmark & \cmark & & & \cmark & & & \\
        \midrule
        MedQA & Text & Medical & \cmark & & & & & & & & & & \cmark \\
        PathCLS & Vision & Medical & \cmark & \cmark & \cmark & & & & & & & \cmark & \\
        PathVQA & Vision & Medical & \cmark & \cmark & \cmark & & & & & & & & \\
        SLAKE & Vision & Medical & \cmark & \cmark & \cmark & & & & & & & & \\
        \midrule
        GAIA-Text & Text & Agentic & \cmark & & & & \cmark & \cmark & \cmark & & \cmark & & \\
        \bottomrule
    \end{tabular}
    \caption{Optimized tool sets for each benchmark following our Algorithm\ref{alg:tool_selection_optimization}. A \cmark indicates that the tool is used for that benchmark. }
    \vspace{-2mm}
    \label{tab:optimized_toolsets}
\end{table}

\begin{figure*}[th!]
    \centering
    \includegraphics[width=0.95\textwidth]{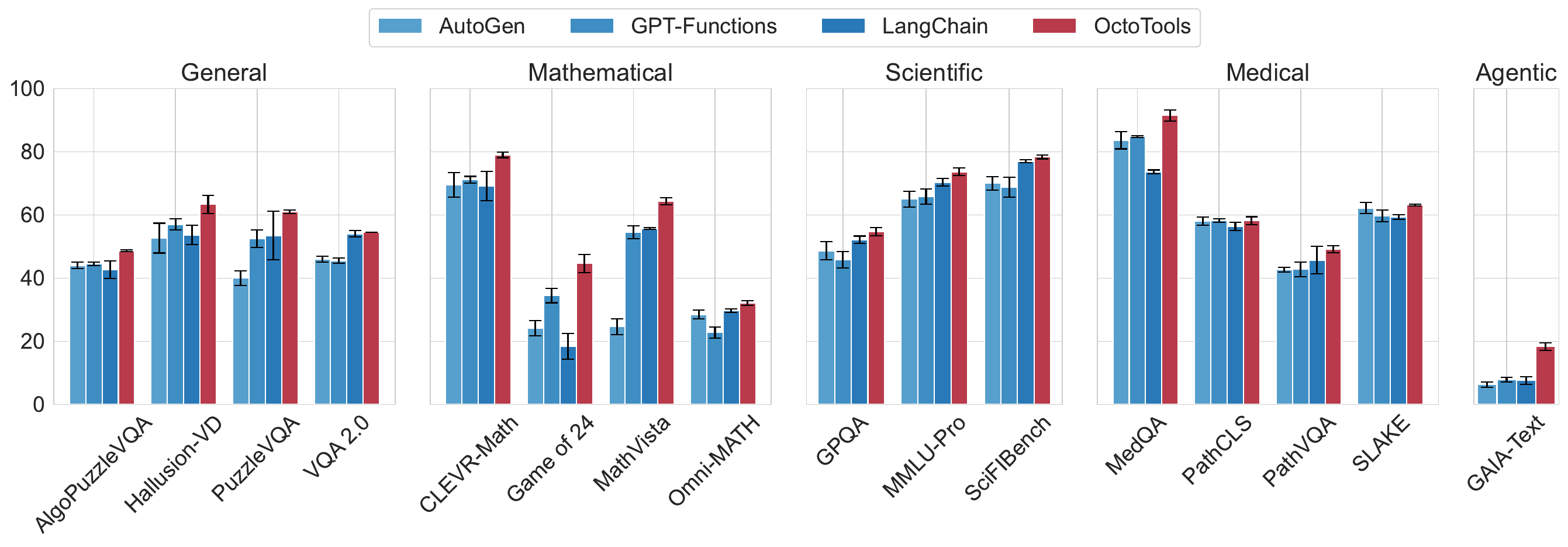}
    \caption{\textbf{Performance ours vs. other agents}. Our system consistently outperforms agent baselines across all benchmarks. Bar values represent accuracy and error bars represent standard deviation.}
\label{fig:agent_baselines_bar_plot}
\end{figure*}

\begin{figure*}[th!]
    \centering
    \includegraphics[width=0.95\textwidth]{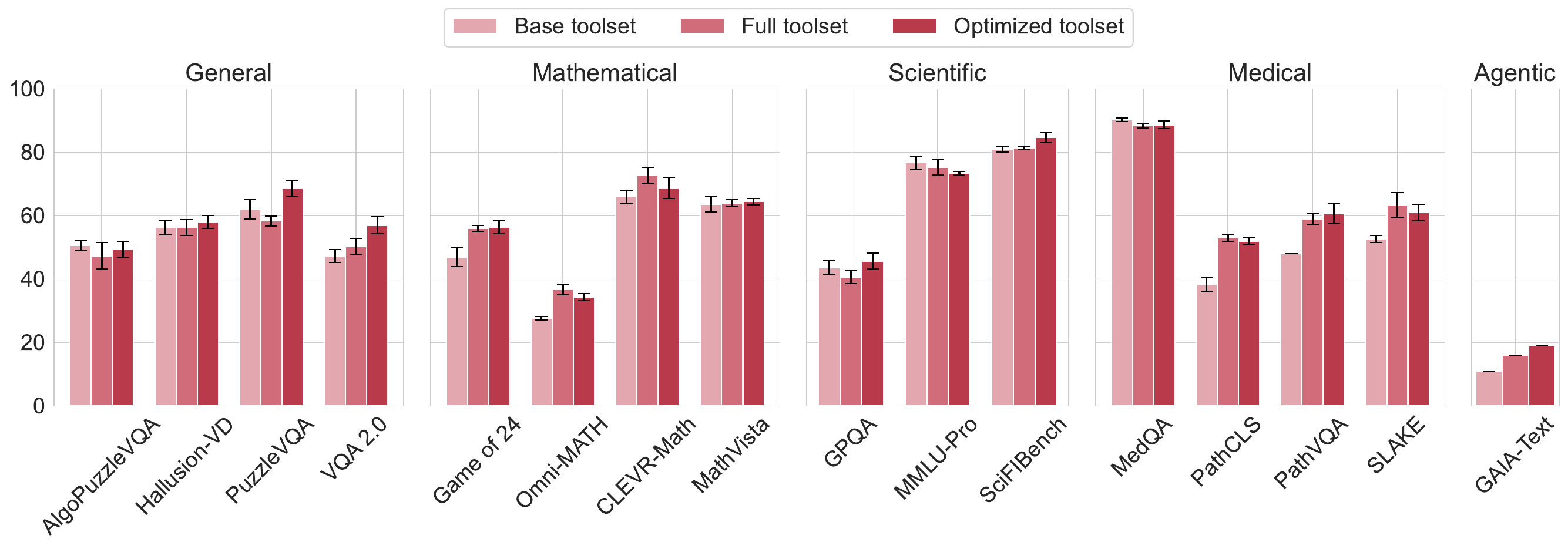}
    \vspace{-2mm}
    \caption{\textbf{Performance with vs. without tool selection}. While toolset optimization increases performance over using the full toolset in most tasks, even without it, our system achieves similar performance by naively enabling all possible tools. Bar values represent accuracy and error bars represent standard deviation.}
\label{fig:all_tools_ablation_bar_plot}
\end{figure*}

\begin{figure*}[th!]
    \centering
    \includegraphics[width=0.95\textwidth]{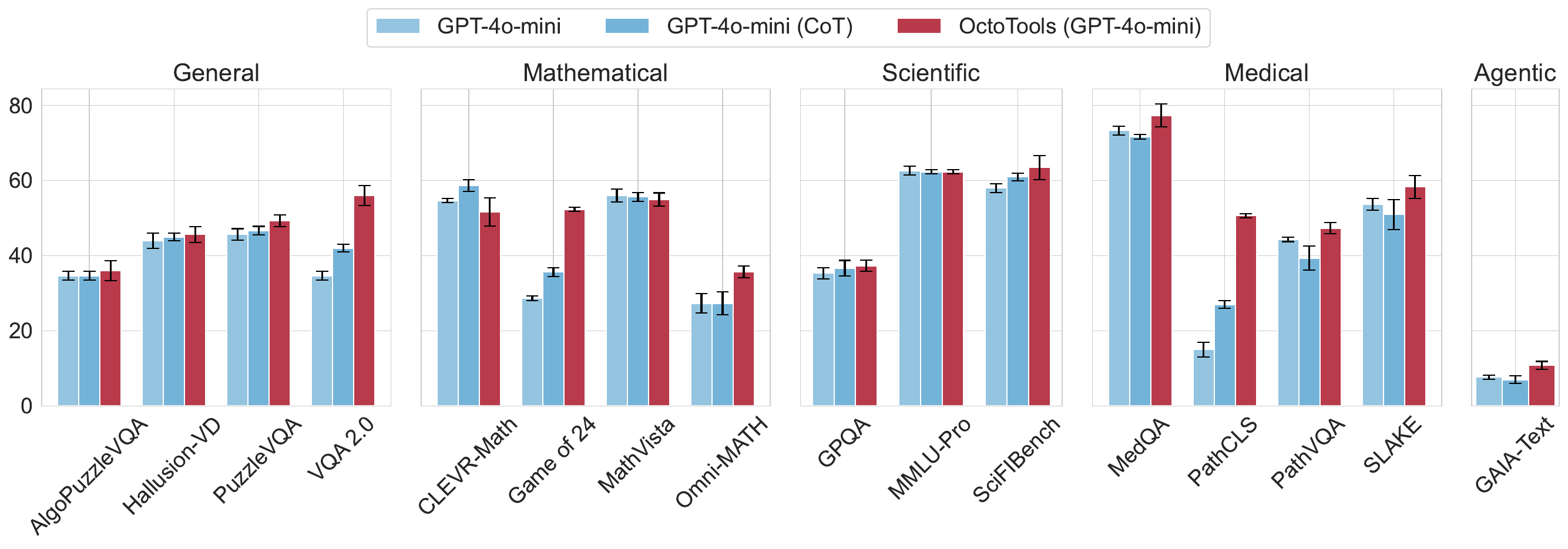}
    \vspace{-2mm}
    \caption{\textbf{Performance on a weaker LLM (\gptmini)}. We observe similar trends using \model with a weaker base LLM. Bar values represent accuracy and error bars represent standard deviation.}
\label{fig:weaker_llm_performance}
\end{figure*}

\begin{figure*}[th!]
    \centering
    \includegraphics[width=0.8\textwidth]{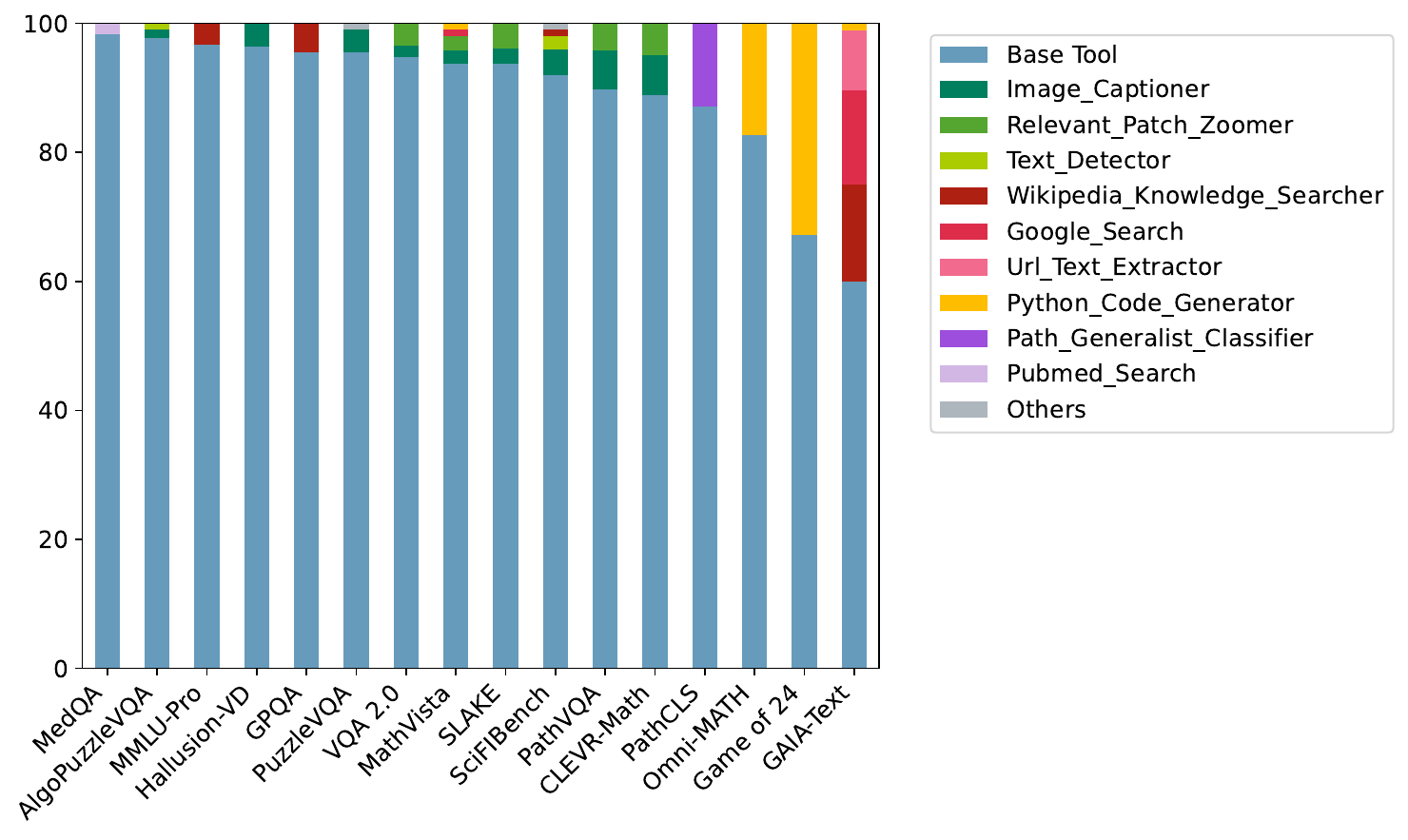}
    \vspace{-2mm}
    \caption{\textbf{Distribution of tools usage.} Frequency of tools used by the \autogen agent for each benchmark.
}
\label{fig:tool_usage_16_tasks_AutoGen}
\end{figure*}

\begin{figure*}[th!]
    \centering
    \includegraphics[width=0.8\textwidth]{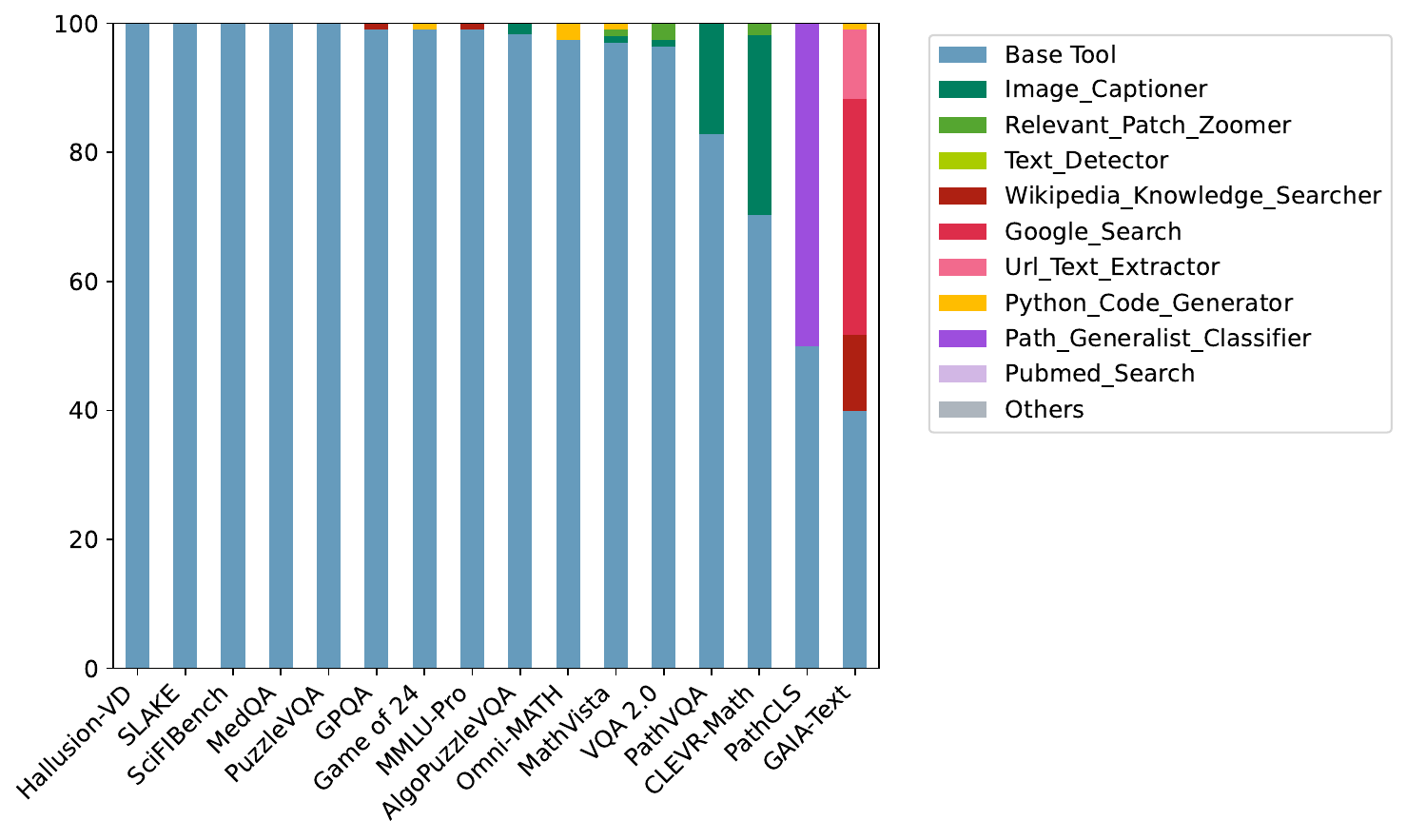}
    \caption{\textbf{Distribution of tools usage.} Frequency of tools used by the \langchain agent for each benchmark.
}
\label{fig:tool_usage_16_tasks_LangChain}
\end{figure*}

\begin{figure*}[th!]
    \centering
    \includegraphics[width=0.8\textwidth]{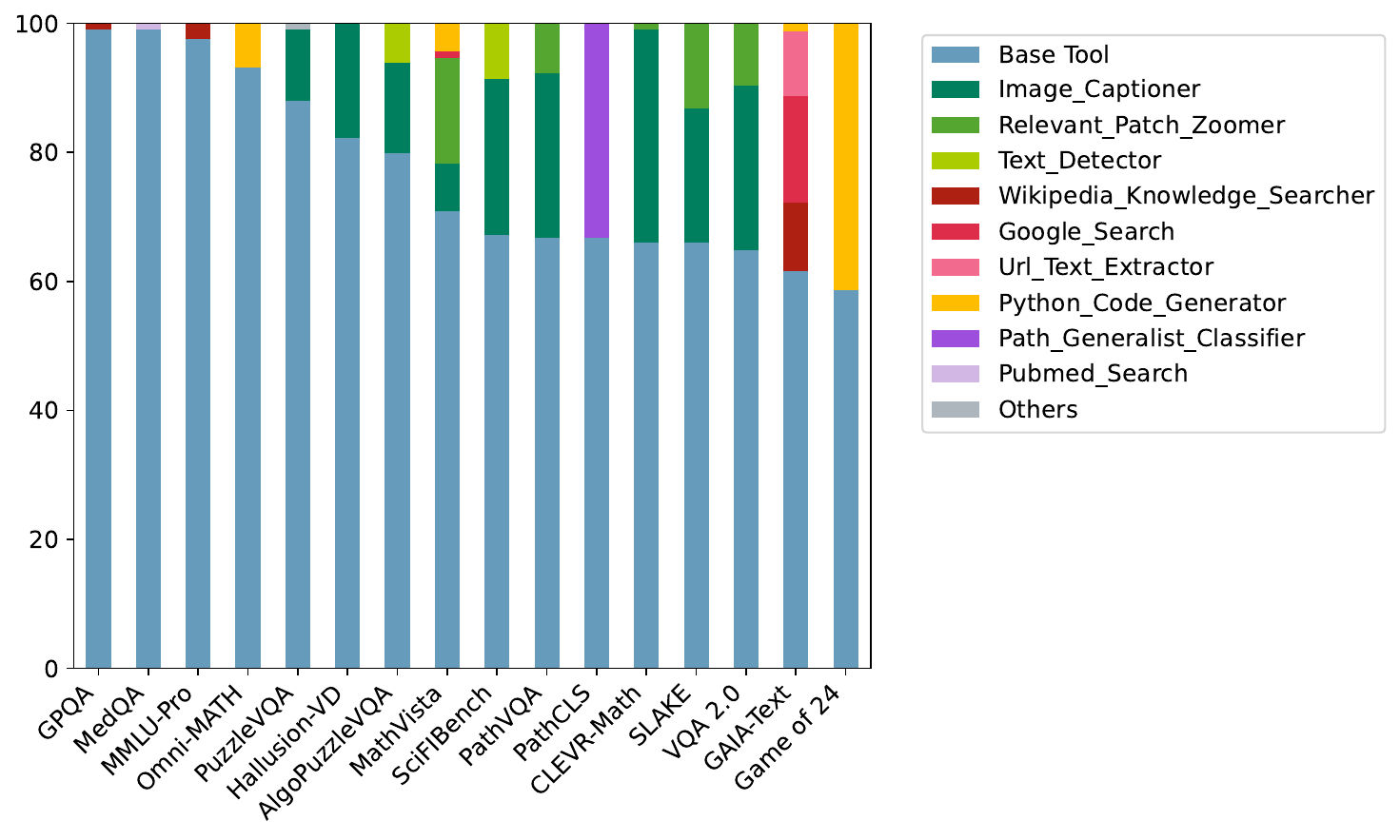}
    \caption{\textbf{Distribution of tools usage.} Frequency of tools used by the \gptplugin agent for each benchmark.
}
\label{fig:tool_usage_16_tasks_GPT4o-Plugin}
\end{figure*}

\begin{figure}[th!]
    \centering
    \includegraphics[width=0.45\textwidth]{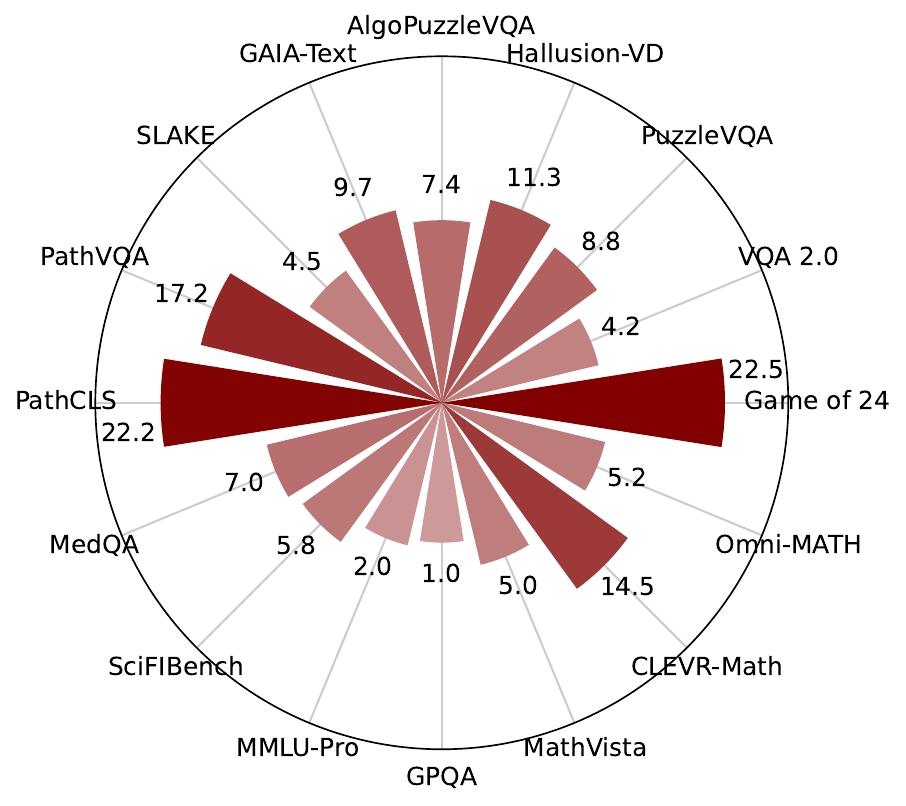}
    \caption{Performance gains across different benchmarks from our \model framework over the base \gpt model.}
\label{fig:benchmark_gains_radar_barplot}
\end{figure}

\begin{figure*}[th!]
    \centering
    \includegraphics[width=0.99\textwidth]{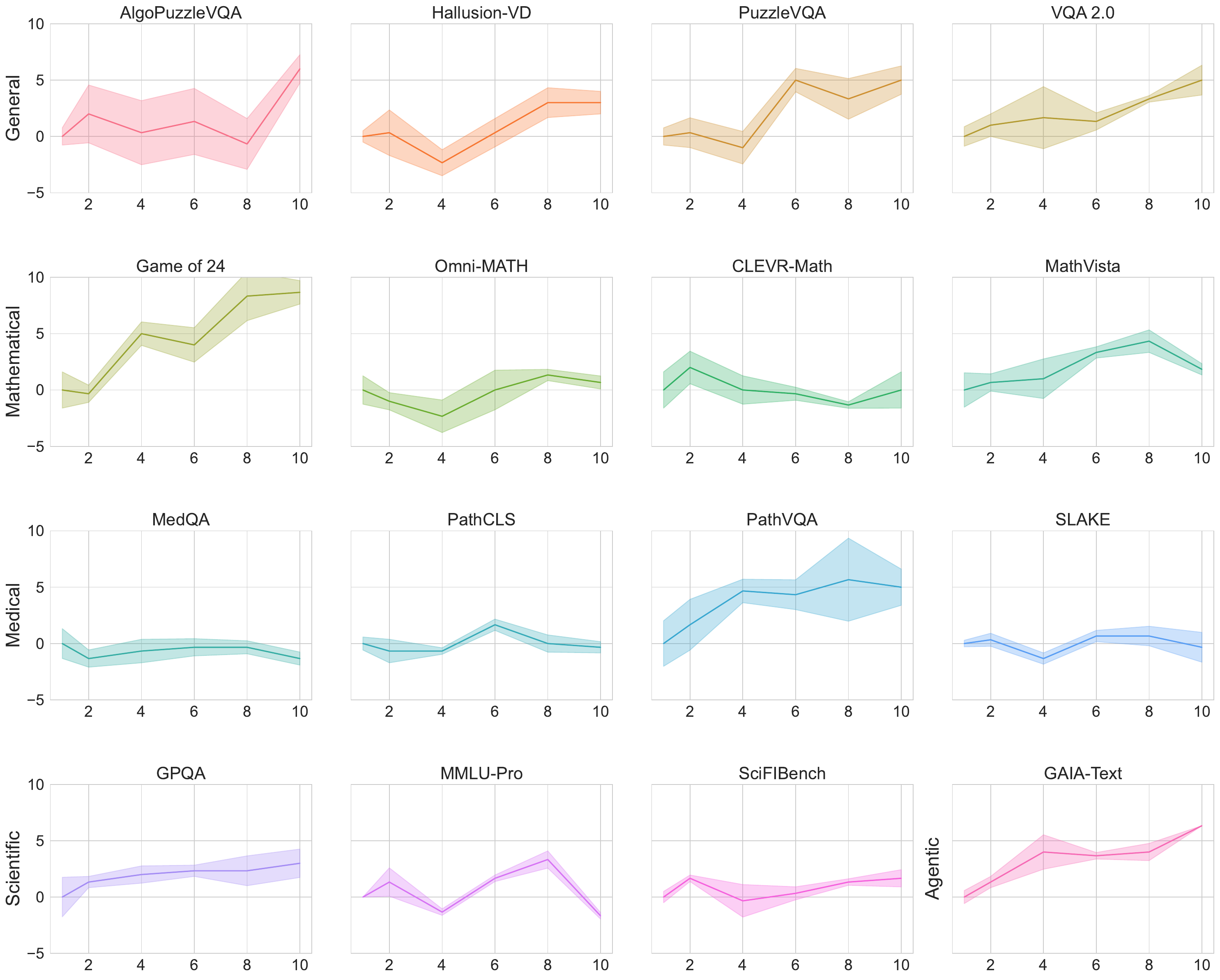}
    \caption{\textbf{Accuracy vs number of maximum steps.} The change in accuracy from a maximum step of 1 is plotted. Most benchmarks improve in performance with the number of allowed steps.}
\label{fig:accuracy_vs_max_steps}
\end{figure*}

\clearpage
\section{Configurations in \model}
\label{app:configuration}

\subsection{Query Analyzer}
\label{app:query_analysis}
\begin{textcolorbox}[Prompt for Query Analyzer]
\textbf{Task:} Analyze the given query with accompanying inputs and determine the skills and tools needed to address it effectively.
\\\\
\textbf{Available tools:} \texttt{\{self.available\_tools\}}
\\\\
\textbf{Metadata for the tools:} \texttt{\{self.toolbox\_metadata\}}
\\\\
\textbf{Image:} \texttt{\{image\_info\}}
\\\\
\textbf{Query:} \texttt{\{question\}}
\\\\
\textbf{Instructions:}
\\
1. Carefully read and understand the query and any accompanying inputs.
\\
2. Identify the main objectives or tasks within the query.
\\
3. List the specific skills that would be necessary to address the query comprehensively.
\\
4. Examine the available tools in the toolbox and determine which ones might relevant and useful for addressing the query. Make sure to consider the user metadata for each tool, including limitations and potential applications (if available).
\\
5. Provide a brief explanation for each skill and tool you've identified, describing how it would contribute to answering the query.
\\\\
\textbf{Your response should include:}
\\
1. A concise summary of the query's main points and objectives, as well as content in any accompanying inputs.
\\
2. A list of required skills, with a brief explanation for each.
\\
3. A list of relevant tools from the toolbox, with a brief explanation of how each tool would be utilized and its potential limitations.
\\
4. Any additional considerations that might be important for addressing the query effectively.
\\
Please present your analysis in a clear, structured format.
\end{textcolorbox}

\clearpage
\subsection{Action Predictor}
\label{app:action_prediction}
\begin{textcolorbox}[Prompt for Action Predictor]

\textbf{Task:} Determine the optimal next step to address the given query based on the provided analysis, available tools, and previous steps taken.
\\\\
\textbf{Query:} \texttt{\{question\}}
\\
\textbf{Image:} \texttt{\{image\}}
\\
\textbf{Query Analysis:} \texttt{\{query\_analysis\}}
\\
\textbf{Available Tools:}
\texttt{\{self.available\_tools\}}
\\
\textbf{Tool Metadata:}
\texttt{\{self.toolbox\_metadata\}}

\textbf{Previous Steps and Their Results:}
\texttt{\{memory.get\_actions()\}}

\textbf{Current Step:} \texttt{\{step\_count\}} in \texttt{\{max\_step\_count\}} steps

\textbf{Remaining Steps:} \texttt{\{max\_step\_count - step\_count\}}
\\\\
\textbf{Instructions:}

1. Analyze the context thoroughly, including the query, its analysis, any image, available tools and their metadata, and previous steps taken.

2. Determine the most appropriate next step by considering:

\quad - Key objectives from the query analysis
   
\quad - Capabilities of available tools
   
\quad - Logical progression of problem-solving
   
\quad - Outcomes from previous steps
   
\quad - Current step count and remaining steps

3. Select ONE tool best suited for the next step, keeping in mind the limited number of remaining steps.

4. Formulate a specific, achievable sub-goal for the selected tool that maximizes progress towards answering the query.
\\\\
\textbf{Output Format:}

\textless{}\texttt{justification}\textgreater{}: detailed explanation of why the selected tool is the best choice for the next step, considering the context and previous outcomes.

\textless{}\texttt{context}\textgreater{}: MUST include ALL necessary information for the tool to function, structured as follows:

\quad * Relevant data from previous steps
    
\quad * File names or paths created or used in previous steps (list EACH ONE individually)
    
\quad * Variable names and their values from previous steps' results
    
\quad * Any other context-specific information required by the tool
    
\textless{}\texttt{sub\_goal}\textgreater{}: a specific, achievable objective for the tool, based on its metadata and previous outcomes. It MUST contain any involved data, file names, and variables from Previous Steps and Their Results that the tool can act upon.

\textless{}\texttt{tool\_name}\textgreater{}: MUST be the exact name of a tool from the available tools list.
\\\\
\textbf{Rules:}

\quad - Select only ONE tool for this step.

\quad - The sub-goal MUST directly address the query and be achievable by the selected tool.

\quad - The Context section MUST include ALL necessary information for the tool to function, including ALL relevant file paths, data, and variables from previous steps.

\quad - The tool name MUST exactly match one from the available tools list: \texttt{\{self.available\_tools\}}.

\quad - Avoid redundancy by considering previous steps and building on prior results.
% \\\\
% \textbf{Example (do not copy, use only as reference):}

% \textless{}\texttt{justification}\textgreater{}: [Your detailed explanation here]

% \textless{}\texttt{context}\textgreater{}: Image path: \texttt{"example/image.jpg"}, Previous detection results: [list of objects]

% \textless{}\texttt{sub\_goal}\textgreater{}: Detect and count the number of specific objects in the image \texttt{"example/image.jpg"}

% \textless{}\texttt{tool\_name}\textgreater{}: \texttt{Object\_Detector\_Tool}
\end{textcolorbox}

\clearpage
\subsection{Command Predictor}
\label{app:command_prediction}
\begin{textcolorbox}[Prompt for Command Predictor]
\textbf{Task:} Generate a precise command to execute the selected tool based on the given information.
\\\\
\textbf{Query:} \texttt{\{question\}}
\\
\textbf{Image:} \texttt{\{image\}}
\\
\textbf{Context:} \texttt{\{context\}}
\\
\textbf{Sub-Goal:} \texttt{\{sub\_goal\}}
\\
\textbf{Selected Tool:} \texttt{\{tool\_name\}}
\\
\textbf{Tool Metadata:} \texttt{\{tool\_metadata\}}
\\\\
\textbf{Instructions:}

1. Carefully review all provided information: the query, image path, context, sub-goal, selected tool, and tool metadata.

2. Analyze the tool's \texttt{input\_types} from the metadata to understand required and optional parameters.

3. Construct a command or series of commands that aligns with the tool's usage pattern and addresses the sub-goal.

4. Ensure all required parameters are included and properly formatted.

5. Use appropriate values for parameters based on the given context, particularly the \texttt{Context} field which may contain relevant information from previous steps.

6. If multiple steps are needed to prepare data for the tool, include them in the command construction.
\\\\
\textbf{Output Format:}

\textless{}\texttt{analysis}\textgreater{}: a step-by-step analysis of the context, sub-goal, and selected tool to guide the command construction.

\textless{}\texttt{explanation}\textgreater{}: a detailed explanation of the constructed command(s) and their parameters.

\textless{}\texttt{command}\textgreater{}: the Python code to execute the tool, which can be one of the following types:

\quad a. A single line command with \texttt{execution = tool.execute()}.

\quad b. A multi-line command with complex data preparation, ending with \texttt{execution = tool.execute()}.

\quad  c. Multiple lines of \texttt{execution = tool.execute()} calls for processing multiple items.
\begin{codebox}
```
python
<your command here>
```
\end{codebox}
\textbf{Rules:}

1. The command MUST be valid Python code and include at least one call to \texttt{tool.execute()}.

2. Each \texttt{tool.execute()} call MUST be assigned to the \texttt{execution} variable in the format \texttt{execution = tool.execute(...)}.

3. For multiple executions, use separate \texttt{execution = tool.execute()} calls for each execution.

4. The final output MUST be assigned to the \texttt{execution} variable, either directly from \texttt{tool.execute()} or as a processed form of multiple executions.

5. Use the exact parameter names as specified in the tool's \texttt{input\_types}.

6. Enclose string values in quotes, use appropriate data types for other values (e.g., lists, numbers).

7. Do not include any code or text that is not part of the actual command.

8. Ensure the command directly addresses the sub-goal and query.

9. Include ALL required parameters, data, and paths to execute the tool in the command itself.

10. If preparation steps are needed, include them as separate Python statements before the \texttt{tool.execute()} calls.
\end{textcolorbox}

\begin{textcolorbox}[Prompt for Command Prediction (Continued)]

\textbf{Examples (Not to use directly unless relevant):}
\\\\
\textbf{Example 1 (Single line command):}

 \textless{}\texttt{analysis}\textgreater{}: The tool requires an image path and a list of labels for object detection.
 
\textless{}\texttt{explanation}\textgreater{}: We pass the image path and a list containing ``baseball'' as the label to detect.

\textless{}\texttt{command}\textgreater{}:
\begin{codebox}
```
python
execution = tool.execute(image="path/to/image", labels=["baseball"])
```
\end{codebox}

\textbf{Example 2 (Multi-line command with data preparation):}

\textless{}\texttt{analysis}\textgreater{}: The tool requires an image path, multiple labels, and a threshold for object detection.

\textless{}\texttt{explanation}\textgreater{}: We prepare the data by defining variables for the image path, labels, and threshold, then pass these to the \texttt{tool.execute()} function.

\textless{}\texttt{command}\textgreater{}:
\begin{codebox}
```
python
image = "path/to/image"
labels = ["baseball", "football", "basketball"]
threshold = 0.5
execution = tool.execute(image=image, labels=labels, threshold=threshold)
```
\end{codebox}

\textbf{Example 3 (Multiple executions):}

\textless{}\texttt{analysis}\textgreater{}: We need to process multiple images for baseball detection.

\textless{}\texttt{explanation}\textgreater{}: We call the tool for each image path, using the same label and threshold for all.

\textless{}\texttt{command}\textgreater{}:
\begin{codebox}
```
python
execution = tool.execute(image="path/to/image1", labels=["baseball"], threshold=0.5)
execution = tool.execute(image="path/to/image2", labels=["baseball"], threshold=0.5)
execution = tool.execute(image="path/to/image3", labels=["baseball"], threshold=0.5)
```
\end{codebox}
\end{textcolorbox}

\begin{textcolorbox}[Prompt for Command Predictor (Continued)]
\textbf{Some Wrong Examples:}

\textless{}\texttt{command}\textgreater{}:
\begin{codebox}
```
python
execution1 = tool.execute(query="...")
execution2 = tool.execute(query="...")
```
\end{codebox}

\textbf{Reason:} only \texttt{execution = tool.execute} is allowed, not \texttt{execution1} or \texttt{execution2}.
\\\\
\textless{}\texttt{command}\textgreater{}:
\begin{codebox}
```
python
urls = [
    "https://example.com/article1",
    "https://example.com/article2"
]

execution = tool.execute(url=urls[0])
execution = tool.execute(url=urls[1])
```
\end{codebox}

\textbf{Reason:} The command should process multiple items in a single execution, not separate executions for each item.
\\\\
\textbf{Remember:} Your \textless{}\texttt{command}\textgreater{} field MUST be valid Python code including any necessary data preparation steps and one or more \texttt{execution = tool.execute()} calls, without any additional explanatory text. The format \texttt{execution = tool.execute} must be strictly followed, and the last line must begin with \texttt{execution = tool.execute} to capture the final output.
\end{textcolorbox}

\clearpage
\subsection{Context Verifier}
\label{app:verification}
\begin{textcolorbox}[Prompt for Verifier]
\textbf{Task:} Thoroughly evaluate the completeness and accuracy of the memory for fulfilling the given query, considering the potential need for additional tool usage.
\\\\
\textbf{Query:} \texttt{\{question\}}
\\
\textbf{Image:} \texttt{\{image\_info\}}
\\
\textbf{Available Tools:} \texttt{\{self.available\_tools\}}
\\
\textbf{Toolbox Metadata:} \texttt{\{self.toolbox\_metadata\}}
\\
\textbf{Initial Analysis:} \texttt{\{query\_analysis\}}
\\
\textbf{Memory (tools used and results):} \texttt{\{memory.get\_actions()\}}
\\\\
\textbf{Detailed Instructions:}

1. Carefully analyze the query, initial analysis, and image (if provided):

\quad - Identify the main objectives of the query.

\quad - Note any specific requirements or constraints mentioned.

\quad - If an image is provided, consider its relevance and what information it contributes.
\\
2. Review the available tools and their metadata:

\quad - Understand the capabilities and limitations and best practices of each tool.

\quad - Consider how each tool might be applicable to the query.
\\
3. Examine the memory content in detail:

\quad - Review each tool used and its execution results.

\quad - Assess how well each tool's output contributes to answering the query.
\\
4. Critical Evaluation (address each point explicitly):

\quad a) Completeness: Does the memory fully address all aspects of the query?

\quad\quad - Identify any parts of the query that remain unanswered.

\quad\quad - Consider if all relevant information has been extracted from the image (if applicable).

\quad b) Unused Tools: Are there any unused tools that could provide additional relevant information?

\quad\quad - Specify which unused tools might be helpful and why.

\quad c) Inconsistencies: Are there any contradictions or conflicts in the information provided?

\quad\quad - If yes, explain the inconsistencies and suggest how they might be resolved.

\quad d) Verification Needs: Is there any information that requires further verification due to tool limitations?

\quad\quad - Identify specific pieces of information that need verification and explain why.

\quad e) Ambiguities: Are there any unclear or ambiguous results that could be clarified by using another tool?

\quad\quad - Point out specific ambiguities and suggest which tools could help clarify them.
\\
5. Final Determination:
   Based on your thorough analysis, decide if the memory is complete and accurate enough to generate the final output, or if additional tool usage is necessary.
\\\\
\textbf{Response Format:}

\textless{}\texttt{analysis}\textgreater{}: Provide a detailed analysis of why the memory is sufficient. Reference specific information from the memory and explain its relevance to each aspect of the task. Address how each main point of the query has been satisfied.

\textless{}\texttt{stop\_signal}\textgreater{}: Whether to stop the problem solving process and proceed to generating the final output.

\quad * \texttt{"True"}: if the memory is sufficient for addressing the query to proceed and no additional available tools need to be used. If ONLY manual verification without tools is needed, choose \texttt{"True"}.

\quad * \texttt{"False"}: if the memory is insufficient and needs more information from additional tool usage.
\end{textcolorbox}

\clearpage
\subsection{Solution Summarizer}
\label{app:solution_summarization}
\begin{textcolorbox}[Prompt for Solution Summarizer]
\textbf{Task:} Generate the final output based on the query, image, and tools used in the process.
\\\\
\textbf{Query:} \texttt{\{question\}}
\\
\textbf{Image:} \texttt{\{image\_info\}}
\\
\textbf{Actions Taken:} \texttt{\{memory.get\_actions()\}}
\\\\
\textbf{Instructions:}

1. Review the query, image, and all actions taken during the process.

2. Consider the results obtained from each tool execution.

3. Incorporate the relevant information from the memory to generate the step-by-step final output.

4. The final output should be consistent and coherent using the results from the tools.
\\\\
\textbf{Output Structure:}

Your response should be well-organized and include the following sections:

1. Summary:

\quad - Provide a brief overview of the query and the main findings.
\\\\
2. Detailed Analysis:

\quad - Break down the process of answering the query step-by-step.

\quad - For each step, mention the tool used, its purpose, and the key results obtained.

\quad - Explain how each step contributed to addressing the query.
\\\\
3. Key Findings:

\quad - List the most important discoveries or insights gained from the analysis.

\quad - Highlight any unexpected or particularly interesting results.
\\\\
4. Answer to the Query:

\quad - Directly address the original question with a clear and concise answer.

\quad - If the query has multiple parts, ensure each part is answered separately.
\\\\
5. Additional Insights (if applicable):

\quad - Provide any relevant information or insights that go beyond the direct answer to the query.

\quad - Discuss any limitations or areas of uncertainty in the analysis.
\\\\
6. Conclusion:

\quad - Summarize the main points and reinforce the answer to the query.

\quad - If appropriate, suggest potential next steps or areas for further investigation.
\end{textcolorbox}

\clearpage
\section{Tool Cards in \model}
\label{app:toolcards}

\subsection{ArXiv Paper Searcher Tool}
\label{app:arxiv_paper_searcher_tool}
\begin{codecolorbox}[ArXiv Paper Searcher Tool: Metadata]{python}
tool_name="ArXiv_Paper_Searcher_Tool",

tool_description="A tool that searches arXiv for papers based on a given query.",

input_types={
    "query": "str - The search query for arXiv papers.",
    "size": "int - The number of results per page (25, 50, 100, or 200). If None, use 25.",
    "max_results": "int - The maximum number of papers to return (default: 25). Should be less than or equal to 100."
},

output_type="list - A list of dictionaries containing paper information.",

demo_commands=[
    {
        "command": 'execution = tool.execute(query="tool agents with large language models")',
        "description": "Search for papers about tool agents with large language models."
    },
    {
        "command": 'execution = tool.execute(query="quantum computing", size=100, max_results=50)',
        "description": "Search for quantum computing papers, with 100 results per page, returning a maximum of 50 papers."
    },
    {
        "command": 'execution = tool.execute(query="machine learning", max_results=75)',
        "description": "Search for machine learning papers, returning a maximum of 75 papers."
    },
],

user_metadata={
    "valid_sizes": [25, 50, 100, 200],
    "base_url": "https://arxiv.org/search/"
}
\end{codecolorbox}

\begin{textcolorbox}[ArXiv Paper Searcher Tool: Example 1]
\toolinput{query}
enhance mathematical reasoning with large language models
\\\\
% \toolinput{command}
% \\
% \command{tool.execute(query="enhance mathematical reasoning with large language models")}
% \\\\
\tooloutput{article 1, search date: Jan 14, 2025}
\\\\
\textbf{Title:} Enhancing Reasoning through Process Supervision with Monte Carlo Tree Search

\textbf{Authors:} Shuangtao Li, Shuaihao Dong, Kexin Luan, Xinhan Di, Chaofan Ding

\textbf{Abstract:} Large language models (LLMs) have demonstrated their remarkable capacity across a variety of tasks. 
...
% However, reasoning remains a challenge for LLMs. To improve LLMs' reasoning ability, process supervision has proven to be better than outcome supervision. In this work, we study using Monte Carlo Tree Search (MCTS) to generate process supervision data with LLMs themselves for training them. We sample reasoning steps with an LLM and assign each step a score that captures its \"relative correctness,\" and the LLM is then trained by minimizing weighted log-likelihood of generating the reasoning steps. This generate-then-train process is repeated iteratively until convergence.Our experimental results demonstrate that the proposed methods considerably improve the performance of LLMs on two mathematical reasoning datasets. 
Furthermore, models trained on one dataset also exhibit improved performance on the other, showing the transferability of the enhanced reasoning ability.

\textbf{Link:} \url{https://arxiv.org/abs/2501.01478}
\\\\
\tooloutput{article 2, search date: Jan 14, 2025}
\\\\
\textbf{Title:} Open Eyes, Then Reason: Fine-grained Visual Mathematical Understanding in MLLMs

\textbf{Authors:} Shan Zhang, Aotian Chen, Yanpeng Sun, Jindong Gu, Yi-Yu Zheng, Piotr Koniusz, Kai Zou, Anton van den Hengel, Yuan Xue

\textbf{Abstract:}  Current multimodal large language models (MLLMs) often underperform on mathematical problem-solving tasks that require fine-grained visual understanding. ...
% The limitation is largely attributable to inadequate perception of geometric primitives during image-level contrastive pre-training (e.g., CLIP). While recent efforts to improve math MLLMs have focused on scaling up mathematical visual instruction datasets and employing stronger LLM backbones, they often overlook persistent errors in visual recognition. In this paper, we systematically evaluate the visual grounding capabilities of state-of-the-art MLLMs and reveal a significant negative correlation between visual grounding accuracy and problem-solving performance, underscoring the critical role of fine-grained visual understanding. Notably, advanced models like GPT-4o exhibit a 70% error rate when identifying geometric entities, highlighting that this remains a key bottleneck in visual mathematical reasoning. To address this, we propose a novel approach, SVE-Math (Selective Vision-Enhanced Mathematical MLLM), featuring a geometric-grounded vision encoder and a feature router that dynamically adjusts the contribution of hierarchical visual feature maps. Our model recognizes accurate visual primitives and generates precise visual prompts tailored to the language model's reasoning needs. In experiments, SVE-Math-Qwen2.5-7B outperforms other 7B models by 15% on MathVerse and is compatible with GPT-4V on MathVista. Despite being trained on smaller datasets, SVE-Math-7B achieves competitive performance on GeoQA, rivaling models trained on significantly larger datasets. 
Our findings emphasize the importance of incorporating fine-grained visual understanding into MLLMs and provide a promising direction for future research.

\textbf{Link:} \url{https://arxiv.org/abs/2501.06430}
\\\\
\tooloutput{structured result}
\begin{codebox}
[
    {
        "title": "Enhancing Reasoning through Process Supervision with Monte Carlo Tree Search",
        "authors": "Shuangtao Li, Shuaihao Dong, Kexin Luan, Xinhan Di, Chaofan Ding",
        "abstract": "Large language models (LLMs) have demonstrated their remarkable capacity across a variety of tasks. ... Furthermore, models trained on one dataset also exhibit improved performance on the other, showing the transferability of the enhanced reasoning ability.",
        "link": "https://arxiv.org/abs/2501.01478"
    },
    {
        "title": "Open Eyes, Then Reason: Fine-grained Visual Mathematical Understanding in MLLMs",
        "authors": "Shan Zhang, Aotian Chen, Yanpeng Sun, Jindong Gu, Yi-Yu Zheng, Piotr Koniusz, Kai Zou, Anton van den Hengel, Yuan Xue",
        "abstract": "Current multimodal large language models (MLLMs) often underperform on mathematical problem-solving tasks that require fine-grained visual understanding. ... Our findings emphasize the importance of incorporating fine-grained visual understanding into MLLMs and provide a promising direction for future research.",
        "link": "https://arxiv.org/abs/2501.06430"
    }
]
\end{codebox}

\end{textcolorbox}

\begin{textcolorbox}[ArXiv Paper Searcher Tool: Example 2]
\toolinput{query}
automated scientific discovery
\\\\
\tooloutput{article 1, search date: Jan 14, 2025}
\\\\
\textbf{Title:} BoxingGym: Benchmarking Progress in Automated Experimental Design and Model Discover

\textbf{Authors:} Kanishk Gandhi, Michael Y. Li, Lyle Goodyear, Louise Li, Aditi Bhaskar, Mohammed Zaman, Noah D. Goodman

\textbf{Abstract:} Understanding the world and explaining it with scientific theories is a central aspiration of artificial intelligence research. Proposing theories, designing experiments to test them, and then revising them based on data are fundamental to scientific discovery. 
...
% Despite the significant promise of LLM-based scientific agents, no benchmarks systematically test LLM's ability to propose scientific models, collect experimental data, and revise them in light of new data. We introduce BoxingGym, a benchmark with 10 environments for systematically evaluating both experimental design (e.g. collecting data to test a scientific theory) and model discovery (e.g. proposing and revising scientific theories). To enable tractable and quantitative evaluation, we implement each environment as a generative probabilistic model with which a scientific agent can run interactive experiments. These probabilistic models are drawn from various real-world scientific domains ranging from psychology to ecology. To quantitatively evaluate a scientific agent's ability to collect informative experimental data, we compute the expected information gain (EIG), an information-theoretic quantity which measures how much an experiment reduces uncertainty about the parameters of a generative model. A good scientific theory is a concise and predictive explanation. Therefore, to quantitatively evaluate model discovery, we ask a scientific agent to explain their model and then assess whether this explanation enables another scientific agent to make reliable predictions about this environment. In addition to this explanation-based evaluation, we compute standard model evaluation metrics such as prediction errors. We find that current LLMs, such as GPT-4o, struggle with both experimental design and model discovery.
We find that augmenting the LLM-based agent with an explicit statistical model does not reliably improve these results.

\textbf{Link:} \url{https://arxiv.org/abs/2501.01540}
\\\\
\tooloutput{article 2, search date: Jan 14, 2025}
\\\\
\textbf{Title:} Automating the Search for Artificial Life with Foundation Models

\textbf{Authors:} Akarsh Kumar, Chris Lu, Louis Kirsch, Yujin Tang, Kenneth O. Stanley, Phillip Isola, David Ha

\textbf{Abstract:}  With the recent Nobel Prize awarded for radical advances in protein discovery, foundation models (FMs) for exploring large combinatorial spaces promise to revolutionize many scientific fields. 
...
% Artificial Life (ALife) has not yet integrated FMs, thus presenting a major opportunity for the field to alleviate the historical burden of relying chiefly on manual design and trial-and-error to discover the configurations of lifelike simulations. This paper presents, for the first time, a successful realization of this opportunity using vision-language FMs. The proposed approach, called Automated Search for Artificial Life (ASAL), (1) finds simulations that produce target phenomena, (2) discovers simulations that generate temporally open-ended novelty, and (3) illuminates an entire space of interestingly diverse simulations. Because of the generality of FMs, ASAL works effectively across a diverse range of ALife substrates including Boids, Particle Life, Game of Life, Lenia, and Neural Cellular Automata. A major result highlighting the potential of this technique is the discovery of previously unseen Lenia and Boids lifeforms, as well as cellular automata that are open-ended like Conway's Game of Life. Additionally, the use of FMs allows for the quantification of previously qualitative phenomena in a human-aligned way. 
This new paradigm promises to accelerate ALife research beyond what is possible through human ingenuity alone.

\textbf{Link:} \url{https://arxiv.org/abs/2412.17799}
\\\\
\tooloutput{structured result}
\begin{codebox}
[
    {
        "title": "BoxingGym: Benchmarking Progress in Automated Experimental Design and Model Discovery",
        "authors": "Kanishk Gandhi, Michael Y. Li, Lyle Goodyear, Louise Li, Aditi Bhaskar, Mohammed Zaman, Noah D. Goodman",
        "abstract": "Understanding the world and explaining it with scientific theories is a central aspiration of artificial intelligence research. ... We find that augmenting the LLM-based agent with an explicit statistical model does not reliably improve these results.",
        "link": "https://arxiv.org/abs/2501.01540"
    },
    {
        "title": "Automating the Search for Artificial Life with Foundation Models",
        "authors": "Akarsh Kumar, Chris Lu, Louis Kirsch, Yujin Tang, Kenneth O. Stanley, Phillip Isola, David Ha",
        "abstract": "With the recent Nobel Prize awarded for radical advances in protein discovery, foundation models (FMs) for exploring large combinatorial spaces promise to revolutionize many scientific fields. ... This new paradigm promises to accelerate ALife research beyond what is possible through human ingenuity alone.",
        "link": "https://arxiv.org/abs/2412.17799"
    }
]
\end{codebox}

\end{textcolorbox}

\clearpage
\subsection{Generalist Solution Generator Tool}
\label{app:generalist_solution_generator_tool}
\begin{codecolorbox}[Generalist Solution Generator Tool: Metadata]{python}
tool_name="Generalist_Solution_Generator_Tool",

tool_description="A generalized tool that takes query from the user as prompt, and answers the question step by step to the best of its ability. It can also accept an image.",

input_types={
    "prompt": "str - The prompt that includes query from the user to guide the agent to generate response (Examples: 'Describe this image in detail').",
    "image": "str - The path to the image file if applicable (default: None).",
},

output_type="str - The generated response to the original query prompt",

demo_commands=[
    {
        "command": 'execution = tool.execute(prompt="Summarize the following text in a few lines")',
        "description": "Generate a short summary given the prompt from the user."
    },
    {
        "command": 'execution = tool.execute(prompt="Explain the mood of this scene.", image="path/to/image1.png")',
        "description": "Generate a caption focusing on the mood using a specific prompt and image."
    },
    {
        "command": 'execution = tool.execute(prompt="Give your best coordinate estimate for the pacemaker in the image and return (x1, y1, x2, y2)", image="path/to/image2.png")',
        "description": "Generate bounding box coordinates given the image and prompt from the user. The format should be (x1, y1, x2, y2)."
    },
    {
        "command": 'execution = tool.execute(prompt="Is the number of tiny objects that are behind the small metal jet less than the number of tiny things left of the tiny sedan?", image="path/to/image2.png")',
        "description": "Answer a question step by step given the image."
    }
],

user_metadata = {
    "limitation": "The Generalist_Solution_Generator_Tool may provide hallucinated or incorrect responses.",
    "best_practice": "Use the Generalist_Solution_Generator_Tool for general queries or tasks that don't require specialized knowledge or specific tools in the toolbox. For optimal results:\n\n"
    "1) Provide clear, specific prompts.\n"
    "2) Use it to answer the original query through step by step reasoning for tasks without complex or multi-step reasoning.\n"
    "3) For complex queries, break them down into subtasks and use the tool multiple times.\n"
    "4) Use it as a starting point for complex tasks, then refine with specialized tools.\n"
    "5) Verify important information from its responses.\n"
    "6) For image-related tasks, ensure the image path is correct and the prompt is relevant to the image content."
}
\end{codecolorbox}

\begin{textcolorbox}[Generalist Solution Generator Tool: Example 1]
\toolinput{query} Describe the image in detail.

\toolinput{image} \texttt{"baseballs.png"}

\vspace{0.2cm}
\includegraphics[width=0.4\linewidth]{tools/examples/mathvista_113.png}
\vspace{0.2cm}

\tooloutput{response}

The image shows four blue buckets, each containing five baseballs. The baseballs are white with red stitching, and the buckets have handles on the sides. The arrangement of the baseballs within each bucket is slightly different, but each bucket consistently holds five baseballs. The background is plain, emphasizing the buckets and their contents.
\\\\
\tooloutput{structured result}
\begin{codebox}
{
    "The image shows four blue buckets, each containing five baseballs. The baseballs are white with red stitching, and the buckets have handles on the sides. The arrangement of the baseballs within each bucket is slightly different, but each bucket consistently holds five baseballs. The background is plain, emphasizing the buckets and their contents."
}
\end{codebox}

\end{textcolorbox}

\clearpage
\subsection{Google Search Tool}
\label{app:google_search_tool}
\begin{codecolorbox}[Google Search Tool: Metadata]{python}
tool_name="Google_Search_Tool",

tool_description="A tool that performs Google searches based on a given text query.",

input_types={
    "query": "str - The search query to be used for the Google search.",
    "num_results": "int - The number of search results to return (default: 10).",
},

output_type="list - A list of dictionaries containing search result information.",

demo_commands=[
    {
        "command": 'execution = tool.execute(query="Python programming")',
        "description": "Perform a Google search for 'Python programming' and return the default number of results."
    },
    {
        "command": 'execution = tool.execute(query="Machine learning tutorials", num_results=5)',
        "description": "Perform a Google search for 'Machine learning tutorials' and return 5 results."
    },
]
\end{codecolorbox}

\begin{textcolorbox}[Google Search Tool: Example 1]
\toolinput{query}
nobel prize winners in chemistry 2024
\\\\
\tooloutput{result 1, search date: Jan 14, 2025}

\textbf{Title:} Nobel Prize in Chemistry Laureates

\textbf{URL:} \url{https://www.nobelprize.org/prizes/chemistry/}
   
\textbf{Snippet:} The Nobel Prize in Chemistry 2024 was awarded with one half to David Baker “for computational protein design” and the other half jointly to Demis Hassabis and ...
\\\\
\tooloutput{result 2, search date: Jan 14, 2025}

\textbf{Title:} NSF congratulates laureates of the 2024 Nobel Prize in chemistry ...

\textbf{URL:} \url{https://new.nsf.gov/news/nsf-congratulates-laureates-2024-nobel-prize-chemistry}
   
\textbf{Snippet:} Oct 9, 2024 ... The U.S. National Science Foundation congratulates David Baker, Demis Hassabis and John Jumper on being awarded the 2024 Nobel Prize in ...
\\\\
\tooloutput{result 3, search date: Jan 14, 2025}

\textbf{Title:} Press release: The Nobel Prize in Chemistry 2024 - NobelPrize.org

\textbf{URL:} \url{https://www.nobelprize.org/prizes/chemistry/2024/press-release/}
   
\textbf{Snippet:} Oct 9, 2024 ... David Baker has succeeded with the almost impossible feat of building entirely new kinds of proteins. Demis Hassabis and John Jumper have ...
\\\\
\tooloutput{structured result}
\begin{codebox}
[
    {
        "title": "Nobel Prize in Chemistry Laureates",
        "url": "https://www.nobelprize.org/prizes/chemistry/",
        "snippet": "The Nobel Prize in Chemistry 2024 was awarded with one half to David Baker ``for computational protein design'' and the other half jointly to Demis Hassabis and ..."
    },
    {
        "title": "NSF congratulates laureates of the 2024 Nobel Prize in chemistry ...",
        "url": "https://new.nsf.gov/news/nsf-congratulates-laureates-2024-nobel-prize-chemistry",
        "snippet": "Oct 9, 2024 ... The U.S. National Science Foundation congratulates David Baker, Demis Hassabis and John Jumper on being awarded the 2024 Nobel Prize in ..."
    },
    {
        "title": "Press release: The Nobel Prize in Chemistry 2024 - NobelPrize.org",
        "url": "https://www.nobelprize.org/prizes/chemistry/2024/press-release/",
        "snippet": "Oct 9, 2024 ... David Baker has succeeded with the almost impossible feat of building entirely new kinds of proteins. Demis Hassabis and John Jumper have ..."
    }
]
\end{codebox}

\end{textcolorbox}

\clearpage
\subsection{Image Captioner Tool}
\label{app:image_captioner_tool}
\begin{codecolorbox}[Image Captioner Tool: Metadata]{python}
tool_name="Image_Captioner_Tool",

tool_description="A tool that generates captions for images using OpenAI's multimodal model.",

input_types={
    "image": "str - The path to the image file.",
    "prompt": "str - The prompt to guide the image captioning (default: 'Describe this image in detail.').",
},

output_type="str - The generated caption for the image.",

demo_commands=[
    {
        "command": 'execution = tool.execute(image="path/to/image.png")',
        "description": "Generate a caption for an image using the default prompt and model."
    },
    {
        "command": 'execution = tool.execute(image="path/to/image.png", prompt="Explain the mood of this scene.")',
        "description": "Generate a caption focusing on the mood using a specific prompt and model."
    }
],

user_metadata = {
    "limitation": "The Image_Captioner_Tool provides general image descriptions but has limitations: 1) May make mistakes in complex scenes, counting, attribute detection, and understanding object relationships. 2) Might not generate comprehensive captions, especially for images with multiple objects or abstract concepts. 3) Performance varies with image complexity. 4) Struggles with culturally specific or domain-specific content. 5) May overlook details or misinterpret object relationships. For precise descriptions, consider: using it with other tools for context/verification, as an initial step before refinement, or in multi-step processes for ambiguity resolution. Verify critical information with specialized tools or human expertise when necessary."
}
\end{codecolorbox}

\begin{textcolorbox}[Image Captioner Tool: Example 1]
\toolinput{image} \texttt{"baseballs.png"}

\vspace{0.2cm}
\includegraphics[width=0.4\linewidth]{tools/examples/mathvista_113.png}
\vspace{0.2cm}

\tooloutput{caption}

The image shows four blue buckets, each containing five baseballs. The buckets are arranged in a grid pattern with two on the top row and two on the bottom row. Each bucket has a handle on the side, and the baseballs inside are white with red stitching, typical of standard baseballs. The background is plain white, emphasizing the buckets and their contents.
\\\\
\tooloutput{structured result}
\begin{codebox}
{
    "The image shows four blue buckets, each containing five baseballs. The buckets are arranged in a grid pattern with two on the top row and two on the bottom row. Each bucket has a handle on the side, and the baseballs inside are white with red stitching, typical of standard baseballs. The background is plain white, emphasizing the buckets and their contents."
}
\end{codebox}

\end{textcolorbox}

\clearpage
\subsection{Path Generalist Classifier Tool}
\label{app:path_generalist_classifier_tool} 
\begin{codecolorbox}[Path Generalist Classifier Tool: Metadata]{python}
tool_name="Path_Generalist_Classifier_Tool",

tool_description="A tool for answering multiple choice questions about H&E microscopy images. Do NOT use for open-ended questions. Do NOT use for images that are not H&E-stained.",

input_types={
    "image": "str - The path to the histopathology image.",
    "options": "list[str] - A list of options to classify the image against."
},

output_type="str - The classification result.",

demo_commands=[
    {
        "command": 'execution = tool.execute(image="path/to/image.jpg", options=["lung adenocarcinoma", "lung squamous cell carcinoma"])',
        "description": "Classify the image into one of the given options."
    },
    {
        "command": 'execution = tool.execute(image="path/to/image.png", options=["debris" "cancer-associated stroma", "adipose", "normal colon mucosa", "colorectal adenocarcinoma epithelium", "none of the above"])',
        "description": "Classify the image into one of the given options."
    }
],

user_metadata={
    "limitations": "This tool is designed for answering classification questions about H&E-stained microscopy images. This tool is not suitable for open ended questions. Do NOT use this tool if the input is a natural image, a medical image of other domains (such as IHC, CT, MRI, or X-ray images), or a raw whole slide image (i.e., svs, ndpi, czi, etc). This tool is not always reliable and the result should be cross-referenced by other tools or your own knowledge.",
    "best_practice": "Provide clear and specific options for classification. This tool is ideal for classification tasks where the options are well-defined and specific to histopathology (H&E) images."
}
\end{codecolorbox}

\begin{textcolorbox}[Path Generalist Classifier Tool: Example 1]

\toolinput{query} \texttt{["Non-tumor", "Necrotic tumor", "Viable tumor"]}

\toolinput{image} \texttt{"tissue.png"}

\vspace{0.2cm}
\includegraphics[width=0.4\linewidth]{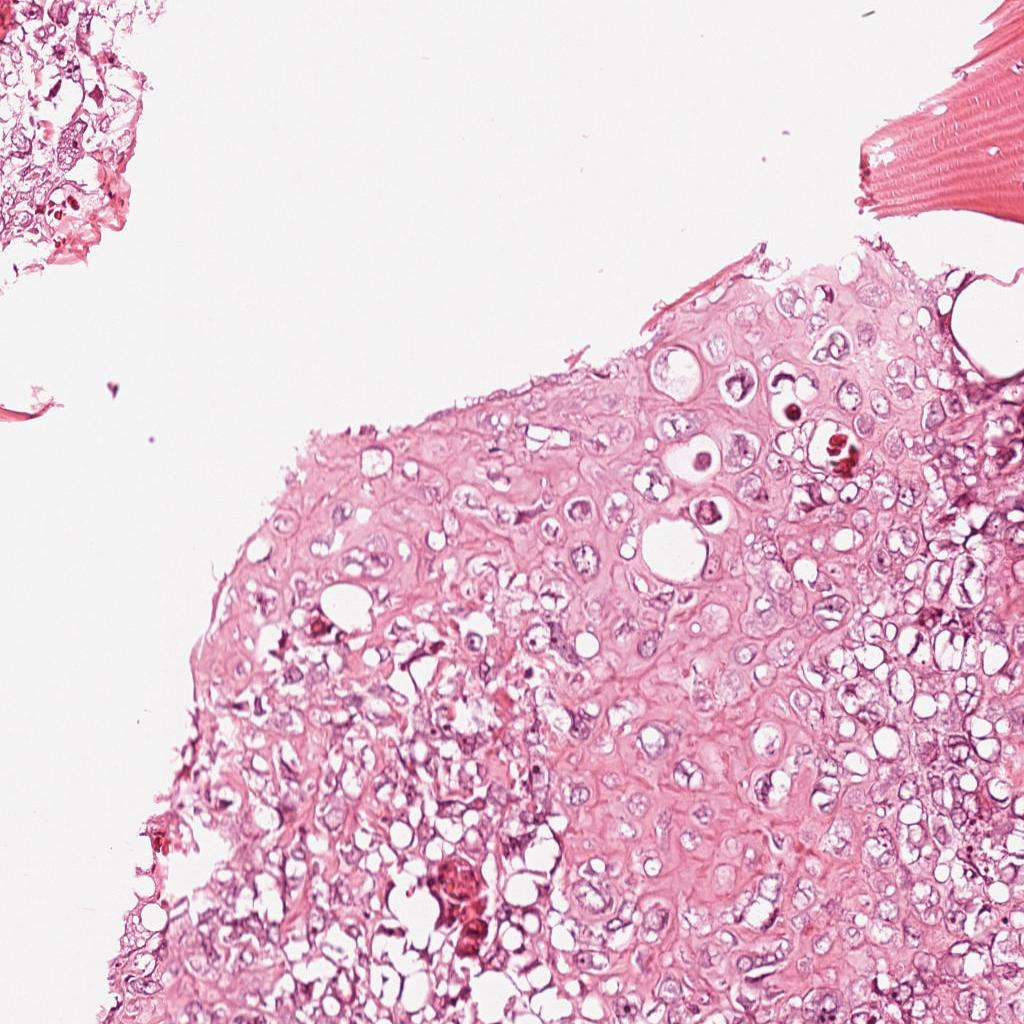}
\\\\
\tooloutput{type}
\begin{codebox}
Viable tumor
\end{codebox}
\tooloutput{structured result}
\begin{codebox}
{
    "result": "Viable tumor"
}
\end{codebox}

\end{textcolorbox}

\clearpage
\subsection{Pubmed Search Tool}
\label{app:pubmed_search_tool}
\begin{codecolorbox}[Pubmed Search Tool: Metadata]{python}
tool_name="Pubmed_Search_Tool",

tool_description="A tool that searches PubMed Central to retrieve relevant article abstracts based on a given list of text queries. Use this ONLY if you cannot use the other more specific ontology tools.",

input_types={
    "queries": "list[str] - list of queries terms for searching PubMed."
},

output_type="list - List of items matching the search query. Each item consists of the title, abstract, keywords, and URL of the article. If no results found, a string message is returned.",

demo_commands=[
    {
        "command": 'execution = tool.execute(queries=["scoliosis", "injury"])',
        "description": "Search for PubMed articles mentioning 'scoliosis' OR 'injury'."
    },
    {
        "command": 'execution = tool.execute(queries=["COVID", "vaccine", "occupational health"])',
        "description": "Search for PubMed articles mentioning 'COVID' OR 'vaccine' OR 'occupational health'."
    }
],

user_metadata={
    "limitations": "Try to use shorter and more general search queries."
}
\end{codecolorbox}

\begin{textcolorbox}[Pubmed Search Tool: Example 1]
\toolinput{query}
COVID occupational health

\tooloutput{result 1}

\textbf{Title:} COVID-19 workplace countermeasures that occupational physicians...

\textbf{Abstract:} BACKGROUND: During the COVID-19 pandemic, information and circumstances changed from moment to moment...

\textbf{Keywords:} COVID-19, Japan, Occupational health,...

\textbf{URL:} \url{https://ncbi.nlm.nih.gov/pubmed/39780108}
\\\\
\tooloutput{result 2}

\textbf{Title:} Rapid COVID-19 Testing of Symptomatic Health Care Personnel...

\textbf{Abstract:} Determine performance characteristics and safety outcomes of two rapid COVID-19 screening methods ...

\textbf{Keywords:}

\textbf{URL:} \url{https://ncbi.nlm.nih.gov/pubmed/39739739}
\\\\
\tooloutput{result 3}
\\\\
\textbf{Title:} Satisfaction and Workload as Predictors of Psychological Distress in Professionals of Psychosocial Care Centers During the COVID-19 Pandemic.

\textbf{Abstract:} BACKGROUND AND AIMS: The COVID-19 pandemic significantly impacted the mental health of healthcare professionals...

\textbf{Keywords:} COVID-19, health personnel, job satisfaction,...

\textbf{URL:} \url{https://ncbi.nlm.nih.gov/pubmed/39728651}
\\\\
\tooloutput{structured result}
\begin{codebox}
[
    {
        "title": "COVID-19 workplace countermeasures that occupational physicians...",
        "abstract": "BACKGROUND: During the COVID-19 pandemic, information and circumstances changed from moment to moment...",
        "keywords": ["COVID-19", "Japan", "Occupational health", ...],
        "url": "https://ncbi.nlm.nih.gov/pubmed/39780108"
    },
    {
        "title": "Rapid COVID-19 Testing of Symptomatic Health Care Personnel...",
        "abstract": "Determine performance characteristics and safety outcomes of two rapid COVID-19 screening methods...",
        "keywords": [],
        "url": "https://ncbi.nlm.nih.gov/pubmed/39739739"
    },
    {
        "title": "Satisfaction and Workload as Predictors of Psychological Distress in Professionals of Psychosocial Care Centers During the COVID-19 Pandemic.",
        "abstract": "BACKGROUND AND AIMS: The COVID-19 pandemic significantly impacted the mental health of healthcare professionals...",
        "keywords": ["COVID-19", "health personnel", "job satisfaction", ...],
        "url": "https://ncbi.nlm.nih.gov/pubmed/39728651"
    }
]
\end{codebox}

\end{textcolorbox}

\clearpage
\subsection{Python Code Generator Tool}
\label{app:python_code_generator_tool}
\begin{codecolorbox}[Python Code Generator Tool: Metadata]{python}
tool_name="Python_Code_Generator_Tool",

tool_description="A tool that generates and executes simple Python code snippets for basic arithmetical calculations and math-related problems. The generated code runs in a highly restricted environment with only basic mathematical operations available.",

input_types={
    "query": "str - A clear, specific description of the arithmetic calculation or math problem to be solved, including any necessary numerical inputs."},

output_type="dict - A dictionary containing the generated code, calculation result, and any error messages.",

demo_commands=[
    {
        "command": 'execution = tool.execute(query="Calculate the factorial of 5")',
        "description": "Generate a Python code snippet to calculate the factorial of 5."
    },
    {
        "command": 'execution = tool.execute(query="Find the sum of prime numbers up to 50")',
        "description": "Generate a Python code snippet to find the sum of prime numbers up to 50."
    },
    {
        "command": 'query="Given the list [1, 2, 3, 4, 5, 6, 7, 8, 9, 10], calculate the sum of squares of odd numbers"\nexecution = tool.execute(query=query)',
        "description": "Generate a Python function for a specific mathematical operation on a given list of numbers."
    },
],
user_metadata = {
    "limitations": [
        "Restricted to basic Python arithmetic operations and built-in mathematical functions.",
        "Cannot use any external libraries or modules, including those in the Python standard library.",
        "Limited to simple mathematical calculations and problems.",
        "Cannot perform any string processing, data structure manipulation, or complex algorithms.",
        "No access to any system resources, file operations, or network requests.",
        "Cannot use 'import' statements.",
        "All calculations must be self-contained within a single function or script.",
        "Input must be provided directly in the query string.",
        "Output is limited to numerical results or simple lists/tuples of numbers."
    ],
    "best_practices": [
        "Provide clear and specific queries that describe the desired mathematical calculation.",
        "Include all necessary numerical inputs directly in the query string.",
        "Keep tasks focused on basic arithmetic, algebraic calculations, or simple mathematical algorithms.",
        "Ensure all required numerical data is included in the query.",
        "Verify that the query only involves mathematical operations and does not require any data processing or complex algorithms.",
        "Review generated code to ensure it only uses basic Python arithmetic operations and built-in math functions."
    ]
}
\end{codecolorbox}

\begin{textcolorbox}[Python Code Generator Tool: Example 1]
\toolinput{query}
Given the number list: [1, 2, 3, 4, 5], calculate the sum of all the numbers in the list.
\\\\
\tooloutput{code}
\begin{codebox}
    numbers = [1, 2, 3, 4, 5]
    result = sum(numbers)
    print(f'The sum is: {result}')
\end{codebox}
\tooloutput{structured result}
\begin{codebox}
{
    "generated_code": """
    numbers = [1, 2, 3, 4, 5]
    result = sum(numbers)
    print(f'The sum is: {result}')
    """,
    "execution_result": "The sum is: 15",
    "error": None
}
\end{codebox}

\end{textcolorbox}

% \end{codecolorbox}
% \begin{codecolorbox}[Python Code Generator Tool: Metadata (Continued)]{python}

\clearpage
\subsection{Relevant Patch Zoomer Tool}
\label{app:relevant_patch_zoomer_tool}
\begin{codecolorbox}[Relevant Patch Zoomer Tool: Metadata]{python}
tool_name="Relevant_Patch_Zoomer_Tool",

tool_description="A tool that analyzes an image, divides it into 5 regions (4 quarters + center), and identifies the most relevant patches based on a question. The returned patches are zoomed in by a factor of 2.",

input_types={
    "image": "str - The path to the image file.",
    "question": "str - The question about the image content.",
},

output_type="dict - Contains analysis text and list of saved zoomed patch paths.",

demo_commands=[
    {
        "command": 'execution = tool.execute(image="path/to/image.jpg", question="What is the color of the car?")',
        "description": "Analyze image and return relevant zoomed patches that show the car's color."
    }
],

user_metadata = {
    "best_practices": [
        "It might be helpful to zoom in on the image first to get a better look at the object(s).",
        "It might be helpful if the question requires a close-up view of the object(s), symbols, texts, etc.",
        "The tool should be used to provide a high-level analysis first, and then use other tools for fine-grained analysis. For example, you can use Relevant_Patch_Zoomer_Tool first to get a zoomed patch of specific objects, and then use Image_Captioner_Tool to describe the objects in detail."
    ]
}
\end{codecolorbox}

\begin{textcolorbox}[Relevant Patch Zoomer Tool: Example 1]
\toolinput{query} What is the color of the car?

\toolinput{image} \texttt{"car.png"}

\vspace{0.2cm}
\includegraphics[width=0.3\linewidth]{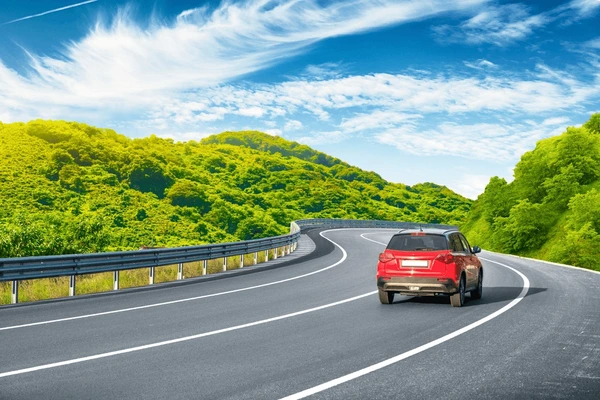}
\vspace{0.2cm}

\tooloutput{analysis}

The image shows a red car on a curvy road with green hills and a blue sky with clouds in the background. Here is what is visible in each region:

\quad (A) Top-left quarter: Contains the blue sky, clouds, and a small part of green hills. No car is visible here.

\quad (B) Top-right quarter: Contains the sky, clouds, and the tops of the green hills. The car is not visible.

\quad (C) Bottom-left quarter: Shows part of the road and guardrail. The car is partially visible.

\quad (D) Bottom-right quarter: Contains the road, guardrail, and most of the car, including its rear view.

\quad (E) Center region: Overlaps with parts of the road, hills, and car. The car's rear is partially visible.
\\\\
The task is to identify the color of the car. Regions (D) and (E) are relevant as they show the car, but region (D) contains the full view of the car, making region (E) unnecessary.
\\\\
\tooloutput{patch region} (D) bottom-right quarter

\tooloutput{patch image} \texttt{"car\_bottom-right\_zoomed\_2x.png"}

\vspace{0.2cm}
\includegraphics[width=0.3\linewidth]{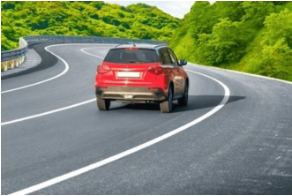}
\vspace{0.2cm}

\tooloutput{structured result}
\begin{codebox}
{
    "analysis": "The image shows a red car on a curvy road with green hills and a blue sky with clouds in the background. ... The task is to identify the color of the car. Regions (D) and (E) are relevant as they show the car, but region (D) contains the full view of the car, making region (E) unnecessary.",
    "patches": 
     {
         "path": "car_bottom-right_zoomed_2x.png",
         "description": "The bottom-right region of the image: car.png."
     }
}
\end{codebox}

\end{textcolorbox}

\begin{textcolorbox}[Relevant Patch Zoomer Tool: Example 2]
\toolinput{query} What is the color of the car?

\toolinput{image} \texttt{"car.png"}

\vspace{0.2cm}
\includegraphics[width=0.3\linewidth]{tools/examples/car1.png}
\vspace{0.2cm}

\tooloutput{analysis}

The image depicts a red car driving on a curved road surrounded by green trees. Here's a breakdown of the five regions:

\quad (A) Top-left quarter: Contains part of the road and green trees in the background. No car is visible.

\quad (B) Top-right quarter: Includes trees in the background and part of the road. The rear portion of the car is partially visible.

\quad (C) Bottom-left quarter: Shows the road with no visible car.

\quad (D) Bottom-right quarter: Displays a clear view of the rear part of the red car, along with some of the road and green surroundings.

\quad (E) Center region: Contains the clearest and most focused view of the rear of the red car.
\\\\
To answer the question, ``What is the color of the car?'': The most relevant region is (E), as it contains the clearest depiction of the car's color without including unnecessary background or less visible parts of the car.
\\\\
\tooloutput{patch region} (E) center quarter

\tooloutput{patch image} \texttt{"car\_bottom-right\_zoomed\_2x\_center\_zoomed\_2x.png"}

\vspace{0.2cm}
\includegraphics[width=0.3\linewidth]{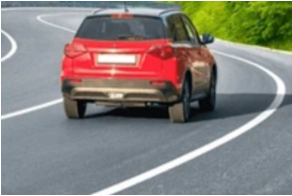}
\vspace{0.2cm}

\tooloutput{returned result}
\begin{codebox}
{
    "analysis": "The image depicts a red car driving on a curved road surrounded by green trees. ... To answer the question, 'What is the color of the car?': The most relevant region is (E), as it contains the clearest depiction of the car's color without including unnecessary background or less visible parts of the car.",
    "patches": 
     {
         "path": "car_bottom-right_zoomed_2x_center_zoomed_2x.png",
         "description": "The center region of the image: car_bottom-right_zoomed_2x.png."
     }
}
\end{codebox}

\end{textcolorbox}

\clearpage
\subsection{Text Detector Tool}
\label{app:text_detector_tool}
\begin{codecolorbox}[Text Detector Tool: Metadata]{python}
tool_name="Text_Detector_Tool",

tool_description="A tool that detects text in an image using EasyOCR.",

input_types={
    "image": "str - The path to the image file.",
    "languages": "list - A list of language codes for the OCR model.",
    "detail": "int - The level of detail in the output. Set to 0 for simpler output, 1 for detailed output."
},

output_type="list - A list of detected text blocks.",

demo_commands=[
    {
        "command": 'execution = tool.execute(image="path/to/image.png", languages=["en"])',
        "description": "Detect text in an image using the default language (English)."
    },
    {
        "command": 'execution = tool.execute(image="path/to/image.png", languages=["en", "de"])',
        "description": "Detect text in an image using multiple languages (English and German)."
    },
    {
        "command": 'execution = tool.execute(image="path/to/image.png", languages=["en"], detail=0)',
        "description": "Detect text in an image with simpler output (text without coordinates and scores)."
    },
],

user_metadata={
    "frequently_used_language": {
        "ch_sim": "Simplified Chinese",
        "ch_tra": "Traditional Chinese",
        "de": "German",
        "en": "English",
        "es": "Spanish",
        "fr": "French",
        "hi": "Hindi",
        "ja": "Japanese",
    }
}
\end{codecolorbox}

\begin{textcolorbox}[Text Detector Tool: Example 1]
\toolinput{image} \texttt{"english.png"}

\vspace{0.2cm}
\includegraphics[width=0.7\linewidth]{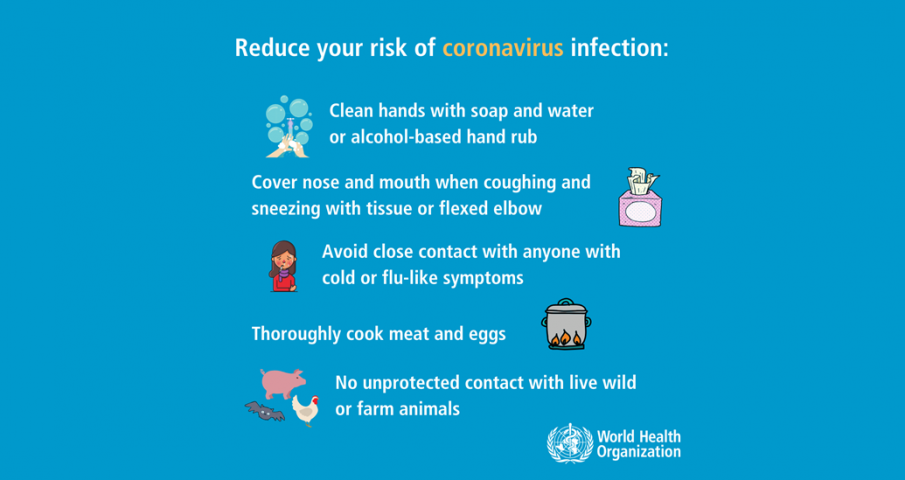}
\vspace{0.2cm}

\tooloutput{detected text}
{\footnotesize
\begin{codebox}
[[[[231, 32], [674, 32], [674, 64], [231, 64]], "Reduce your risk of coronavirus infection:", 0.84], 
[[[326, 98], [598, 98], [598, 124], [326, 124]], "Clean hands with soap and water", 0.96], 
[[[328, 124], [542, 124], [542, 148], [328, 148]], "or alcohol-based hand rub", 0.89], 
[[[246, 169], [595, 169], [595, 196], [246, 196]], "Cover nose and mouth when coughing and", 1.0], 
[[[245, 194], [546, 194], [546, 222], [245, 222]], "sneezing with tissue or flexed elbow", 0.96], 
[[[320, 240], [624, 240], [624, 266], [320, 266]], "Avoid close contact with anyone with", 0.86], 
[[[318, 266], [528, 266], [528, 292], [318, 292]], "cold or flu-like symptoms", 0.77], 
[[[248, 322], [510, 322], [510, 348], [248, 348]], "Thoroughly cook meat and eggs", 0.72], 
[[[332, 370], [640, 370], [640, 396], [332, 396]], "No unprotected contact with live wild", 0.83], 
[[[334, 396], [464, 396], [464, 420], [334, 420]], "or farm animals", 0.72],
[[[595, 427], [683, 427], [683, 447], [595, 447]], "World Health", 1.0], 
[[[595, 445], [685, 445], [685, 463], [595, 463]], "Organization", 1.0]]
\end{codebox}
}
\tooloutput{structured result}
{\footnotesize
\begin{codebox}
{
    "[[[[231, 32], [674, 32], [674, 64], [231, 64]], `Reduce your risk of coronavirus infection:`, 0.84], [[[326, 98], [598, 98], [598, 124], [326, 124]], `Clean hands with soap and water`, 0.96], [[[328, 124], [542, 124], [542, 148], [328, 148]], `or alcohol-based hand rub`, 0.89], [[[246, 169], [595, 169], [595, 196], [246, 196]], `Cover nose and mouth when coughing and`, 1.0], [[[245, 194], [546, 194], [546, 222], [245, 222]], `sneezing with tissue or flexed elbow`, 0.96], [[[320, 240], [624, 240], [624, 266], [320, 266]], `Avoid close contact with anyone with`, 0.86], [[[318, 266], [528, 266], [528, 292], [318, 292]], `cold or flu-like symptoms`, 0.77], [[[248, 322], [510, 322], [510, 348], [248, 348]], `Thoroughly cook meat and eggs`, 0.72], [[[332, 370], [640, 370], [640, 396], [332, 396]], `No unprotected contact with live wild`, 0.83], [[[334, 396], [464, 396], [464, 420], [334, 420]], `or farm animals`, 0.72], [[[595, 427], [683, 427], [683, 447], [595, 447]], `World Health`, 1.0], [[[595, 445], [685, 445], [685, 463], [595, 463]], `Organization`, 1.0]]"
}
\end{codebox}
}

\end{textcolorbox}

\begin{textcolorbox}[Text Detector Tool: Example 2]
\toolinput{image} \texttt{"chinese.jpg"}

\vspace{0.2cm}
\includegraphics[width=0.4\linewidth]{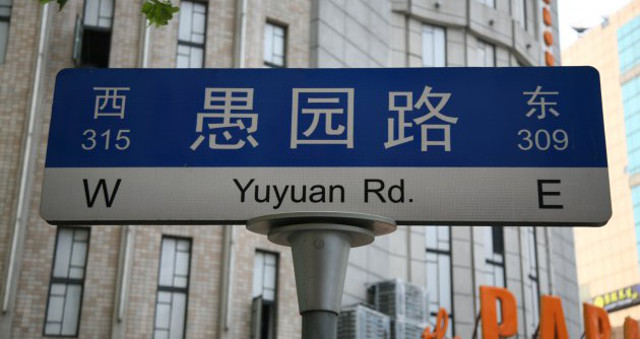}
\vspace{0.2cm}

\tooloutput{detected text}
\begin{codebox}
{
    [([[86, 80], [134, 80], [134, 128], [86, 128]], "(*@\begin{CJK}{UTF8}{gbsn}西\end{CJK}@*)", 0.81), 
    ([[189, 75], [469, 75], [469, 165], [189, 165]], "(*@\begin{CJK}{UTF8}{gbsn}愚园路\end{CJK}@*)", 0.96), 
    ([[517, 81], [565, 81], [565, 123], [517, 123]], "(*@\begin{CJK}{UTF8}{gbsn}东\end{CJK}@*)", 0.99), 
    ([[78, 126], [136, 126], [136, 156], [78, 156]], "315", 1.0), 
    ([[514, 126], [574, 126], [574, 156], [514, 156]], "309", 1.0), 
    ([[79, 173], [125, 173], [125, 213], [79, 213]], "W", 0.32), 
    ([[226, 170], [414, 170], [414, 220], [226, 220]], "Yuyuan Rd.", 0.86), 
    ([[529, 173], [569, 173], [569, 213], [529, 213]], "E", 0.56)]
}
\end{codebox}
\tooloutput{structured result}
\begin{codebox}
{
    "[[[[86, 80], [134, 80], [134, 128], [86, 128]], '\u897f', 0.81], [[[189, 75], [469, 75], [469, 165], [189, 165]], '\u611a\u56ed\u8def', 0.96], [[[517, 81], [565, 81], [565, 123], [517, 123]], '\u4e1c', 0.99], [[[78, 126], [136, 126], [136, 156], [78, 156]], '315', 1.0], [[[514, 126], [574, 126], [574, 156], [514, 156]], '309', 1.0], [[[79, 173], [125, 173], [125, 213], [79, 213]], 'W', 0.32], [[[226, 170], [414, 170], [414, 220], [226, 220]], 'Yuyuan Rd\u3002', 0.86], [[[529, 173], [569, 173], [569, 213], [529, 213]], 'E', 0.56]]"
}
\end{codebox}

\end{textcolorbox}

\clearpage
\subsection{URL Text Extractor Tool}
\label{app:url_text_extractor_tool}
\begin{codecolorbox}[URL Text Extractor Tool: Metadata]{python}
tool_name="URL_Text_Extractor_Tool",

tool_description="A tool that extracts all text from a given URL.",

input_types={
    "url": "str - The URL from which to extract text.",
},

output_type="dict - A dictionary containing the extracted text and any error messages.",

demo_commands=[
    {
        "command": 'execution = tool.execute(url="https://example.com")',
        "description": "Extract all text from the example.com website."
    },
    {
        "command": 'execution = tool.execute(url="https://en.wikipedia.org/wiki/Python_(programming_language)")',
        "description": "Extract all text from the Wikipedia page about Python programming language."
    },
]
\end{codecolorbox}

\begin{textcolorbox}[URL Text Extractor Tool: Example 1]
\toolinput{query}
\url{https://example.com}
\\\\
\tooloutput{text}

Example Domain

Example Domain

This domain is for use in illustrative examples in documents. You may use this
    domain in literature without prior coordination or asking for permission.
    
More information...
\\\\
\tooloutput{structured result}
\begin{codebox}
{
    "url": "https://example.com",
    "extracted_text": "Example Domain\nExample Domain\nThis domain is for use in illustrative examples in documents. You may use this\n    domain in literature without prior coordination or asking for permission.\nMore information..."
}
\end{codebox}

\end{textcolorbox}

\begin{textcolorbox}[URL Text Extractor Tool: Example 2]
\toolinput{query}
\url{https://arxiv.org/abs/1706.03762}
\\\\
\tooloutput{text}

[1706.03762] Attention Is All You Need

...

[Submitted on 12 Jun 2017 (
v1
), last revised 2 Aug 2023 (this version, v7)]

Title:
Attention Is All You Need

Authors:
Ashish Vaswani
,
Noam Shazeer
,
Niki Parmar
,
Jakob Uszkoreit
,
Llion Jones
,
Aidan N. Gomez
,
Lukasz Kaiser
,
Illia Polosukhin

...

Abstract:
The dominant sequence transduction models are based on complex recurrent or convolutional neural networks in an encoder-decoder configuration. The best performing models also connect the encoder and decoder through an attention mechanism. We propose a new simple network architecture, the Transformer, based solely on attention mechanisms, dispensing with recurrence and convolutions entirely. Experiments on two machine translation tasks show these models to be superior in quality while being more parallelizable and requiring significantly less time to train. Our model achieves 28.4 BLEU on the WMT 2014 English-to-German translation task, improving over the existing best results, including ensembles by over 2 BLEU. On the WMT 2014 English-to-French translation task, our model establishes a new single-model state-of-the-art BLEU score of 41.8 after training for 3.5 days on eight GPUs, a small fraction of the training costs of the best models from the literature. We show that the Transformer generalizes well to other tasks by applying it successfully to English constituency parsing both with large and limited training data.

...
\\\\
\tooloutput{structured result}
\begin{codebox}
{
    "url": "https://arxiv.org/abs/1706.03762",
    "extracted_text": "[1706.03762] Attention Is All You Need\n\n...\n\n[Submitted on 12 Jun 2017 (\nv1\n), last revised 2 Aug 2023 (this version, v7)]\n\nTitle:\nAttention Is All You Need\n\nAuthors:\nAshish Vaswani\n,\nNoam Shazeer\n,\nNiki Parmar\n,\nJakob Uszkoreit\n,\nLlion Jones\n,\nAidan N. Gomez\n,\nLukasz Kaiser\n,\nIllia Polosukhin\n\n...\n\nAbstract:\nThe dominant sequence transduction models are based on complex recurrent or convolutional neural networks in an encoder-decoder configuration. The best performing models also connect the encoder and decoder through an attention mechanism. We propose a new simple network architecture, the Transformer, based solely on attention mechanisms, dispensing with recurrence and convolutions entirely. Experiments on two machine translation tasks show these models to be superior in quality while being more parallelizable and requiring significantly less time to train. Our model achieves 28.4 BLEU on the WMT 2014 English-to-German translation task, improving over the existing best results, including ensembles by over 2 BLEU. On the WMT 2014 English-to-French translation task, our model establishes a new single-model state-of-the-art BLEU score of 41.8 after training for 3.5 days on eight GPUs, a small fraction of the training costs of the best models from the literature. We show that the Transformer generalizes well to other tasks by applying it successfully to English constituency parsing both with large and limited training data.\n\n..."
}
\end{codebox}

\end{textcolorbox}

\clearpage
\subsection{Wikipedia Knowledge Searcher Tool}
\label{app:wikipedia_knowledge_searcher_tool}
\begin{codecolorbox}[Wikipedia Knowledge Searcher Tool: Metadata]{python}
tool_name="Wikipedia_Knowledge_Searcher_Tool",

tool_description="A tool that searches Wikipedia and returns web text based on a given query.",

input_types={
    "query": "str - The search query for Wikipedia.",            },

output_type="dict - A dictionary containing the search results, extracted text, and any error messages.",

demo_commands=[
    {
        "command": 'execution = tool.execute(query="Python programming language")',
        "description": "Search Wikipedia for information about Python programming language."
    },
    {
        "command": 'execution = tool.execute(query="Artificial Intelligence")',
        "description": "Search Wikipedia for information about Artificial Intelligence"
    },
    {
        "command": 'execution = tool.execute(query="Theory of Relativity")',
        "description": "Search Wikipedia for the full article about the Theory of Relativity."
    },
]
\end{codecolorbox}

\begin{textcolorbox}[Wikipedia Knowledge Searcher Tool: Example 1]
\toolinput{query}
kidney
\\\\
\tooloutput{type}

Search results for `kidney':

1. Kidney
2. Kidney disease
3. Kidney failure
4. Kidney dialysis
5. Kidney transplantation
6. Kidney bean
7. Kidney cancer
8. Nephrology
9. Ectopic kidney
10. Kidney dish
\\\\
Extracted text:

In humans, the kidneys are two reddish-brown bean-shaped blood-filtering organs that are a multilobar, multipapillary form of mammalian kidneys, usually without signs of external lobulation. They are located on the left and right in the retroperitoneal space, and in adult humans are about 12 centimetres (4+1⁄2 inches) in length. They receive blood from the paired renal arteries; blood exits into the paired renal veins. Each kidney is attached to a ureter, a tube that carries excreted urine to the bladder.
...
\\\\
\tooloutput{structured result}
\begin{codebox}
{
    "output": "Search results for 'kidney':\n1. Kidney\n2. Kidney disease\n3. Kidney failure\n4. Kidney dialysis\n5. Kidney transplantation\n6. Kidney bean\n7. Kidney cancer\n8. Nephrology\n9. Ectopic kidney\n10. Kidney dish\n\nExtracted text:\nIn humans, the kidneys are two reddish-brown bean-shaped blood-filtering organs that are a multilobar, multipapillary form of mammalian kidneys, usually without signs of external lobulation. They are located on the left and right in the retroperitoneal space, and in adult humans are about 12 centimetres (4+1\u20442 inches) in length. They receive blood from the paired renal arteries; blood exits into the paired renal veins. Each kidney is attached to a ureter, a tube that carries excreted urine to the bladder. ..."
}
\end{codebox}

\end{textcolorbox}

\clearpage
\section{Additional Tool Cards in \model}
\label{app:additional_toolcards}

% \clearpage
\subsection{Object Detector Tool}
\label{app:object_detector_tool}
\begin{codecolorbox}[Object Detector Tool: Metadata]{python}
tool_name="Object_Detector_Tool",

tool_description="A tool that detects objects in an image using the Grounding DINO model and saves individual object images with empty padding.",

input_types={
    "image": "str - The path to the image file.",
    "labels": "list - A list of object labels to detect.",
    "threshold": "float - The confidence threshold for detection (default: 0.35).",
    "model_size": "str - The size of the model to use ('tiny' or 'base', default: 'tiny').",
    "padding": "int - The number of pixels to add as empty padding around detected objects (default: 20)."
},

output_type="list - A list of detected objects with their scores, bounding boxes, and saved image paths.",

demo_commands=[
    {
        "command": 'execution = tool.execute(image="path/to/image.png", labels=["baseball", "basket"])',
        "description": "Detect baseball and basket in an image, save the detected objects with default empty padding, and return their paths."
    },
    {
        "command": 'execution = tool.execute(image="path/to/image.png", labels=["car", "person"], threshold=0.5, model_size="base", padding=15)',
        "description": "Detect car and person in an image using the base model, save the detected objects with 15 pixels of empty padding, and return their paths."
    }
],

user_metadata={
    "limitation": "The model may not always detect objects accurately, and its performance can vary depending on the input image and the associated labels. It typically struggles with detecting small objects, objects that are uncommon, or objects with limited or specific attributes. For improved accuracy or better detection in certain situations, consider using supplementary tools or image processing techniques to provide additional information for verification."
}
\end{codecolorbox}

\begin{textcolorbox}[Object Detector Tool: Example 1]
\toolinput{image} \texttt{"example.png"}

\vspace{0.2cm}
\includegraphics[width=0.4\linewidth]{tools/examples/mathvista_113.png}
\vspace{0.2cm}

\toolinput{label} \texttt{["baseball"]}
\\\\
\tooloutput{detected objects}
\\\\
\includegraphics[width=0.05\linewidth]{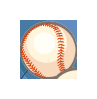}
\includegraphics[width=0.05\linewidth]{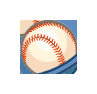}
\includegraphics[width=0.05\linewidth]{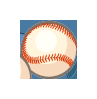}
\includegraphics[width=0.05\linewidth]{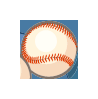}
\includegraphics[width=0.05\linewidth]{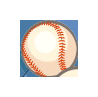}

\includegraphics[width=0.05\linewidth]{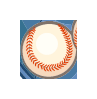}
\includegraphics[width=0.05\linewidth]{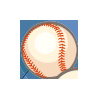}
\includegraphics[width=0.05\linewidth]{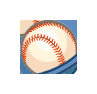}
\includegraphics[width=0.05\linewidth]{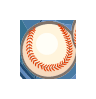}
\includegraphics[width=0.05\linewidth]{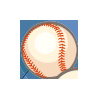}

\includegraphics[width=0.05\linewidth]{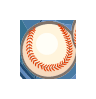}
\includegraphics[width=0.05\linewidth]{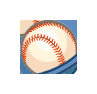}
\includegraphics[width=0.05\linewidth]{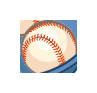}
\includegraphics[width=0.05\linewidth]{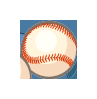}
\includegraphics[width=0.05\linewidth]{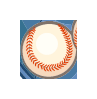}

\includegraphics[width=0.05\linewidth]{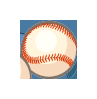}
\includegraphics[width=0.05\linewidth]{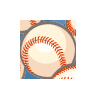}
\includegraphics[width=0.05\linewidth]{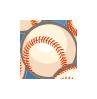}
\includegraphics[width=0.05\linewidth]{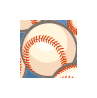}
\includegraphics[width=0.05\linewidth]{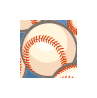}
    
\tooloutput{structured result}

\begin{codebox}
[
    {
        "label": "baseball",
        "confidence score": 0.69,
        "box": [558, 48, 615, 107],
        "saved_image_path": "example_baseball_1.png"
    },
    {
        "label": "baseball",
        "confidence score": 0.69,
        "box": [614, 137, 671, 191],
        "saved_image_path": "example_baseball_2.png"
    },
    ...
    {
        "label": "baseball",
        "confidence score": 0.65,
        "box": [86, 335, 143, 393],
        "saved_image_path": "example_baseball_19.png"
    },
    {
        "label": "baseball",
        "confidence score": 0.63,
        "box": [336, 95, 393, 153],
        "saved_image_path": "example_baseball_20.png"
    }
]
\end{codebox}

\end{textcolorbox}

\begin{textcolorbox}[Object Detector Tool: Example 2]
\toolinput{image} \texttt{"example.png"}

\vspace{0.2cm}
\includegraphics[width=0.4\linewidth]{tools/examples/mathvista_113.png}
\vspace{0.2cm}                                                                   

\toolinput{label} \texttt{["basket"]}
\\\\
\tooloutput{detected objects}
\vspace{0.2cm} 

\includegraphics[width=0.14\linewidth]{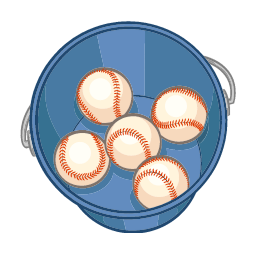}    
\includegraphics[width=0.14\linewidth]{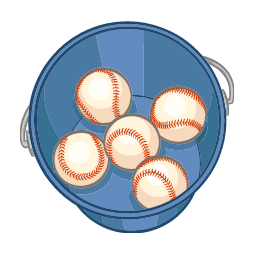}    
\includegraphics[width=0.14\linewidth]{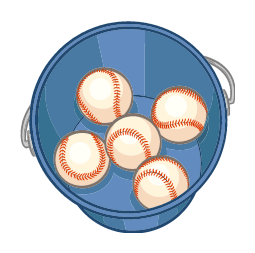}
\includegraphics[width=0.14\linewidth]{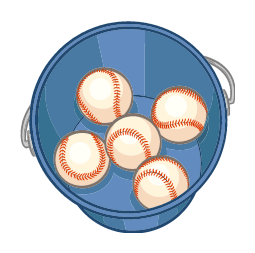}
\\\\
\tooloutput{structured result}

\begin{codebox}
[
    {
        "label": "basket",
        "confidence score": 0.59,
        "box": [252, 2, 468, 215],
        "saved_image_path": "example_basket_1.png"
    },
    {
        "label": "basket",
        "confidence score": 0.55,
        "box": [503, 2, 717, 215],
        "saved_image_path": "example_basket_2.png"
    },
    {
        "label": "basket",
        "confidence score": 0.54,
        "box": [2, 2, 217, 215],
        "saved_image_path": "example_basket_3.png"
    },
    {
        "label": "basket",
        "confidence score": 0.5,
        "box": [2, 242, 217, 455],
        "saved_image_path": "example_basket_4.png"
    }
]
\end{codebox}
\end{textcolorbox}

\clearpage
\subsection{Advanced Object Detector Tool}
\label{app:advanced_object_detector_tool}
\begin{codecolorbox}[Advanced Object Detector Tool: Metadata]{python}
tool_name="Advanced Object Detector Tool",

tool_description="A tool that detects objects in an image using the Grounding DINO-X model and saves individual object images with empty padding.",

input_types={
    "image": "str - The path to the image file.",
    "labels": "list - A list of object labels to detect.",
    "threshold": "float - The confidence threshold for detection (default: 0.35).",
    "padding": "int - The number of pixels to add as empty padding around detected objects (default: 20)."
},

output_type="list - A list of detected objects with their scores, bounding boxes, and saved image paths.",

demo_commands=[
    {
        "command": 'execution = tool.execute(image="path/to/image.png", labels=["baseball", "basket"])',
        "description": "Detect baseball and basket in an image, save the detected objects with default empty padding, and return their paths."
    },
    {
        "command": 'execution = tool.execute(image="path/to/image.png", labels=["car", "person"], threshold=0.5, model_size="base", padding=15)',
        "description": "Detect car and person in an image using the base model, save the detected objects with 15 pixels of empty padding, and return their paths."
    }
],

user_metadata={
    "limitation": "The model may not always detect objects accurately, and its performance can vary depending on the input image and the associated labels. It typically struggles with detecting small objects, objects that are uncommon, or objects with limited or specific attributes. For improved accuracy or better detection in certain situations, consider using supplementary tools or image processing techniques to provide additional information for verification."
}
\end{codecolorbox}

\begin{textcolorbox}[Advanced Object Detector Tool: Example 1]
\toolinput{image} \texttt{"example.png"}

\vspace{0.2cm}
\includegraphics[width=0.4\linewidth]{tools/examples/mathvista_113.png}
\vspace{0.2cm}

\toolinput{label} \texttt{["baseball"]}
\\\\
\tooloutput{detected objects}

\includegraphics[width=0.05\linewidth]{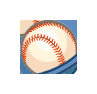}
\includegraphics[width=0.05\linewidth]{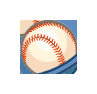}
\includegraphics[width=0.05\linewidth]{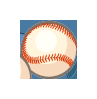}
\includegraphics[width=0.05\linewidth]{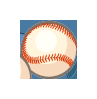}
\includegraphics[width=0.05\linewidth]{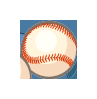}

\includegraphics[width=0.05\linewidth]{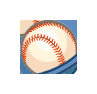}
\includegraphics[width=0.05\linewidth]{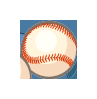}
\includegraphics[width=0.05\linewidth]{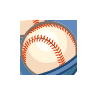}
\includegraphics[width=0.05\linewidth]{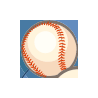}
\includegraphics[width=0.05\linewidth]{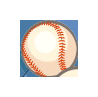}

\includegraphics[width=0.05\linewidth]{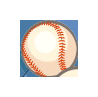}
\includegraphics[width=0.05\linewidth]{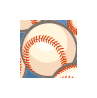}
\includegraphics[width=0.05\linewidth]{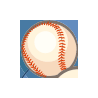}
\includegraphics[width=0.05\linewidth]{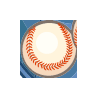}
\includegraphics[width=0.05\linewidth]{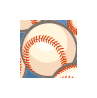}

\includegraphics[width=0.05\linewidth]{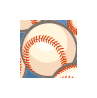}
\includegraphics[width=0.05\linewidth]{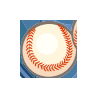}
\includegraphics[width=0.05\linewidth]{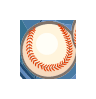}
\includegraphics[width=0.05\linewidth]{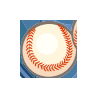}
\includegraphics[width=0.05\linewidth]{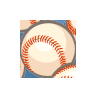}
    
\tooloutput{structured result}

\begin{codebox}
[
    {
        "label": "baseball",
        "confidence score": 0.73,
        "box": [614, 137, 671, 191],
        "saved_image_path": "example_baseball_1.png"
    },
    {
        "label": "baseball",
        "confidence score": 0.73,
        "box": [114, 377, 171, 431],
        "saved_image_path": "example_baseball_2.png"
    },
    ...
    {
        "label": "baseball",
        "confidence score": 0.66,
        "box": [535, 111, 592, 169],
        "saved_image_path": "example_baseball_19.png"
    },
    {
        "label": "baseball",
        "confidence score": 0.66,
        "box": [337, 95, 393, 153],
        "saved_image_path": "example_baseball_20.png"
    }
]
\end{codebox}

\end{textcolorbox}

\clearpage
\subsection{Nature News Fetcher Tool}
\label{app:nature_news_fetcher_tool}
\begin{codecolorbox}[Nature News Fetcher Tool: Metadata]{python}
tool_name="Nature_News_Fetcher_Tool",

tool_description="A tool that fetches the latest news articles from Nature.",

input_types={
    "num_articles": "int - The number of articles to fetch (default: 100).",
    "max_pages": "int - The maximum number of pages to fetch (default: 5).",
},

output_type="list - A list of dictionaries containing information about the latest Nature news articles.",

demo_commands=[
    {
        "command": 'execution = tool.execute()',
        "description": "Fetch the latest 100 news articles from Nature."
    },
    {
        "command": 'execution = tool.execute(num_articles=50, max_pages=3)',
        "description": "Fetch the latest 50 news articles from Nature, searching up to 3 pages."
    },
]
\end{codecolorbox}

\begin{textcolorbox}[Nature News Fetcher Tool: Example 1]
\tooloutput{article 1, search date: Jan 14, 2025}
\\\\
\textbf{Title:} Has Bluesky replaced X for scientists? Take Nature’s poll

\textbf{URL:} \url{https://www.nature.com/articles/d41586-025-00037-y}

\textbf{Description:} The research community has flocked to the social-media platform Bluesky. Tell us about your experien...

\textbf{Authors:} No authors found

\textbf{Date:} 2025-01-14

\textbf{Image URL:} 

\url{https://media.springernature.com/w290h158/magazine-assets/d41586-025-00037-y/d41586-025-00037-y_50440812.jpg}
\\\\
\tooloutput{article 2, search date: Jan 14, 2025}
\\\\
\textbf{Title:} How should we test AI for human-level intelligence? OpenAI’s o3 electrifies quest

\textbf{URL:} \url{https://www.nature.com/articles/d41586-025-00110-6}

\textbf{Description:} Experimental model’s record-breaking performance on science and maths tests wows researchers....

\textbf{Authors:} Nicola Jones

\textbf{Date:} 2025-01-14

\textbf{Image URL:} 

\url{https://media.springernature.com/w290h158/magazine-assets/d41586-025-00110-6/d41586-025-00110-6_50469004.jpg}
\\\\
\tooloutput{structured result}
\begin{codebox}
[
    {
        "title": "Has Bluesky replaced X for scientists? Take Nature\u2019s poll",
        "url": "https://www.nature.com/articles/d41586-025-00037-y",
        "description": "The research community has flocked to the social-media platform Bluesky. Tell us about your experience.",
        "authors": [
            "No authors found"
        ],
        "date": "2025-01-14",
        "image_url": "https://media.springernature.com/w290h158/magazine-assets/d41586-025-00037-y/d41586-025-00037-y_50440812.jpg"
    },
    {
        "title": "How should we test AI for human-level intelligence? OpenAI\u2019s o3 electrifies quest",
        "url": "https://www.nature.com/articles/d41586-025-00110-6",
        "description": "Experimental model\u2019s record-breaking performance on science and maths tests wows researchers.",
        "authors": [
            "Nicola Jones"
        ],
        "date": "2025-01-14",
        "image_url": "https://media.springernature.com/w290h158/magazine-assets/d41586-025-00110-6/d41586-025-00110-6_50469004.jpg"
    }
]
\end{codebox}

\end{textcolorbox}

\clearpage
\section{Case Study Examples}
\label{app:exp_examples}

\subsection{Demonstration Example}
\label{app:demo_example}

Below is an example from MathVista \cite{lu2024mathvista}, which demonstrates the detailed steps in our \model framework. The base \gpt model fails to provide a correct answer due to its limited visual perception of abstract scenes.

%%%%%%%%%%%%%%%%%%%%%%%%%%%%%%%%%%%%%%%%%%%%%%%%%%%%%%%%%%%%%%%%
\begin{querybox}[The Query]
\begin{minipage}{0.3\linewidth}
    \includegraphics[width=1\linewidth]{tools/examples/mathvista_113.png}
\end{minipage}
\hspace{5pt}
\begin{minipage}{0.55\linewidth}
    \textbf{Image path:} \texttt{baseball.png}
    
    \textbf{Question:} How many baseballs are there?
    \\\\
    \textbf{Ground truth:} 20
\end{minipage}
\end{querybox}
%%%%%%%%%%%%%%%%%%%%%%%%%%%%%%%%%%%%%%%%%%%%%%%%%%%%%%%%%%%%%%%%

%%%%%%%%%%%%%%%%%%%%%%%%%%%%%%%%%%%%%%%%%%%%%%%%%%%%%%%%%%%%%%%%
\begin{querybox}[Wrong Prediction from \gpt]
\red{There are a total of 19 baseballs across the four buckets:}

- \red{The first bucket (top left) contains 4 baseballs.}

- The second bucket (top right) contains 5 baseballs.

- \red{The third bucket (bottom left) contains 4 baseballs.}

- \red{The fourth bucket (bottom right) contains 6 baseballs.}
\\\\
\red{Adding them together: 4 + 5 + 4 + 6 = 19 baseballs.}
\end{querybox}
%%%%%%%%%%%%%%%%%%%%%%%%%%%%%%%%%%%%%%%%%%%%%%%%%%%%%%%%%%%%%%%%

In contrast, \model decomposes the query into two steps by calling two vision tools to enhance visual perception. First, it calls the \texttt{Image\_Captioner\_Tool} to provide a general description of the associated image. Then, it uses the \texttt{Object\_Detector\_Tool} to detect and count the baseballs. These two steps provide consistent information about the number of baseballs, thus resulting in a correct answer.

%%%%%%%%%%%%%%%%%%%%%%%%%%%%%%%%%%%%%%%%%%%%%%%%%%%%%%%%%%%%%%%%
\begin{plannerbox}[Step 0: Initial Plan from Query Analyzer]

\textbf{Concise summary:}
The query asks, ``How many baseballs are there?'' accompanied by an image showing several buckets containing baseballs. The objective is to determine the total number of baseballs present in the image.
\\\\
\textbf{Required skills:}

1. Image Analysis: Understanding and interpreting visual content. \\
2. Tool Operation: Ability to operate and execute commands using the provided tools. \\
3. Critical Thinking: Evaluating tool outputs and making decisions based on them.
\\\\
\textbf{Relevant tools:}

1. \texttt{Image\_Captioner\_Tool}: Used to generate a description of the image, which can provide context and identify objects present. \\
2. \texttt{Object\_Detector\_Tool}: Used to detect and count the number of baseballs in the image, providing specific object identification and quantification.
\\\\
\textbf{Additional considerations:}
Consider the limitations of each tool, such as potential inaccuracies in complex scenes or object detection. Verify results if possible, and be aware of the need for potential manual verification or supplementary tools for precise counting.
\end{plannerbox}
%%%%%%%%%%%%%%%%%%%%%%%%%%%%%%%%%%%%%%%%%%%%%%%%%%%%%%%%%%%%%%%%

%%%%%%%%%%%%%%%%%%%%%%%%%%%%%%%%%%%%%%%%%%%%%%%%%%%%%%%%%%%%%%%%
\begin{plannerbox}[Step 1: Action from Action Predictor]

\textbf{Justification:}
The \texttt{Image\_Captioner\_Tool} is the best choice for the first step because it provides a general description of the image, which can help identify the presence of baseballs and other relevant objects. This initial step is crucial for understanding the context of the image before proceeding to count specific objects using the \texttt{Object\_Detector\_Tool}.
\\\\
\textbf{Context:}
Image path: \texttt{baseball.png}
\\\\
\textbf{Sub-goal:}
Generate a detailed description of the image located at \texttt{"baseball.png"} to identify the presence of baseballs and other relevant objects.
\\\\
\textbf{Tool name:} \texttt{Image\_Captioner\_Tool}
\end{plannerbox}
%%%%%%%%%%%%%%%%%%%%%%%%%%%%%%%%%%%%%%%%%%%%%%%%%%%%%%%%%%%%%%%%

%%%%%%%%%%%%%%%%%%%%%%%%%%%%%%%%%%%%%%%%%%%%%%%%%%%%%%%%%%%%%%%%
\begin{executorbox}[Step 1: Command from Command Generator]

\textbf{Analysis:}
The task requires two steps: first, using the \texttt{Image\_Captioner\_Tool} to generate a caption for the image, and second, using the \texttt{Object\_Detector\_Tool} to count the number of baseballs. The current focus is on the first step, which involves generating a descriptive caption for the image located at \texttt{'baseball.png'}. The \texttt{Image\_Captioner\_Tool} requires an image path and optionally a prompt. The default prompt is sufficient for generating a general description of the image.
\\\\
\textbf{Explanation:}
The command uses the \texttt{Image\_Captioner\_Tool} to generate a caption for the image. The image path is provided as \texttt{'baseball.png'}, and the default prompt is used to describe the image. This step is necessary to provide context for the subsequent object detection task.
\\\\
\textbf{Command:}
\command{execution = tool.execute(image="baseball.png")}
\end{executorbox}
%%%%%%%%%%%%%%%%%%%%%%%%%%%%%%%%%%%%%%%%%%%%%%%%%%%%%%%%%%%%%%%%

%%%%%%%%%%%%%%%%%%%%%%%%%%%%%%%%%%%%%%%%%%%%%%%%%%%%%%%%%%%%%%%%
\begin{executorbox}[Step 1: Result from Command Executor]
\textbf{Result:}
The image shows four blue buckets, each containing five baseballs. The buckets are arranged in a grid with three on the top row and one on the bottom left. Each bucket is depicted from a top-down perspective, and the baseballs inside are white with red stitching. The buckets have handles on the sides. The background is plain white, emphasizing the buckets and baseballs.
\end{executorbox}
%%%%%%%%%%%%%%%%%%%%%%%%%%%%%%%%%%%%%%%%%%%%%%%%%%%%%%%%%%%%%%%%

%%%%%%%%%%%%%%%%%%%%%%%%%%%%%%%%%%%%%%%%%%%%%%%%%%%%%%%%%%%%%%%%
\begin{plannerbox}[Step 1: Verification from Context Verifier]
\textbf{Completeness:}
The memory addresses the query by using the \texttt{Image\_Captioner\_Tool} to describe the image, which mentions that there are four buckets, each containing five baseballs. This provides a total count of 20 baseballs. However, the \texttt{Object\_Detector\_Tool} has not been used yet to verify this count, which is a requirement of the task.
\\\\
\textbf{Unused Tools:}
The \texttt{Object\_Detector\_Tool} is available and should be used to count the baseballs as per the task description. This tool is specifically designed to detect and count objects, which is necessary to fulfill the query accurately.
\\\\
\textbf{Inconsistencies:}
There are no inconsistencies in the information provided by the \texttt{Image\_Captioner\_Tool}. However, the task requires verification using the \texttt{Object\_Detector\_Tool}, which has not been done.
\\\\
\textbf{Verification needs:}
The count of baseballs needs verification using the \texttt{Object\_Detector\_Tool} due to the task's explicit requirement to use this tool for counting.
\end{plannerbox}
%%%%%%%%%%%%%%%%%%%%%%%%%%%%%%%%%%%%%%%%%%%%%%%%%%%%%%%%%%%%%%%%

%%%%%%%%%%%%%%%%%%%%%%%%%%%%%%%%%%%%%%%%%%%%%%%%%%%%%%%%%%%%%%%%
\begin{plannerbox}[Step 1: Verification from Context Verifier (Continued)]
\textbf{Ambiguities:}
There are no ambiguities in the current description, but the task's requirement to use the \texttt{Object\_Detector\_Tool} suggests that the count should be confirmed using this tool.
\\\\
\textbf{Final determination:}
The memory is not complete as it lacks the use of the \texttt{Object\_Detector\_Tool} to count the baseballs, which is a critical part of the task. Therefore, additional tool usage is necessary to fulfill the query accurately.
\\\\
\textbf{Conclusion:} \command{\textbf{CONTINUE}}
\end{plannerbox}
%%%%%%%%%%%%%%%%%%%%%%%%%%%%%%%%%%%%%%%%%%%%%%%%%%%%%%%%%%%%%%%%

%%%%%%%%%%%%%%%%%%%%%%%%%%%%%%%%%%%%%%%%%%%%%%%%%%%%%%%%%%%%%%%%
\begin{plannerbox}[Step 2: Action from Action Predictor]

\textbf{Justification:}
The next logical step is to use the \texttt{Object\_Detector\_Tool} to count the number of baseballs in the image. The \texttt{Image\_Captioner\_Tool} has already provided a detailed description indicating the presence of baseballs in buckets. The \texttt{Object\_Detector\_Tool} is specifically designed to detect and quantify objects, making it the most suitable choice for counting the baseballs accurately. This step directly addresses the query's requirement to determine the number of baseballs, leveraging the tool's capability to identify and count specific objects within an image.
\\\\
\textbf{Context:}
Image path: \texttt{baseball.png}
\\\\
\textbf{Previous description:}
The image shows four blue buckets, each containing five baseballs. The buckets are arranged in a grid with two on the top row and two on the bottom row. Each bucket has a handle on the side, and the baseballs inside are white with red stitching. The background is plain white, emphasizing the buckets and baseballs.
\\\\
\textbf{Sub-goal:}
Use the \texttt{Object\_Detector\_Tool} to detect and count the number of baseballs in the image located at \texttt{"baseball.png"}.
\\\\
\textbf{Tool name:} \texttt{Object\_Detector\_Tool}
\end{plannerbox}
%%%%%%%%%%%%%%%%%%%%%%%%%%%%%%%%%%%%%%%%%%%%%%%%%%%%%%%%%%%%%%%%

%%%%%%%%%%%%%%%%%%%%%%%%%%%%%%%%%%%%%%%%%%%%%%%%%%%%%%%%%%%%%%%%
\begin{executorbox}[Step 2: Command from Command Generator]

\textbf{Analysis:}
The task requires detecting and counting the number of baseballs in a given image using the \texttt{Object\_Detector\_Tool}. The image path is provided, and the context suggests that the image contains baseballs. The tool's metadata indicates that it requires an image path and a list of labels to detect. The label relevant to our task is `baseball'. The tool also allows setting a confidence threshold, model size, and padding, but these are optional. For this task, we will use the default values for these optional parameters.
\\\\
\textbf{Explanation:}
The command is constructed to use the \texttt{Object\_Detector\_Tool} to detect baseballs in the specified image. We provide the image path and a list containing the label `baseball' to focus the detection on baseballs. The default values for threshold, model size, and padding are used, as they are not specified in the task requirements. This setup will allow the tool to detect and count the baseballs in the image.
\\\\
\textbf{Command:}
\command{execution = tool.execute(image="baseball.png", labels=["baseball"])}
\end{executorbox}
%%%%%%%%%%%%%%%%%%%%%%%%%%%%%%%%%%%%%%%%%%%%%%%%%%%%%%%%%%%%%%%%

%%%%%%%%%%%%%%%%%%%%%%%%%%%%%%%%%%%%%%%%%%%%%%%%%%%%%%%%%%%%%%%%
\begin{executorbox}[Step 2: Result from Command Executor]
\begin{codebox}
[
    {
        "label": "baseball",
        "confidence score": 0.69,
        "box": "(558, 48, 615, 107)",
        "saved_image_path": "solver_cache/baseball_1.png"
    },
    ...
    {
        "label": "baseball",
        "confidence score": 0.63,
        "box": "(336, 95, 393, 153)",
        "saved_image_path": "solver_cache/baseball_20.png"
    }
]
\end{codebox}
\end{executorbox}
%%%%%%%%%%%%%%%%%%%%%%%%%%%%%%%%%%%%%%%%%%%%%%%%%%%%%%%%%%%%%%%%

%%%%%%%%%%%%%%%%%%%%%%%%%%%%%%%%%%%%%%%%%%%%%%%%%%%%%%%%%%%%%%%%
\begin{plannerbox}[Step 2: Verification from Context Verifier]

\textbf{Completeness:}
The memory addresses the query by using both the \texttt{Image\_Captioner\_Tool} and the \texttt{Object\_Detector\_Tool}. The \texttt{Image\_Captioner\_Tool} provided a detailed description of the image, stating that there are four blue buckets, each containing five baseballs, which totals to 20 baseballs. This description aligns with the task's requirement to identify the presence of baseballs.
The \texttt{Object\_Detector\_Tool} was then used to count the baseballs, detecting 20 baseballs in the image. This matches the description provided by the \texttt{Image\_Captioner\_Tool}, confirming the count.
\\\\
\textbf{Unused Tools:}
All relevant tools have been used. The \texttt{Image\_Captioner\_Tool} provided context, and the \texttt{Object\_Detector\_Tool} confirmed the count of baseballs.
\\\\
\textbf{Inconsistencies:}
There are no inconsistencies between the outputs of the two tools. Both tools indicate the presence of 20 baseballs.
\\\\
\textbf{Verification Needs:}
The results from both tools are consistent, reducing the need for further verification. The \texttt{Object\_Detector\_Tool}'s output aligns with the \texttt{Image\_Captioner\_Tool}'s description, providing confidence in the accuracy of the count.
\\\\
\textbf{Ambiguities:}
There are no ambiguities in the results. The tools have provided clear and consistent information regarding the number of baseballs.
\\\\
\textbf{Final Determination:}
The memory is complete and accurate enough to generate the final output. The use of both tools has provided a consistent and verified count of the baseballs, fulfilling the query's requirements.
\\\\
\textbf{Conclusion:} \command{\textbf{STOP}}
\end{plannerbox}
%%%%%%%%%%%%%%%%%%%%%%%%%%%%%%%%%%%%%%%%%%%%%%%%%%%%%%%%%%%%%%%%

%%%%%%%%%%%%%%%%%%%%%%%%%%%%%%%%%%%%%%%%%%%%%%%%%%%%%%%%%%%%%%%%
\begin{answerbox}[Full Trajectory in the Context after Two Steps]
\begin{answercodebox}
"Action Step 1": {
    "tool_name": "Image_Captioner_Tool",
    "sub_goal": "Generate a descriptive caption for the image located at \"baseball.png\" to provide context for subsequent object detection.",
    "command": "execution = tool.execute(image=\"baseball.png\")",
    "result": [
        "The image shows four blue buckets, each containing five baseballs. The buckets are arranged in a grid with three on the top row and one on the bottom left. Each bucket is depicted from a top-down perspective, and the baseballs inside are white with red stitching. The buckets have handles on the sides. The background is plain white, emphasizing the buckets and baseballs."
    ]
},
"Action Step 2": {
    "tool_name": "Object_Detector_Tool",
    "sub_goal": "Use the Object_Detector_Tool to detect and count the number of baseballs in the image located at \"baseball.png\".",
    "command": "execution = tool.execute(image=\"baseball_113.png\", labels=[\"baseball\"])",
    "result": [
        [
            {
                "label": "baseball",
                "confidence score": 0.69,
                "box": "(558, 48, 615, 107)",
                "saved_image_path": "solver_cache/baseball_1.png"
            },
            ...
            {
                "label": "baseball",
                "confidence score": 0.63,
                "box": "(336, 95, 393, 153)",
                "saved_image_path": "solver_cache/0/baseball_20.png"
            }
        ]
    ]
}
\end{answercodebox}
\end{answerbox}
%%%%%%%%%%%%%%%%%%%%%%%%%%%%%%%%%%%%%%%%%%%%%%%%%%%%%%%%%%%%%%%%

The final answer is summarized by the solution summarizer in the planner, given the full trajectory in the context described above. This two-step flow overcomes the base \gpt model’s limitation in visual perception.

%%%%%%%%%%%%%%%%%%%%%%%%%%%%%%%%%%%%%%%%%%%%%%%%%%%%%%%%%%%%%%%%
\begin{answerbox}[Final Answer]
\textbf{Detailed analysis:}

\quad\textbf{Step 1:} \texttt{Image\_Captioner\_Tool}

\quad\quad\quad\quad Result: The image contains four blue buckets, each with five baseballs, arranged in a grid pattern.
   
\quad\textbf{Step 2}: \texttt{Object\_Detector\_Tool}

\quad\quad\quad\quad Result: Detected 20 baseballs with varying confidence scores.
\\\\
\textbf{Key findings:}
The image contains a total of 20 baseballs, distributed evenly across four buckets.
Each bucket contains five baseballs, as confirmed by both tools.
\\\\
\textbf{Answer to the query:}
The image shows four blue buckets, each containing five baseballs. Therefore, \textbf{there are a total of 20 baseballs.}
\\\\
\textbf{Additional insights:}
The consistent results from both tools reinforce the accuracy of the analysis.
The arrangement of the buckets and baseballs is clear and well-organized, aiding in accurate detection.
\end{answerbox}

\clearpage
\subsection{Example for Multi-step Mathematical Reasoning}

%%%%%%%%%%%%%%%%%%%%%%%%%%%%%%%%%%%%%%%%%%%%%%%%%%%%%%%%%%%%%%%%
\begin{querybox}[The Query]
\textbf{Question:}
Using the numbers [1, 1, 6, 9], create an expression that equals 24. You must use basic arithmetic operations (+, -, ×, /) and parentheses. 

\textbf{Example:} for [1, 2, 3, 4], one solution is (1+2+3)×4.
\end{querybox}
%%%%%%%%%%%%%%%%%%%%%%%%%%%%%%%%%%%%%%%%%%%%%%%%%%%%%%%%%%%%%%%%

%%%%%%%%%%%%%%%%%%%%%%%%%%%%%%%%%%%%%%%%%%%%%%%%%%%%%%%%%%%%%%%%
\begin{querybox}[Wrong Prediction from \gpt]
One possible expression using the numbers [1, 1, 6, 9] to make 24 is:
\[
\red{(9 + 1) \times (6 / 2)}
\]
Breaking it down:

\quad - \( 9 + 1 = 10 \)

\quad - \( 6 / 2 = 3 \)

\quad - \red{\( 10 \times 3 = 24 \)}

This satisfies the conditions using basic arithmetic operations and parentheses. 
\end{querybox}
%%%%%%%%%%%%%%%%%%%%%%%%%%%%%%%%%%%%%%%%%%%%%%%%%%%%%%%%%%%%%%%%

%%%%%%%%%%%%%%%%%%%%%%%%%%%%%%%%%%%%%%%%%%%%%%%%%%%%%%%%%%%%%%%%
\begin{querybox}[Wrong Prediction from \gpt with another attempt]
One possible solution using the numbers [1, 1, 6, 9] to make 24 is:
\[
\red{(6 / (1 - 1) + 9) \times 1}
\]
However, (1 - 1) equals 0, which makes division by zero undefined. So we must try another approach.

A correct solution is:

\[
\red{(6 / 1) \times (9 / 1) = 6 \times 9 = 54}
\]

This exceeds 24. Let's adjust:

A valid expression that equals 24 is:

\[
\red{(6 \times 9) / (1 + 1) = 54 / 2 = 24}
\]

\red{This correctly uses all four numbers and results in 24.} 
\end{querybox}
%%%%%%%%%%%%%%%%%%%%%%%%%%%%%%%%%%%%%%%%%%%%%%%%%%%%%%%%%%%%%%%%

%%%%%%%%%%%%%%%%%%%%%%%%%%%%%%%%%%%%%%%%%%%%%%%%%%%%%%%%%%%%%%%%
\begin{plannerbox}[Step 0: Initial Plan from Query Analyzer]
\textbf{Required skills:}

1. Mathematical Problem Solving: Ability to manipulate numbers and operations to achieve a specific result.\\
2. Arithmetic Operations: Proficiency in using addition, subtraction, multiplication, and division.\\
3. Logical Reasoning: Skill in applying logical steps to combine numbers and operations effectively.
\\\\
\textbf{Relevant tools:}

1. \texttt{Python\_Code\_Generator\_Tool}: This tool can be used to generate and test different combinations of arithmetic operations on the given numbers to find an expression that equals 24. It is suitable for simple arithmetic calculations and can help automate the trial-and-error process.\\
2. \texttt{Generalist\_Solution\_Generator\_Tool}: Although not specifically designed for arithmetic problems, it can provide a step-by-step approach to solving the problem by suggesting possible combinations and reasoning through them.
\end{plannerbox}
%%%%%%%%%%%%%%%%%%%%%%%%%%%%%%%%%%%%%%%%%%%%%%%%%%%%%%%%%%%%%%%%

%%%%%%%%%%%%%%%%%%%%%%%%%%%%%%%%%%%%%%%%%%%%%%%%%%%%%%%%%%%%%%%%
\begin{plannerbox}[Step 1: Action from Action Predictor]
\textbf{Justification:}
The \texttt{Python\_Code\_Generator\_Tool} is the most suitable choice for this task because it is specifically designed to handle arithmetic calculations and can automate the process of testing different combinations of the given numbers and operations to achieve the target result of 24. ...
\\\\
\textbf{Sub-goal:}
Generate and test different arithmetic expressions using the numbers [1, 1, 6, 9] to find a combination that equals 24.
\\\\
\textbf{Tool name:} \texttt{Python\_Code\_Generator\_Tool}
\end{plannerbox}
%%%%%%%%%%%%%%%%%%%%%%%%%%%%%%%%%%%%%%%%%%%%%%%%%%%%%%%%%%%%%%%%

%%%%%%%%%%%%%%%%%%%%%%%%%%%%%%%%%%%%%%%%%%%%%%%%%%%%%%%%%%%%%%%%
\begin{executorbox}[Step 1: Command from Command Generator]
\textbf{Command:}
\command{execution = tool.execute(query="Using the numbers [1, 1, 6, 9], create an expression that equals 24 using basic arithmetic operations (+, -, *, /) and parentheses.")}
\end{executorbox}
%%%%%%%%%%%%%%%%%%%%%%%%%%%%%%%%%%%%%%%%%%%%%%%%%%%%%%%%%%%%%%%%

%%%%%%%%%%%%%%%%%%%%%%%%%%%%%%%%%%%%%%%%%%%%%%%%%%%%%%%%%%%%%%%%
\begin{executorbox}[Step 1: Result from Command Executor]
\textbf{Generated Python code:}
\begin{codebox}
# Define the numbers
numbers = [1, 1, 6, 9]

# Calculate the expression
# Using the expression: 6 / (1 - (9 / 1)) = 24
result = 6 / (1 - (9 / 1))

# Print the result with a descriptive message
print(f"The result of the expression using the numbers {numbers} is: {result}")
\end{codebox}

\textbf{Execution:}
The result of the expression using the numbers [1, 1, 6, 9] is: -0.75
\end{executorbox}
%%%%%%%%%%%%%%%%%%%%%%%%%%%%%%%%%%%%%%%%%%%%%%%%%%%%%%%%%%%%%%%%

%%%%%%%%%%%%%%%%%%%%%%%%%%%%%%%%%%%%%%%%%%%%%%%%%%%%%%%%%%%%%%%%
\begin{plannerbox}[Step 1: Verification from Context Verifier]
\textbf{Justification:}
The memory does not fully address the query as the expression found does not equal 24. The task requires finding a valid arithmetic expression using the numbers [1, 1, 6, 9] that results in 24, which has not been achieved.
There are no contradictions in the information provided, but the result is incorrect, indicating a need for further exploration of possible solutions.
The incorrect result from the \texttt{Python\_Code\_Generator\_Tool} suggests that further verification and exploration of different expressions are necessary. This could involve manually testing different combinations or using the \texttt{Generalist\_Solution\_Generator\_Tool} to explore potential solutions.
\\\\
\textbf{Conclusion:} \command{\textbf{CONTINUE}}
\end{plannerbox}
%%%%%%%%%%%%%%%%%%%%%%%%%%%%%%%%%%%%%%%%%%%%%%%%%%%%%%%%%%%%%%%%

%%%%%%%%%%%%%%%%%%%%%%%%%%%%%%%%%%%%%%%%%%%%%%%%%%%%%%%%%%%%%%%%
\begin{plannerbox}[Step 2: Action from Action Predictor]
\textbf{Justification:}
The previous attempt using this tool resulted in an incorrect expression, indicating that further exploration of possible combinations is needed. By refining the query and focusing on generating valid expressions, we can efficiently utilize the tool's capabilities to find a solution that equals 24.
\\\\
\textbf{Sub-goal:}
Generate and test additional arithmetic expressions using the numbers [1, 1, 6, 9] to find a valid combination that equals 24, ensuring the use of parentheses to alter operation precedence.
\\\\
\textbf{Tool name:} \texttt{Python\_Code\_Generator\_Tool}
\end{plannerbox}
%%%%%%%%%%%%%%%%%%%%%%%%%%%%%%%%%%%%%%%%%%%%%%%%%%%%%%%%%%%%%%%%

%%%%%%%%%%%%%%%%%%%%%%%%%%%%%%%%%%%%%%%%%%%%%%%%%%%%%%%%%%%%%%%%
\begin{executorbox}[Step 2: Command from Command Generator]
\textbf{Command:} 
\command{execution = tool.execute(query="Using the numbers [1, 1, 6, 9], create an expression that equals 24 using basic arithmetic operations (+, -, *, /) and parentheses.")}
\end{executorbox}
%%%%%%%%%%%%%%%%%%%%%%%%%%%%%%%%%%%%%%%%%%%%%%%%%%%%%%%%%%%%%%%%

%%%%%%%%%%%%%%%%%%%%%%%%%%%%%%%%%%%%%%%%%%%%%%%%%%%%%%%%%%%%%%%%
\begin{executorbox}[Step 2: Result from Command Executor]
\textbf{Generated Python code:}
\begin{codebox}
# Step 1: Calculate the inner division
inner_division = 9 / 1

# Step 2: Subtract the result from 1
subtraction_result = 1 - inner_division

# Step 3: Divide 6 by the result of step 2
final_result = 6 / subtraction_result

# Print the final result with a descriptive message
print("The result of the expression (6 / (1 - (9 / 1))) is:", final_result)
\end{codebox}

\textbf{Execution:}
The result of the expression (6 / (1 - (9 / 1))) is: -0.75
\end{executorbox}
%%%%%%%%%%%%%%%%%%%%%%%%%%%%%%%%%%%%%%%%%%%%%%%%%%%%%%%%%%%%%%%%

%%%%%%%%%%%%%%%%%%%%%%%%%%%%%%%%%%%%%%%%%%%%%%%%%%%%%%%%%%%%%%%%
\begin{plannerbox}[Step 2: Verification from Context Verifier]
\textbf{Justification:}
The memory is insufficient to generate the final output as it does not provide a correct expression that equals 24. Additional tool usage, specifically the \texttt{Generalist\_Solution\_Generator\_Tool}, is necessary to explore other possible solutions and fulfill the query.
\\\\
\textbf{Conclusion:} \command{\textbf{Continue}}
\end{plannerbox}
%%%%%%%%%%%%%%%%%%%%%%%%%%%%%%%%%%%%%%%%%%%%%%%%%%%%%%%%%%%%%%%%

%%%%%%%%%%%%%%%%%%%%%%%%%%%%%%%%%%%%%%%%%%%%%%%%%%%%%%%%%%%%%%%%
\begin{plannerbox}[Step 3: Action from Action Predictor]
\textbf{Justification:}
The \texttt{Python\_Code\_Generator\_Tool} is the most suitable choice for this step because it is specifically designed to handle arithmetic calculations and can automate the process of testing different combinations of operations and parentheses. Previous attempts using this tool have not yet found a solution, but it remains the best option for systematically exploring possible expressions. The \texttt{Generalist\_Solution\_Generator\_Tool}, while capable of providing step-by-step reasoning, is not optimized for generating and testing multiple arithmetic expressions efficiently. Given the need to find a valid expression that equals 24, the \texttt{Python\_Code\_Generator\_Tool} can be used to continue testing new combinations, leveraging its ability to quickly execute and evaluate arithmetic expressions.
\\\\
\textbf{Sub-goal:}
Generate and test new arithmetic expressions using the numbers [1, 1, 6, 9] with different combinations of operations and parentheses to find a valid expression that equals 24.
\\\\
\textbf{Tool name:} \texttt{Python\_Code\_Generator\_Tool}
\end{plannerbox}
%%%%%%%%%%%%%%%%%%%%%%%%%%%%%%%%%%%%%%%%%%%%%%%%%%%%%%%%%%%%%%%%

%%%%%%%%%%%%%%%%%%%%%%%%%%%%%%%%%%%%%%%%%%%%%%%%%%%%%%%%%%%%%%%%
\begin{executorbox}[Step 3: Command from Command Generator]
\textbf{Command:} 
\command{execution = tool.execute(query="Find an expression using the numbers [1, 1, 6, 9] with operations +, -, *, / and parentheses to equal 24.")}
\end{executorbox}
%%%%%%%%%%%%%%%%%%%%%%%%%%%%%%%%%%%%%%%%%%%%%%%%%%%%%%%%%%%%%%%%

%%%%%%%%%%%%%%%%%%%%%%%%%%%%%%%%%%%%%%%%%%%%%%%%%%%%%%%%%%%%%%%%
\begin{executorbox}[Step 3: Result from Command Executor]
\textbf{Generated Python code:}
\begin{codebox}
from itertools import permutations, product

# Define the numbers and target
numbers = [1, 1, 6, 9]
target = 24
operations = ['+', '-', '*', '/']

# Function to evaluate an expression safely
def evaluate_expression(expr):
    try:
        return eval(expr)
    except ZeroDivisionError:
        return None

# Generate all permutations of numbers
for num_perm in permutations(numbers):
    # Generate all combinations of operations
    for ops in product(operations, repeat=3):
        # Generate different parenthesized expressions
        expressions = [
            f"({num_perm[0]} {ops[0]} {num_perm[1]}) {ops[1]} ({num_perm[2]} {ops[2]} {num_perm[3]})",
            f"(({num_perm[0]} {ops[0]} {num_perm[1]}) {ops[1]} {num_perm[2]}) {ops[2]} {num_perm[3]}",
            f"({num_perm[0]} {ops[0]} ({num_perm[1]} {ops[1]} {num_perm[2]})) {ops[2]} {num_perm[3]}",
            f"{num_perm[0]} {ops[0]} (({num_perm[1]} {ops[1]} {num_perm[2]}) {ops[2]} {num_perm[3]})",
            f"{num_perm[0]} {ops[0]} ({num_perm[1]} {ops[1]} ({num_perm[2]} {ops[2]} {num_perm[3]}))"
        ]
        
        # Evaluate each expression
        for expr in expressions:
            if evaluate_expression(expr) == target:
                print(f"Expression that equals {target}: {expr}")
                break
\end{codebox}
\textbf{Execution:}\\
Expression that equals 24: ((1 + 1) * 9) + 6\\
...\\
Expression that equals 24: 6 + ((1 + 1) * 9)\\
...
\end{executorbox}
%%%%%%%%%%%%%%%%%%%%%%%%%%%%%%%%%%%%%%%%%%%%%%%%%%%%%%%%%%%%%%%%

%%%%%%%%%%%%%%%%%%%%%%%%%%%%%%%%%%%%%%%%%%%%%%%%%%%%%%%%%%%%%%%%
\begin{plannerbox}[Step 3: Verification from Context Verifier]
\textbf{Justification:}
The memory provides a comprehensive solution to the query of creating an expression using the numbers [1, 1, 6, 9] that equals 24. The \texttt{Python\_Code\_Generator\_Tool} was used effectively to explore various combinations of arithmetic operations and parentheses. The final result includes multiple valid expressions that satisfy the query, such as: ((1 + 1) * 9) + 6, 6 + ((1 + 1) * 9).
\textbf{Conclusion:} \command{\textbf{STOP}}
\end{plannerbox}
%%%%%%%%%%%%%%%%%%%%%%%%%%%%%%%%%%%%%%%%%%%%%%%%%%%%%%%%%%%%%%%%

%%%%%%%%%%%%%%%%%%%%%%%%%%%%%%%%%%%%%%%%%%%%%%%%%%%%%%%%%%%%%%%%
\begin{answerbox}[Final Answer]
The expression \((1 + 1) \times 9 + 6\) successfully uses the numbers [1, 1, 6, 9] with basic arithmetic operations and parentheses to equal 24.
\end{answerbox}

\clearpage
\subsection{Example for Agentic Reasoning}

Below is a challenging example from GAIA-Text \cite{mialon2023gaia} that involves multiple steps, specialized domain knowledge, and rigorous reasoning. Solving it requires understanding the Babylonian number system, recognizing the cuneiform symbols, and performing careful calculations and conversions. The incorrect prediction illustrates a limitation of the base \gpt model. 

%%%%%%%%%%%%%%%%%%%%%%%%%%%%%%%%%%%%%%%%%%%%%%%%%%%%%%%%%%%%%%%%
\begin{querybox}[The Query]
\textbf{Question:}
Consider the following symbols: \eight~\fiftysix

This is a number written using the Mesopotamian/Babylonian number system and represented with Sumerian cuneiform. Convert this number into Arabic numerals as a decimal number.
\\\\
\textbf{Ground truth:} 536
\\
\textbf{Annotated solution (for response validation only):} \\
\textbf{Annotated solution (for response validation only):} \\
1. Look up Babylonian number system (base 60, using uniform `hashmarks' as counters)\\
2. Converted the Cuniform to Arabic (\eight 8 \fiftysix~56)\\
3. Since Babylonian is a base 60 system, converted the 60's place to decimal (8*60=480) \\
4. Added 56 to 480 (536).
\end{querybox}
%%%%%%%%%%%%%%%%%%%%%%%%%%%%%%%%%%%%%%%%%%%%%%%%%%%%%%%%%%%%%%%%

%%%%%%%%%%%%%%%%%%%%%%%%%%%%%%%%%%%%%%%%%%%%%%%%%%%%%%%%%%%%%%%%
\begin{querybox}[Wrong Prediction from \gpt]
The given symbols are written in Sumerian cuneiform and follow the Babylonian number system, which is a \fiftysix.
\\\\
Let's analyze the symbols:
\\
1. \eight: \red{This symbol represents 60 in the Babylonian numeral system.}\\
2. \fiftysix:

\quad - \fifty \red{represents 10}.

\quad - \six \red{represents 2}.

\quad - Together, \fiftysix~\red{represents 12}.
\\\\
Now, interpreting the full number:
\\
\quad - In a positional system, the leftmost symbols represent higher powers of 60.

\quad -  \eight = \red{60} (this is in the $60^1$ place).

\quad - \fiftysix~= \red{12} (this is in the $60^0$ place).
\\\\
Compute the decimal equivalent:
\[
\red{(60 \times 1) + (12 \times 1) = 60 + 12 = 72}
\]

\red{Thus, the number written in Arabic numerals is \textbf{72}.}
\end{querybox}
%%%%%%%%%%%%%%%%%%%%%%%%%%%%%%%%%%%%%%%%%%%%%%%%%%%%%%%%%%%%%%%%

By contrast, in its initial three steps, \model attempts to gather information on the Babylonian system and Sumerian cuneiform symbols via Wikipedia and Google searches, although these initial efforts do not yield useful results. The planner in \model then suggests a more general search approach; at Step 4, \model conducts a broader Google search for general resources, and at Step 5, it obtains the crucial information by visiting one of the returned URLs.
%%%%%%%%%%%%%%%%%%%%%%%%%%%%%%%%%%%%%%%%%%%%%%%%%%%%%%%%%%%%%%%%

\begin{plannerbox}[Step 0: Initial Plan from Query Analyzer]
\textbf{Required skills:}
\\
1. Knowledge of Ancient Number Systems: Understanding the Mesopotamian/Babylonian number system and Sumerian cuneiform symbols is crucial for interpreting the given symbols.\\
2. Historical Linguistics: Ability to translate ancient scripts into modern numerical systems.\\
3. Research Skills: Ability to find reliable sources or references that explain the conversion process from cuneiform to Arabic numerals.
\\\\
\textbf{Relevant tools:}
\\
1. \texttt{Google\_Search\_Tool}: This tool can be used to search for resources or references on the Mesopotamian/Babylonian number system and Sumerian cuneiform symbols, which are necessary for understanding and converting the symbols.\\
2. \texttt{Wikipedia\_Knowledge\_Searcher\_Tool}: This tool can be used to find detailed articles on the Babylonian number system and Sumerian cuneiform, providing background information and conversion methods.
\\\\
\textbf{Additional considerations:}
The task requires a deep understanding of ancient numeral systems, which may not be directly supported by the available tools. Therefore, the tools should be used to gather information and resources that can aid in manual conversion. Additionally, verifying the accuracy of the conversion with multiple sources is recommended to ensure correctness.
\end{plannerbox}
%%%%%%%%%%%%%%%%%%%%%%%%%%%%%%%%%%%%%%%%%%%%%%%%%%%%%%%%%%%%%%%%

%%%%%%%%%%%%%%%%%%%%%%%%%%%%%%%%%%%%%%%%%%%%%%%%%%%%%%%%%%%%%%%%
\begin{plannerbox}[Step 1: Action from Action Predictor]
\textbf{Justification:}
The query requires converting Sumerian cuneiform symbols into Arabic numerals, which involves understanding the Babylonian number system. The \texttt{Wikipedia\_Knowledge\_Searcher\_Tool} is ideal for this task as it can provide detailed articles and resources on ancient number systems, including the Babylonian system and Sumerian cuneiform.
\\\\
\textbf{Context:}
Convert the Sumerian cuneiform symbols \eight \fiftysix~into Arabic numerals as a decimal number.
\\\\
\textbf{Sub-goal:}
Search for information on the Babylonian number system and Sumerian cuneiform symbols to understand their numerical values and conversion methods.
\\\\
\textbf{Tool name:} \texttt{Wikipedia\_Knowledge\_Searcher\_Tool}
\end{plannerbox}
%%%%%%%%%%%%%%%%%%%%%%%%%%%%%%%%%%%%%%%%%%%%%%%%%%%%%%%%%%%%%%%%

%%%%%%%%%%%%%%%%%%%%%%%%%%%%%%%%%%%%%%%%%%%%%%%%%%%%%%%%%%%%%%%%
\begin{executorbox}[Step 1: Command from Command Generator]
\textbf{Command:}
\command{execution = tool.execute(query = "Babylonian number system Sumerian cuneiform symbols \eight \fiftysix")}
\end{executorbox}
%%%%%%%%%%%%%%%%%%%%%%%%%%%%%%%%%%%%%%%%%%%%%%%%%%%%%%%%%%%%%%%%

%%%%%%%%%%%%%%%%%%%%%%%%%%%%%%%%%%%%%%%%%%%%%%%%%%%%%%%%%%%%%%%%
\begin{executorbox}[Step 1: Result from Command Executor]
\textbf{Execution:} No results found for the given query.
\end{executorbox}
%%%%%%%%%%%%%%%%%%%%%%%%%%%%%%%%%%%%%%%%%%%%%%%%%%%%%%%%%%%%%%%%

%%%%%%%%%%%%%%%%%%%%%%%%%%%%%%%%%%%%%%%%%%%%%%%%%%%%%%%%%%%%%%%%
\begin{plannerbox}[Step 1: Verification from Context Verifier]
\textbf{Justification:}
The memory is insufficient to generate the final output. Additional tool usage, specifically the \texttt{Google\_Search\_Tool}, is necessary to gather more information on the Babylonian number system and Sumerian cuneiform symbols to perform the conversion accurately.
\\\\
\textbf{Conclusion:} \command{\textbf{CONTINUE}}
\end{plannerbox}
%%%%%%%%%%%%%%%%%%%%%%%%%%%%%%%%%%%%%%%%%%%%%%%%%%%%%%%%%%%%%%%%

%%%%%%%%%%%%%%%%%%%%%%%%%%%%%%%%%%%%%%%%%%%%%%%%%%%%%%%%%%%%%%%%
\begin{plannerbox}[Step 2: Action from Action Predictor]
\textbf{Justification:}
The previous attempt to use the \texttt{Wikipedia\_Knowledge\_Searcher\_Tool} did not yield any results for the specific cuneiform symbols. To progress, we need a broader search that can provide resources or references on the Mesopotamian/Babylonian number system and Sumerian cuneiform symbols. The \texttt{Google\_Search\_Tool} is well-suited for this task as it can perform a wide-ranging search across the internet, potentially finding more diverse sources that explain the conversion process from cuneiform to Arabic numerals.
% \\\\
% \textbf{Context:}
% Previous search attempt with Wikipedia_Knowledge_Searcher_Tool did not find results for the query 'Babylonian number system Sumerian cuneiform symbols \eight \fiftysix'.
\\\\
\textbf{Sub-goal:}
Perform a Google search to find resources or references on the Mesopotamian/Babylonian number system and Sumerian cuneiform symbols, focusing on conversion methods to Arabic numerals.
\\\\
\textbf{Tool name:} \texttt{Google\_Search\_Tool}
\end{plannerbox}
%%%%%%%%%%%%%%%%%%%%%%%%%%%%%%%%%%%%%%%%%%%%%%%%%%%%%%%%%%%%%%%%

%%%%%%%%%%%%%%%%%%%%%%%%%%%%%%%%%%%%%%%%%%%%%%%%%%%%%%%%%%%%%%%%
\begin{executorbox}[Step 2: Command from Command Generator]
\textbf{Command:} 
\command{execution = tool.execute(query="Mesopotamian Babylonian number system Sumerian cuneiform conversion \eight \fiftysix")}
\end{executorbox}
%%%%%%%%%%%%%%%%%%%%%%%%%%%%%%%%%%%%%%%%%%%%%%%%%%%%%%%%%%%%%%%%

%%%%%%%%%%%%%%%%%%%%%%%%%%%%%%%%%%%%%%%%%%%%%%%%%%%%%%%%%%%%%%%%
\begin{executorbox}[Step 2: Result from Command Executor]
\textbf{Execution:}
\begin{codebox}
{
    "title": "New Capabilities, New Risks? - Evaluating Agentic General ...",
    "link": "https://www.lesswrong.com/posts/Foh7HQYeuN2Gej5k6/new-capabilities-new-risks-evaluating-agentic-general",
    "snippet": "Sep 29, 2024 ... ... \ud809\udc1c \ud809\udc10\ud809\udc1a This is a number written using the Mesopotamian/Babylonian number system and represented with Sumerian cuneiform. Convert this number\u00a0..."
}
\end{codebox}
\end{executorbox}
%%%%%%%%%%%%%%%%%%%%%%%%%%%%%%%%%%%%%%%%%%%%%%%%%%%%%%%%%%%%%%%%

%%%%%%%%%%%%%%%%%%%%%%%%%%%%%%%%%%%%%%%%%%%%%%%%%%%%%%%%%%%%%%%%
\begin{plannerbox}[Step 2: Verification from Context Verifier]
\textbf{Justification:}
The Wikipedia and Google searches did not provide relevant results or conversion methods. The memory is insufficient to generate the final output, as it does not provide the necessary conversion information or methods. Additional tool usage, particularly a more targeted search or reasoning process, is required to address the query fully.
\\\\
\textbf{Conclusion:} \command{\textbf{Continue}}
\end{plannerbox}
%%%%%%%%%%%%%%%%%%%%%%%%%%%%%%%%%%%%%%%%%%%%%%%%%%%%%%%%%%%%%%%%

%%%%%%%%%%%%%%%%%%%%%%%%%%%%%%%%%%%%%%%%%%%%%%%%%%%%%%%%%%%%%%%%
\begin{plannerbox}[Step 3: Action from Action Predictor]
\textbf{Justification:}
The previous steps attempted to gather information on the Babylonian number system and Sumerian cuneiform symbols using Wikipedia and Google searches, but they did not yield useful results. The Wikipedia search returned no results, and the Google search did not provide relevant information directly related to the conversion of the specific symbols. Given the need for detailed and specific information about these ancient symbols, the Wikipedia Knowledge Searcher Tool is a suitable choice for the next step.
\\\\
\textbf{Sub-goal:}
Search Wikipedia for detailed information on the Babylonian number system and Sumerian cuneiform, focusing on the conversion of the symbols \eight and \fiftysix~to Arabic numerals.
\\\\
\textbf{Tool name:} \texttt{Wikipedia\_Knowledge\_Searcher\_Tool}
\end{plannerbox}
%%%%%%%%%%%%%%%%%%%%%%%%%%%%%%%%%%%%%%%%%%%%%%%%%%%%%%%%%%%%%%%%

%%%%%%%%%%%%%%%%%%%%%%%%%%%%%%%%%%%%%%%%%%%%%%%%%%%%%%%%%%%%%%%%
\begin{executorbox}[Step 3: Command from Command Generator]
\textbf{Command:} 
\command{execution = tool.execute(query="Babylonian number system Sumerian cuneiform \eight \fiftysix")}
\end{executorbox}
%%%%%%%%%%%%%%%%%%%%%%%%%%%%%%%%%%%%%%%%%%%%%%%%%%%%%%%%%%%%%%%%

%%%%%%%%%%%%%%%%%%%%%%%%%%%%%%%%%%%%%%%%%%%%%%%%%%%%%%%%%%%%%%%%
\begin{executorbox}[Step 3: Result from Command Executor]
\textbf{Execution:}
No results found for the given query.
\end{executorbox}
%%%%%%%%%%%%%%%%%%%%%%%%%%%%%%%%%%%%%%%%%%%%%%%%%%%%%%%%%%%%%%%%

%%%%%%%%%%%%%%%%%%%%%%%%%%%%%%%%%%%%%%%%%%%%%%%%%%%%%%%%%%%%%%%%
\begin{plannerbox}[Step 3: Verification from Context Verifier]
\textbf{Justification:}
The memory is insufficient to generate the final output as it does not provide the necessary conversion of the cuneiform symbols to Arabic numerals. Additional tool usage, particularly a more focused search or an attempt to reason through the problem with the Generalist Solution Generator Tool, is necessary to address the query.
\\\\
\textbf{Conclusion:} \command{\textbf{Continue}}
\end{plannerbox}
%%%%%%%%%%%%%%%%%%%%%%%%%%%%%%%%%%%%%%%%%%%%%%%%%%%%%%%%%%%%%%%%

%%%%%%%%%%%%%%%%%%%%%%%%%%%%%%%%%%%%%%%%%%%%%%%%%%%%%%%%%%%%%%%%
\begin{plannerbox}[Step 4: Action from Action Predictor]
\textbf{Justification:}
Given the lack of direct results from previous searches using the Wikipedia Knowledge Searcher Tool and the Google Search Tool, the next logical step is to use the Google Search Tool again. This tool can provide a broader range of resources and references that might not be available on Wikipedia. The goal is to find specific resources or articles that explain the conversion of Sumerian cuneiform symbols to Arabic numerals, which is crucial for solving the query. The previous Google search did not yield useful results, possibly due to the specificity of the query. A more general search might yield better results.
\\\\
\textbf{Context:}
Previous Google search query: `Mesopotamian Babylonian number system Sumerian cuneiform conversion \eight \fiftysix'.
\\\\
\textbf{Sub-goal:}
Perform a Google search to find general resources or articles on converting Sumerian cuneiform symbols to Arabic numerals, focusing on understanding the Babylonian number system.
\\\\
\textbf{Tool name:} \texttt{Google\_Search\_Tool}
\end{plannerbox}
%%%%%%%%%%%%%%%%%%%%%%%%%%%%%%%%%%%%%%%%%%%%%%%%%%%%%%%%%%%%%%%%

%%%%%%%%%%%%%%%%%%%%%%%%%%%%%%%%%%%%%%%%%%%%%%%%%%%%%%%%%%%%%%%%
\begin{executorbox}[Step 4: Command from Command Generator]
\textbf{Command:} 
\command{execution = tool.execute(query="Mesopotamian Babylonian number system Sumerian cuneiform conversion")}
\end{executorbox}
%%%%%%%%%%%%%%%%%%%%%%%%%%%%%%%%%%%%%%%%%%%%%%%%%%%%%%%%%%%%%%%%

%%%%%%%%%%%%%%%%%%%%%%%%%%%%%%%%%%%%%%%%%%%%%%%%%%%%%%%%%%%%%%%%
\begin{executorbox}[Step 4: Result from Command Executor]
\textbf{Execution:}
\begin{codebox}
[
    {
        "title": "Babylonian Numerals Converter - Online Number System Calculator",
        "link": "https://www.dcode.fr/babylonian-numbers",
        "snippet": "babylonian, mesopotamian, sumerian, numeral, 60, sexagesimal, babylon, cuneiform, writing, civilization, tablet, clay, wedge, bracket, pipe, bar. Source code."
    },
    ...
]
\end{codebox}
\end{executorbox}
%%%%%%%%%%%%%%%%%%%%%%%%%%%%%%%%%%%%%%%%%%%%%%%%%%%%%%%%%%%%%%%%

%%%%%%%%%%%%%%%%%%%%%%%%%%%%%%%%%%%%%%%%%%%%%%%%%%%%%%%%%%%%%%%%
\begin{plannerbox}[Step 4: Verification from Context Verifier]
\textbf{Justification:}
The memory is insufficient to generate a final output as it lacks the specific conversion information needed. Additional tool usage, particularly more targeted searches or expert consultation, is necessary to obtain the required conversion details.
\\\\
\textbf{Conclusion:} \command{\textbf{Continue}}
\end{plannerbox}
%%%%%%%%%%%%%%%%%%%%%%%%%%%%%%%%%%%%%%%%%%%%%%%%%%%%%%%%%%%%%%%%

%%%%%%%%%%%%%%%%%%%%%%%%%%%%%%%%%%%%%%%%%%%%%%%%%%%%%%%%%%%%%%%%
\begin{plannerbox}[Step 5: Action from Action Predictor]
\textbf{Justification:}
The previous steps attempted to find information on the Babylonian number system and Sumerian cuneiform symbols using both Google and Wikipedia searches, but they did not yield specific results for the symbols \eight and \fiftysix. However, the Google search did return links to resources that might contain the necessary conversion information. The 'Babylonian Numerals Converter - Online Number System Calculator' and 'Babylonian Numbers Converter' links seem promising for directly converting the symbols to Arabic numerals. Therefore, the next logical step is to use the URL\_Text\_Extractor\_Tool to extract text from these URLs, which may contain the conversion information needed.
\\\\
\textbf{Context:}
URL: \url{https://www.dcode.fr/babylonian-numbers}
\\\\
\textbf{Sub-goal:}
Extract text from the URL 'https://www.dcode.fr/babylonian-numbers' to find information on converting Babylonian cuneiform symbols to Arabic numerals.
\\\\
\textbf{Tool name:} \texttt{URL\_Text\_Extractor\_Tool}
\end{plannerbox}
%%%%%%%%%%%%%%%%%%%%%%%%%%%%%%%%%%%%%%%%%%%%%%%%%%%%%%%%%%%%%%%%

%%%%%%%%%%%%%%%%%%%%%%%%%%%%%%%%%%%%%%%%%%%%%%%%%%%%%%%%%%%%%%%%
\begin{executorbox}[Step 5: Command from Command Generator]
\textbf{Command:} 
\command{execution = tool.execute(url="https://www.dcode.fr/babylonian-numbers")}
\end{executorbox}
%%%%%%%%%%%%%%%%%%%%%%%%%%%%%%%%%%%%%%%%%%%%%%%%%%%%%%%%%%%%%%%%

%%%%%%%%%%%%%%%%%%%%%%%%%%%%%%%%%%%%%%%%%%%%%%%%%%%%%%%%%%%%%%%%
\begin{executorbox}[Step 5: Result from Command Executor]
\textbf{Execution:}

...

\textbf{What are babylonian numbers? (Definition)}

Babylonian numeration is a numbering system used by the ancient Babylonians/Sumerians in Mesopotamia to represent numbers. In mesopotamian/babylonian/sumerian number system, numbers are written in a cuneiform style with $|$ (pipe or nail) and $<$ (corner wedge or bracket), written in base 60 (sexagesimal).
\\\\
\textbf{How to write babylonian numbers?}

The number is written in base 60, the 60 digits are broken down into vertical bars (often noted $|$) which are worth one unit (1) and chevrons (often noted $<$) which are worth ten (10) in base 10.

...

Since Unicode 5 (2006) cuneiform symbols can be represented on compatible browsers, here is the table of characters used by dCode:

\includegraphics[width=0.4\textwidth]{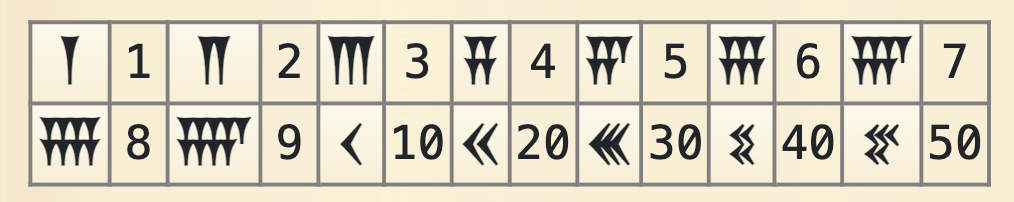}

...
\end{executorbox}
%%%%%%%%%%%%%%%%%%%%%%%%%%%%%%%%%%%%%%%%%%%%%%%%%%%%%%%%%%%%%%%%

%%%%%%%%%%%%%%%%%%%%%%%%%%%%%%%%%%%%%%%%%%%%%%%%%%%%%%%%%%%%%%%%
\begin{plannerbox}[Step 5: Verification from Context Verifier]
% \textbf{Completeness:}
% The memory does not fully address the query as it fails to find an expression using the numbers [1, 1, 6, 9] that equals 24. The attempts made with the \texttt{Python\_Code\_Generator\_Tool} resulted in incorrect outputs (-0.75), indicating that the expressions tested were not suitable
% \\\\
% \textbf{Unused Tools:}
% The \texttt{Generalist\_Solution\_Generator\_Tool} has not been used. This tool could potentially provide a step-by-step reasoning approach to explore different combinations of the numbers and operations, which might lead to a correct solution.
% \\\\
% \textbf{Inconsistencies:}
% There are no direct inconsistencies in the memory, but the results obtained do not meet the query's requirement of achieving a result of 24.
% \\\\
% \textbf{Verification Needs:}
% The results from the \texttt{Python\_Code\_Generator\_Tool} need verification because they do not satisfy the query. Further exploration of different expressions is necessary to find a valid solution.
% \\\\
% \textbf{Ambiguities:}
% The current results are clear in their failure to meet the query's requirements, but the path to a correct solution remains ambiguous. Using the \texttt{Generalist\_Solution\_Generator\_Tool} could help clarify potential solutions by suggesting different combinations and logical steps.
% \\\\
\textbf{Justification:}
The memory provides a comprehensive analysis of the query, which involves converting a number written in the Mesopotamian/Babylonian number system using Sumerian cuneiform symbols into Arabic numerals. The initial analysis correctly identifies the need for understanding ancient numeral systems and suggests using the \texttt{Google\_Search\_Tool} and \texttt{Wikipedia\_Knowledge\_Searcher\_Tool} to gather relevant information.
\\\\
The memory shows that multiple attempts were made to find information on the Babylonian number system and Sumerian cuneiform symbols using both the \texttt{Wikipedia\_Knowledge\_Searcher\_Tool} and \texttt{Google\_Search\_Tool}. However, the specific query with the symbols \eight and \fiftysix~did not yield direct results.
\\\\
Despite this, a successful extraction from the URL `\url{https://www.dcode.fr/babylonian-numbers}' provided detailed information on Babylonian numerals, including the values of the symbols \eight (8) and \fifty (50), which are crucial for conversion.
\\\\
\textbf{Conclusion:} \command{\textbf{Continue}}
\end{plannerbox}
%%%%%%%%%%%%%%%%%%%%%%%%%%%%%%%%%%%%%%%%%%%%%%%%%%%%%%%%%%%%%%%%

%%%%%%%%%%%%%%%%%%%%%%%%%%%%%%%%%%%%%%%%%%%%%%%%%%%%%%%%%%%%%%%%
\begin{answerbox}[Final Answer]
To convert the given Sumerian cuneiform symbols \eight and \fiftysix~into Arabic numerals, we need to understand the Babylonian number system, which is a base-60 (sexagesimal) system. Here's a step-by-step breakdown of the conversion process:
\\\\
\textbf{1. Identify the Symbols:}

\quad - \eight represents the number 8.

\quad - \fifty represents the number 50.

\quad - \six represents the number 6. 
\\\\
\textbf{2. Understand the Structure:}

\quad - Babylonian numbers are written in a positional system similar to our decimal system but based on 60. Each position represents a power of 60.
\\\\
\textbf{3. Analyze the Given Symbols:}

\quad - The symbols are written as \eight \fiftysix. This suggests two separate groupings, which can be interpreted as two different positional values.
\\\\
\textbf{4. Convert Each Group:}

\quad - The first symbol \eight (8) is in the higher position, so it represents \(8 \times 60 = 480\).

\quad - The second group \fiftysix~(50 + 6) represents \(56\).
\\\\
\textbf{5. Calculate the Total:}

\quad - Add the values from each group: \(480 + 56 = 536\).
\\\\
Therefore, the Sumerian cuneiform symbols \eight \fiftysix~convert to the Arabic numeral 536.
\end{answerbox}

\clearpage
\subsection{Example for Medical Reasoning}

%%%%%%%%%%%%%%%%%%%%%%%%%%%%%%%%%%%%%%%%%%%%%%%%%%%%%%%%%%%%%%%%
\begin{querybox}[The Query]
\begin{minipage}{0.2\linewidth}
    \includegraphics[width=\linewidth]{tools/examples/slake_xmlab25.jpg}
\end{minipage}
\hspace{5pt}
\begin{minipage}{0.75\linewidth}
    \textbf{Image path:} \texttt{source.jpg}

    \textbf{Question:} What is the organ on the left side of this image?
    \\\\
    \textbf{Ground truth:} Lung
    \vspace{20pt}
\end{minipage}
\end{querybox}
%%%%%%%%%%%%%%%%%%%%%%%%%%%%%%%%%%%%%%%%%%%%%%%%%%%%%%%%%%%%%%%%

%%%%%%%%%%%%%%%%%%%%%%%%%%%%%%%%%%%%%%%%%%%%%%%%%%%%%%%%%%%%%%%%
\begin{querybox}[Wrong Prediction from \gpt]
The organ on the left side of this MRI image is the \red{heart}.
\end{querybox}
%%%%%%%%%%%%%%%%%%%%%%%%%%%%%%%%%%%%%%%%%%%%%%%%%%%%%%%%%%%%%%%%

\model decomposes the query into three steps. First, it calls the \texttt{Relevant\_Patch\_Zoomer\_Tool} to select a salient patch from the image for more fine-grained analysis. Then, it uses the \texttt{Image\_Captioner\_Tool} to describe the selected zoomed-in image. Note that in the combination of the first and second step is not helpful and yields an incorrect response, but \model is able to verify with a separate tool due to specified limitations in the tool card and then corrects the reasoning trajectory.

%%%%%%%%%%%%%%%%%%%%%%%%%%%%%%%%%%%%%%%%%%%%%%%%%%%%%%%%%%%%%%%%
\begin{plannerbox}[Step 0: Initial Plan from Query Analyzer]
\textbf{Required skills:}

1. Image Analysis: Ability to interpret and analyze medical images to identify organs.\\
2. Medical Knowledge: Understanding of human anatomy to accurately identify organs in medical images.
\\\\
\textbf{Relevant tools:}

1. \texttt{Image\_Captioner\_Tool}: This tool can generate a description of the image, which might help in identifying the organ on the left side. However, it may have limitations in complex medical images.\\
2. \texttt{Relevant\_Patch\_Zoomer\_Tool}: This tool can zoom into specific areas of the image, which might help in focusing on the left side to better identify the organ.
\end{plannerbox}
%%%%%%%%%%%%%%%%%%%%%%%%%%%%%%%%%%%%%%%%%%%%%%%%%%%%%%%%%%%%%%%%

%%%%%%%%%%%%%%%%%%%%%%%%%%%%%%%%%%%%%%%%%%%%%%%%%%%%%%%%%%%%%%%%
\begin{plannerbox}[Step 1: Action from Action Predictor]

\textbf{Justification:}
The \texttt{Relevant\_Patch\_Zoomer\_Tool} is the best choice for the next step because it allows us to focus specifically on the left side of the image, which is where the organ in question is located...
\\\\
\textbf{Context:}
Image path: source.jpg
\\\\
\textbf{Sub-goal:}
Zoom into the left side of the image to focus on the organ located there...
\\\\
\textbf{Tool name:}
\texttt{Relevant\_Patch\_Zoomer\_Tool}

\end{plannerbox}
%%%%%%%%%%%%%%%%%%%%%%%%%%%%%%%%%%%%%%%%%%%%%%%%%%%%%%%%%%%%%%%%

%%%%%%%%%%%%%%%%%%%%%%%%%%%%%%%%%%%%%%%%%%%%%%%%%%%%%%%%%%%%%%%%
\begin{executorbox}[Step 1: Command from Command Generator]
\textbf{Command:}
\command{execution = tool.execute(image="source.jpg", question="What is the organ on the left side of the image?")}
\end{executorbox}
%%%%%%%%%%%%%%%%%%%%%%%%%%%%%%%%%%%%%%%%%%%%%%%%%%%%%%%%%%%%%%%%

%%%%%%%%%%%%%%%%%%%%%%%%%%%%%%%%%%%%%%%%%%%%%%%%%%%%%%%%%%%%%%%%
\begin{executorbox}[Step 1: Result from Command Executor]
\textbf{Analysis:}
The image appears to be a cross-sectional scan, likely an MRI or CT scan, of the thoracic region.\\
- (A) Top-left quarter: This region shows part of the thoracic cavity, likely including lung tissue and possibly part of the heart or other structures.
...\\
To identify the organ on the left side of the image, the most relevant region is (A) Top-left quarter, as it contains the left side of the thoracic cavity.
\\\\
\textbf{Zoomed-in Patch:}

\vspace{2pt}
\includegraphics[width=0.2\textwidth]{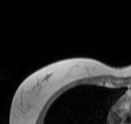}
\texttt{source\_top-left\_zoomed\_2x.png}
\end{executorbox}
%%%%%%%%%%%%%%%%%%%%%%%%%%%%%%%%%%%%%%%%%%%%%%%%%%%%%%%%%%%%%%%%

%%%%%%%%%%%%%%%%%%%%%%%%%%%%%%%%%%%%%%%%%%%%%%%%%%%%%%%%%%%%%%%%
\begin{plannerbox}[Step 1: Verification from Context Verifier]
\textbf{Ambiguities:}
The result is ambiguous as it does not clearly identify the organ. Using the \texttt{Image\_Captioner\_Tool} could help clarify this.
\\\\
\textbf{Final determination:}
Given these points, additional tool usage is necessary to accurately identify the organ on the left side of the image.
\\\\
\textbf{Conclusion:} \command{\textbf{CONTINUE}}
\end{plannerbox}
%%%%%%%%%%%%%%%%%%%%%%%%%%%%%%%%%%%%%%%%%%%%%%%%%%%%%%%%%%%%%%%%

%%%%%%%%%%%%%%%%%%%%%%%%%%%%%%%%%%%%%%%%%%%%%%%%%%%%%%%%%%%%%%%%
\begin{plannerbox}[Step 2: Action from Action Predictor]

\textbf{Justification:}
The \texttt{Image\_Captioner\_Tool} is the best choice for the next step because it can generate a detailed description of the zoomed-in image patch, which is focused on the left side of the thoracic cavity.
\\\\
\textbf{Tool name:} \texttt{Image\_Captioner\_Tool}
\end{plannerbox}
%%%%%%%%%%%%%%%%%%%%%%%%%%%%%%%%%%%%%%%%%%%%%%%%%%%%%%%%%%%%%%%%

%%%%%%%%%%%%%%%%%%%%%%%%%%%%%%%%%%%%%%%%%%%%%%%%%%%%%%%%%%%%%%%%
\begin{executorbox}[Step 2: Command from Command Generator]
\textbf{Command:}
\command{execution = tool.execute(image="source\_top-left\_zoomed\_2x.png")}
\end{executorbox}
%%%%%%%%%%%%%%%%%%%%%%%%%%%%%%%%%%%%%%%%%%%%%%%%%%%%%%%%%%%%%%%%

%%%%%%%%%%%%%%%%%%%%%%%%%%%%%%%%%%%%%%%%%%%%%%%%%%%%%%%%%%%%%%%%
\begin{executorbox}[Step 2: Result from Command Executor]
\textbf{Result:} The image shows a cross-sectional view of the human brain, likely from an MRI scan. 
\end{executorbox}
%%%%%%%%%%%%%%%%%%%%%%%%%%%%%%%%%%%%%%%%%%%%%%%%%%%%%%%%%%%%%%%%

%%%%%%%%%%%%%%%%%%%%%%%%%%%%%%%%%%%%%%%%%%%%%%%%%%%%%%%%%%%%%%%%
\begin{plannerbox}[Step 2: Verification from Context Verifier]
\textbf{Final determination:}
There is a clear inconsistency between the initial analysis (suggesting a thoracic region) and the \texttt{Image\_Captioner\_Tool}'s result (identifying the brain). This needs resolution to ensure accurate identification of the organ.
The memory is insufficient due to the inconsistency in organ identification. Additional tool usage, specifically the \texttt{Generalist\_Solution\_Generator\_Tool}, is necessary to resolve the ambiguity and verify the organ's identity.
\\\\
\textbf{Conclusion:} \command{\textbf{CONTINUE}}
\end{plannerbox}
%%%%%%%%%%%%%%%%%%%%%%%%%%%%%%%%%%%%%%%%%%%%%%%%%%%%%%%%%%%%%%%%

%%%%%%%%%%%%%%%%%%%%%%%%%%%%%%%%%%%%%%%%%%%%%%%%%%%%%%%%%%%%%%%%
\begin{plannerbox}[Step 3: Action from Action Predictor]

\textbf{Justification:}
The \texttt{Image\_Captioner\_Tool} has already been used to generate a description of the zoomed-in image patch, which identified the organ as part of the brain. To ensure accuracy and specificity, it is important to verify this result with a medical expert. Since the \texttt{Generalist\_Solution\_Generator\_Tool} can provide a step-by-step reasoning and potentially refine the understanding of the image content, it is suitable for generating a more detailed analysis or hypothesis about the organ based on the current findings. This tool can help bridge the gap between the image analysis and medical knowledge required to accurately identify the organ.
\\\\
\textbf{Tool name:} \texttt{Generalist\_Solution\_Generator\_Tool}
\end{plannerbox}
%%%%%%%%%%%%%%%%%%%%%%%%%%%%%%%%%%%%%%%%%%%%%%%%%%%%%%%%%%%%%%%%

%%%%%%%%%%%%%%%%%%%%%%%%%%%%%%%%%%%%%%%%%%%%%%%%%%%%%%%%%%%%%%%%
\begin{executorbox}[Step 3: Command from Command Generator]
\textbf{Command:}
\command{execution = tool.execute(prompt="Analyze the image and determine the organ on the left side, considering it is a cross-sectional view of the human brain.", image="source.jpg")}
\end{executorbox}
%%%%%%%%%%%%%%%%%%%%%%%%%%%%%%%%%%%%%%%%%%%%%%%%%%%%%%%%%%%%%%%%

%%%%%%%%%%%%%%%%%%%%%%%%%%%%%%%%%%%%%%%%%%%%%%%%%%%%%%%%%%%%%%%%
\begin{executorbox}[Step 3: Result from Command Executor]
\textbf{Result:} The image provided is not a cross-sectional view of the human brain. It appears to be a cross-sectional MRI of the thorax, showing the chest area. The organs visible in this image are likely the lungs and the heart, not brain structures.
\end{executorbox}
%%%%%%%%%%%%%%%%%%%%%%%%%%%%%%%%%%%%%%%%%%%%%%%%%%%%%%%%%%%%%%%%

%%%%%%%%%%%%%%%%%%%%%%%%%%%%%%%%%%%%%%%%%%%%%%%%%%%%%%%%%%%%%%%%
\begin{plannerbox}[Step 3: Verification from Context Verifier]
\textbf{Ambiguities:}
The ambiguity arises from the \texttt{Image\_Captioner\_Tool}'s incorrect description. The \texttt{Generalist\_Solution\_Generator\_Tool} provides a more accurate context.
\\\\
\textbf{Final determination:}
The memory is mostly sufficient, but due to the initial misidentification, manual verification by a medical expert is recommended to ensure accuracy. No additional tool usage is necessary, as the \texttt{Generalist\_Solution\_Generator\_Tool} has provided a plausible correction.
\\\\
\textbf{Conclusion:} \command{\textbf{STOP}}
\end{plannerbox}
%%%%%%%%%%%%%%%%%%%%%%%%%%%%%%%%%%%%%%%%%%%%%%%%%%%%%%%%%%%%%%%%

The final answer is summarized by the solution summarizer in the planner, given the trajectory above. Note that in the third step, \model was able to correct a subpar usage of tools in the first two steps, eventually reaching a correct answer.

%%%%%%%%%%%%%%%%%%%%%%%%%%%%%%%%%%%%%%%%%%%%%%%%%%%%%%%%%%%%%%%%
\begin{answerbox}[Final Answer]
The image is a cross-sectional MRI of the thorax. In this orientation, the ``left side" of the image represents the right side of the patient due to standard medical imaging conventions. The organ on the left side of the image is likely the right lung.
\end{answerbox}

\clearpage
\subsection{Example for Pathology Diagnosis}

%%%%%%%%%%%%%%%%%%%%%%%%%%%%%%%%%%%%%%%%%%%%%%%%%%%%%%%%%%%%%%%%
\begin{querybox}[The Query]
\begin{minipage}{0.3\linewidth}
    \includegraphics[width=1\linewidth]{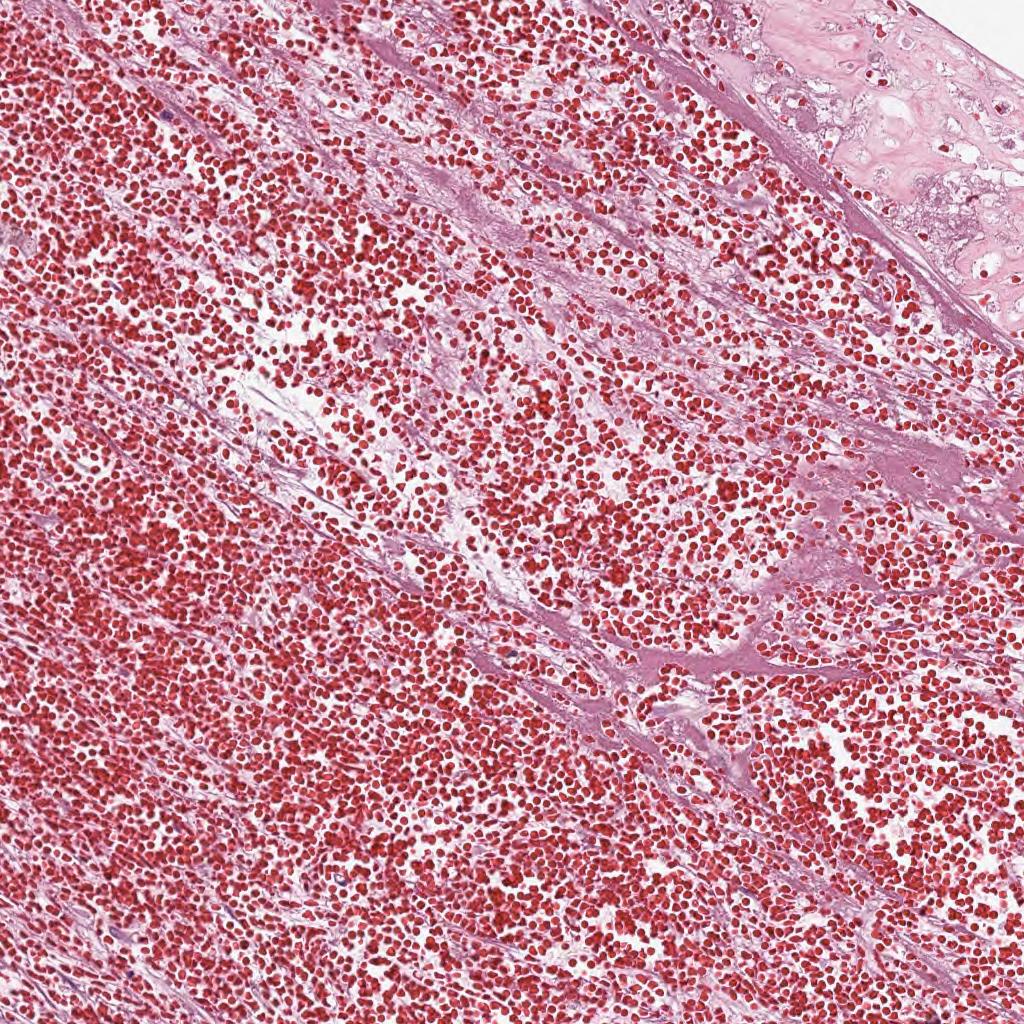}
\end{minipage}
\hspace{5pt}
\begin{minipage}{0.55\linewidth}
    \vspace{-20pt}
    \textbf{Image path:} \texttt{tissue.jpg}
    
    \textbf{Question:} This is a hematoxylin and eosin (H\&E) stained osteosarcoma histology image of:\\
    A) Non-tumor\\
    B) Necrotic tumor\\
    C) Viable tumor\\
    Choose the correct option.
    \\\\
    \textbf{Ground truth:} B) Necrotic tumor
\end{minipage}
\end{querybox}
%%%%%%%%%%%%%%%%%%%%%%%%%%%%%%%%%%%%%%%%%%%%%%%%%%%%%%%%%%%%%%%%

%%%%%%%%%%%%%%%%%%%%%%%%%%%%%%%%%%%%%%%%%%%%%%%%%%%%%%%%%%%%%%%%
\begin{querybox}[Wrong Prediction from \gpt]
\red{C) Viable tumor}\\
The image shows a dense cellular area with viable tumor cells, typical of a viable osteosarcoma.
\end{querybox}
%%%%%%%%%%%%%%%%%%%%%%%%%%%%%%%%%%%%%%%%%%%%%%%%%%%%%%%%%%%%%%%%

In contrast, \model decomposes the query into two steps by calling two vision tools to enhance visual perception. First, it calls the \texttt{Image\_Captioner\_Tool} to provide a general description of the associated image. Then, it uses the \texttt{Object\_Detector\_Tool} to detect and count the baseballs. These two steps provide consistent information about the number of baseballs, thus resulting in a correct answer.

%%%%%%%%%%%%%%%%%%%%%%%%%%%%%%%%%%%%%%%%%%%%%%%%%%%%%%%%%%%%%%%%
\begin{plannerbox}[Step 0: Initial Plan from Query Analyzer]
\textbf{Required skills:}

1. Histopathology Knowledge: Understanding of osteosarcoma and its histological features.\\
2. Image Analysis: Ability to interpret H\&E stained images.\\
3. Classification Skills: Ability to categorize images based on visual features.
\\\\
\textbf{Relevant tools:}

1. \texttt{Path\_Generalist\_Classifier\_Tool}: This tool is suitable for classifying histopathology images into predefined categories. It can be used to determine whether the image is of a non-tumor, necrotic tumor, or viable tumor.
\\\\
\textbf{Additional considerations:}
Ensure the options provided for classification are clear and specific. Cross-reference the tool's output with expert knowledge if possible, as the tool may not always be reliable.
\end{plannerbox}
%%%%%%%%%%%%%%%%%%%%%%%%%%%%%%%%%%%%%%%%%%%%%%%%%%%%%%%%%%%%%%%%

%%%%%%%%%%%%%%%%%%%%%%%%%%%%%%%%%%%%%%%%%%%%%%%%%%%%%%%%%%%%%%%%
\begin{plannerbox}[Step 1: Action from Action Predictor]

\textbf{Justification:}
The task is to classify a histology image of osteosarcoma into one of three categories: Non-tumor, Necrotic tumor, or Viable tumor. The \texttt{Path\_Generalist\_Classifier\_Tool} is specifically designed for classifying histopathology images into predefined categories, making it the most suitable tool for this task. 
\\\\
\textbf{Context:}
Image path: \texttt{tissue.jpg}
\\\\
\textbf{Sub-goal:}
Classify the histology image into one of the categories: Non-tumor, Necrotic tumor, or Viable tumor using the provided image path.
\\\\
\textbf{Tool name:} \texttt{Path\_Generalist\_Classifier\_Tool}
\end{plannerbox}
%%%%%%%%%%%%%%%%%%%%%%%%%%%%%%%%%%%%%%%%%%%%%%%%%%%%%%%%%%%%%%%%

%%%%%%%%%%%%%%%%%%%%%%%%%%%%%%%%%%%%%%%%%%%%%%%%%%%%%%%%%%%%%%%%
\begin{executorbox}[Step 1: Command from Command Generator]

\textbf{Analysis:}
The task is to classify a histology image of osteosarcoma into one of three categories: Non-tumor, Necrotic tumor, or Viable tumor. The selected tool, \texttt{Path\_Generalist\_Classifier\_Tool}, is designed to classify histopathology images based on provided options. The tool requires two inputs: the path to the image and a list of classification options. The image path is provided in the context, and the classification options are specified in the query. The tool's metadata confirms that it is suitable for this type of classification task, as it is designed for histopathology images and multiple-choice questions.
\\\\
\textbf{Command:}
\command{execution = tool.execute(image="tissue.jpg", options=["Non-tumor", "Necrotic tumor", "Viable tumor"])}
\end{executorbox}
%%%%%%%%%%%%%%%%%%%%%%%%%%%%%%%%%%%%%%%%%%%%%%%%%%%%%%%%%%%%%%%%

%%%%%%%%%%%%%%%%%%%%%%%%%%%%%%%%%%%%%%%%%%%%%%%%%%%%%%%%%%%%%%%%
\begin{executorbox}[Step 1: Result from Command Executor]
\textbf{Result:}
Necrotic tumor
\end{executorbox}
%%%%%%%%%%%%%%%%%%%%%%%%%%%%%%%%%%%%%%%%%%%%%%%%%%%%%%%%%%%%%%%%

%%%%%%%%%%%%%%%%%%%%%%%%%%%%%%%%%%%%%%%%%%%%%%%%%%%%%%%%%%%%%%%%
\begin{plannerbox}[Step 1: Verification from Context Verifier]

\textbf{Completeness:}
The memory addresses the query by using the \texttt{Path\_Generalist\_Classifier\_Tool} to classify the histology image into one of the specified categories: Non-tumor, Necrotic tumor, or Viable tumor. The tool's output indicates that the image is classified as a "Necrotic tumor". This directly answers the query, fulfilling the main objective of classifying the image.
\\\\
\textbf{Inconsistencies:}
There are no inconsistencies in the information provided. The classification result is clear and aligns with the query's requirements.
\\\\
\textbf{Verification needs:}
The \texttt{Path\_Generalist\_Classifier\_Tool} has limitations and may not always be reliable. However, the classification of "Necrotic tumor" seems plausible given the context. If further verification is needed, consulting a histopathology expert or using additional histopathology-specific tools could be considered.
\\\\
\textbf{Ambiguities:}
There are no ambiguities in the result. The classification is straightforward and matches one of the provided options.
\\\\
\textbf{Final determination:}
The memory is complete and accurate enough to generate the final output. The classification tool has provided a clear answer to the query, and no additional tool usage is necessary.
\\\\
\textbf{Conclusion:} \command{\textbf{STOP}}
\end{plannerbox}
%%%%%%%%%%%%%%%%%%%%%%%%%%%%%%%%%%%%%%%%%%%%%%%%%%%%%%%%%%%%%%%%

The final answer is summarized by the solution summarizer in the planner, given the full trajectory in the context described above. When a highly specialized tool is present for its corresponding domain, in this case histopathology classification, \model is able to leverage the specialized tool directly to reach the answer.

%%%%%%%%%%%%%%%%%%%%%%%%%%%%%%%%%%%%%%%%%%%%%%%%%%%%%%%%%%%%%%%%
\begin{answerbox}[Final Answer]
The histology image of osteosarcoma stained with hematoxylin and eosin (H\&E) has been classified using the \texttt{Path\_Generalist\_Classifier\_Tool}. Based on the analysis, the image is identified as a:

B) Necrotic tumor
\end{answerbox}

\clearpage
\subsection{Example for Scientific Reasoning}

%%%%%%%%%%%%%%%%%%%%%%%%%%%%%%%%%%%%%%%%%%%%%%%%%%%%%%%%%%%%%%%%
\begin{querybox}[The Query]
    \textbf{Question:} Which of the following describes the fallacy of appeal to indignation?\\
    A) presenting two options as the only possible ones when more options exist \\
    B) attacking an argument by exaggerating, misrepresenting, or even completely fabricating parts of it\\
    C) arguing against a position only based on personal feelings toward the position\\
    D) responding to charges of wrongdoing by saying other people do the same thing\\
    E) arguing that a claim is true because it has not been proven false\\
    F) responding to an argument with an expression of derision instead of a counter argument\\
    G) making a claim based on popularity or majority opinion instead of evidence\\
    H) attacking the character or background of someone making an argument instead of his or her evidence and reasoning\\
    I) making a claim seem true or false based on irrelevant information or associations\\   
    Choose the correct option.
    \\\\
    \textbf{Ground truth:} E
    % \vspace{20pt}
\end{querybox}

%%%%%%%%%%%%%%%%%%%%%%%%%%%%%%%%%%%%%%%%%%%%%%%%%%%%%%%%%%%%%%%%

%%%%%%%%%%%%%%%%%%%%%%%%%%%%%%%%%%%%%%%%%%%%%%%%%%%%%%%%%%%%%%%%
\begin{querybox}[Wrong Prediction from \gpt]
\red{F)}\\
The fallacy of appeal to indignation involves responding to an argument with an expression of derision or outrage instead of providing a logical counterargument. This fallacy focuses on expressing strong emotions rather than addressing the substance of the argument itself.
\end{querybox}
%%%%%%%%%%%%%%%%%%%%%%%%%%%%%%%%%%%%%%%%%%%%%%%%%%%%%%%%%%%%%%%%

In contrast, \model decomposes the query into two steps by calling the knowledge retrieval tool and reasoning tool to answer the question.  First, it calls the \texttt{Wikipedia\_Knowledge\_Searcher\_Tool} to retrieve related knowledge. Then it uses \texttt{Generalist\_Solution\_Generator\_Tool} to perform reasoning over the query and the knowledge retrieved.  

%%%%%%%%%%%%%%%%%%%%%%%%%%%%%%%%%%%%%%%%%%%%%%%%%%%%%%%%%%%%%%%%
\begin{plannerbox}[Step 0: Initial Plan from Query Analyzer]

% \textbf{Concise summary:}
% The task is to classify a histology image of osteosarcoma stained with hematoxylin and eosin (H\&E) into one of three categories: Non-tumor, Necrotic tumor, or Viable tumor.
% \\\\
\textbf{Required skills:}

1. Critical Thinking: To analyze and understand the descriptions of logical fallacies and identify the correct one.\\
2. Knowledge of Logical Fallacies: To accurately recognize and differentiate between various logical fallacies, including the appeal to indignation.\\
3. Decision Making: To choose the correct option from the given list based on the analysis.
\\\\
\textbf{Relevant tools:}

1. \texttt{Wikipedia\_Knowledge\_Searcher\_Tool}: This tool can be used to search for information on logical fallacies, including the appeal to indignation, to verify the correct description. It provides factual information from Wikipedia, which can be cross-referenced with the options provided. \\
2.\texttt{Generalist\_Solution\_Generator\_Tool}: This tool can be used to generate a step-by-step analysis of each option to determine which one describes the fallacy of appeal to indignation. However, it may provide incorrect responses, so verification is necessary.
\\\\
\textbf{Additional considerations:}
Ensure that the final answer is verified for accuracy, as the \texttt{Generalist\_Solution\_Generator\_Tool} may provide hallucinated responses. Cross-referencing with reliable sources is recommended to confirm the correct description of the fallacy.
\end{plannerbox}
%%%%%%%%%%%%%%%%%%%%%%%%%%%%%%%%%%%%%%%%%%%%%%%%%%%%%%%%%%%%%%%%

%%%%%%%%%%%%%%%%%%%%%%%%%%%%%%%%%%%%%%%%%%%%%%%%%%%%%%%%%%%%%%%%
\begin{plannerbox}[Step 1: Action from Action Predictor]

\textbf{Justification:}
The \texttt{Wikipedia\_Knowledge\_Searcher\_Tool} is the best choice for this step because it provides factual information from a reliable source, which is crucial for accurately identifying the description of the fallacy of appeal to indignation. This tool can search for and retrieve specific information about logical fallacies, ensuring that the selected option is verified against authoritative content.
% It can directly take the image and the classification options as input and provide a classification result, which aligns perfectly with the current query's requirements.
\\\\
\textbf{Sub-goal:}
Search Wikipedia for information on the fallacy of appeal to indignation to verify which option correctly describes it.
\\\\
\textbf{Tool name:} \texttt{Wikipedia\_Knowledge\_Searcher\_Tool}
\end{plannerbox}
%%%%%%%%%%%%%%%%%%%%%%%%%%%%%%%%%%%%%%%%%%%%%%%%%%%%%%%%%%%%%%%%

%%%%%%%%%%%%%%%%%%%%%%%%%%%%%%%%%%%%%%%%%%%%%%%%%%%%%%%%%%%%%%%%
\begin{executorbox}[Step 1: Command from Command Generator]

\textbf{Analysis:}
The task is to identify which option describes the fallacy of appeal to indignation. The sub-goal is to search Wikipedia for information on this fallacy to verify the correct option. The selected tool is the \texttt{Wikipedia\_Knowledge\_Searcher\_Tool}, which requires a query string as input. The query should be related to the fallacy of appeal to indignation to retrieve relevant information from Wikipedia.
% explanation="The command is constructed to search Wikipedia for the term 'appeal to indignation fallacy'. This will help in retrieving information that can be used to verify which option correctly describes the fallacy. The tool's input type is a string query, so we pass the search term directly to the \texttt{tool.execute()} function.
% \\\\
% \textbf{Explanation:}
% The command is constructed to use the The selected tool, \texttt{Path\_Generalist\_Classifier\_Tool} by providing the required inputs: the image path and the list of classification options. The image path is directly taken from the context, and the options are extracted from the query. The command follows the tool's input requirements and usage pattern, ensuring that the image is classified into one of the specified categories.
\\\\
\textbf{Command:}
\command{execution = tool.execute(query="appeal to indignation fallac")}
\end{executorbox}
%%%%%%%%%%%%%%%%%%%%%%%%%%%%%%%%%%%%%%%%%%%%%%%%%%%%%%%%%%%%%%%%

%%%%%%%%%%%%%%%%%%%%%%%%%%%%%%%%%%%%%%%%%%%%%%%%%%%%%%%%%%%%%%%%
\begin{executorbox}[Step 1: Result from Command Executor]
\textbf{Result:} The ``fallacy of appeal to indignation'' refers to an argument tactic where someone tries to persuade others by provoking anger or outrage instead of presenting logical evidence, essentially using strong emotions like indignation to sway opinion rather than reasoned facts; it is considered a form of ``appeal to emotion'' fallacy where the emotion being exploited is anger or outrage. 
\end{executorbox}
%%%%%%%%%%%%%%%%%%%%%%%%%%%%%%%%%%%%%%%%%%%%%%%%%%%%%%%%%%%%%%%%

%%%%%%%%%%%%%%%%%%%%%%%%%%%%%%%%%%%%%%%%%%%%%%%%%%%%%%%%%%%%%%%%
\begin{plannerbox}[Step 1: Verification from Context Verifier]

\textbf{Completeness:}
The memory addresses the main objective of identifying the correct description of the fallacy. The \texttt{Generalist\_Solution\_Generator\_Tool}\'s output aligns with the understanding of the fallacy as it involves emotional reactions rather than logical reasoning.
\\\\
\textbf{Inconsistencies:}
There are no direct inconsistencies in the information provided
\\\\
\textbf{Verification needs:}
The information regarding the appeal to indignation fallacy needs further verification as the current memory does not provide a definitive answer.
\\\\
\textbf{Ambiguities:}
 The ambiguity lies in the lack of direct information about the appeal to indignation fallacy. Using the \texttt{Generalist\_Solution\_Generator\_Tool} could help clarify which option best fits the description of this fallacy.
\\\\
\textbf{Final determination:}
The memory is insufficient to generate the final output as it requires reasoning over the information on the appeal to indignation fallacy.
\\\\
\textbf{Conclusion:} \command{\textbf{CONTINUE}}
\end{plannerbox}

%%%%%%%%%%%%%%%%%%%%%%%%%%%%%%%%%%%%%%%%%%%%%%%%%%%%%%%%%%%%%%%%
\begin{plannerbox}[Step 2: Action from Action Predictor]

\textbf{Justification:}
The previous attempts to use the \texttt{Wikipedia\_Knowledge\_Searcher\_Tool} did not yield a direct answer to the query. The next logical step is to use the \texttt{Generalist\_Solution\_Generator\_Tool}. This tool can analyze the provided options and use its general knowledge to identify which option best describes the fallacy of appeal to indignation.
\\\\
\textbf{Sub-goal:}
Use the \texttt{Generalist\_Solution\_Generator\_Tool} to analyze the provided options and determine which one best describes the fallacy of appeal to indignation.
\\\\
\textbf{Tool name:} \texttt{Generalist\_Solution\_Generator\_Tool}
\end{plannerbox}
%%%%%%%%%%%%%%%%%%%%%%%%%%%%%%%%%%%%%%%%%%%%%%%%%%%%%%%%%%%%%%%%

%%%%%%%%%%%%%%%%%%%%%%%%%%%%%%%%%%%%%%%%%%%%%%%%%%%%%%%%%%%%%%%%
\begin{executorbox}[Step 2: Command from Command Generator]

\textbf{Analysis:}
The task is to identify which option describes the fallacy of appeal to indignation. The \texttt{Generalist\_Solution\_Generator\_Tool} is selected to analyze the options and determine the correct one. The tool requires a prompt as input, which should include the query and options for analysis. The context provided includes the query and options, which will be used to construct the prompt for the tool.
\\\\
\textbf{Command:}
\command{execution = tool.execute(prompt="Which of the following describes the fallacy of appeal to indignation?..."}
\end{executorbox}
%%%%%%%%%%%%%%%%%%%%%%%%%%%%%%%%%%%%%%%%%%%%%%%%%%%%%%%%%%%%%%%%

%%%%%%%%%%%%%%%%%%%%%%%%%%%%%%%%%%%%%%%%%%%%%%%%%%%%%%%%%%%%%%%%
\begin{executorbox}[Step 2: Result from Command Executor]
\textbf{Result:} The fallacy of appeal to indignation is best described by option C: arguing against a position only based on personal feelings toward the position. This fallacy involves rejecting an argument or claim not because of its merits or evidence, but because of an emotional reaction or personal feelings about it.
\end{executorbox}
%%%%%%%%%%%%%%%%%%%%%%%%%%%%%%%%%%%%%%%%%%%%%%%%%%%%%%%%%%%%%%%%

%%%%%%%%%%%%%%%%%%%%%%%%%%%%%%%%%%%%%%%%%%%%%%%%%%%%%%%%%%%%%%%%
\begin{plannerbox}[Step 2: Verification from Context Verifier]

\textbf{Completeness:}
The memory does not fully address the query as the search results did not provide a direct description of the appeal to indignation fallacy. The correct option from the list remains unidentified.
\\\\
\textbf{Inconsistencies:}
There are no direct inconsistencies in the information provided.
\\\\
\textbf{Verification needs:}
The information regarding the appeal to indignation fallacy needs further verification as the current memory does not provide a definitive answer.
\\\\
\textbf{Ambiguities:}
 There are no significant ambiguities in the results that require further clarification.
\\\\
\textbf{Final determination:}
The memory is sufficient to generate the final output, as the \texttt{Generalist\_Solution\_Generator\_Tool} has provided a plausible and logical answer to the query.
\\\\
\textbf{Conclusion:} \command{\textbf{STOP}}
\end{plannerbox}

When a knowledge retrieval tool is present for retrieving relevant knowledge, \model is able to leverage the retrieved knowledge directly to reach the answer.

%%%%%%%%%%%%%%%%%%%%%%%%%%%%%%%%%%%%%%%%%%%%%%%%%%%%%%%%%%%%%%%%
\begin{answerbox}[Final Answer]
Based on the analysis, option C is the correct description of the fallacy of appeal to indignation, as it involves arguing against a position based on personal feelings.
\end{answerbox}

\clearpage
\subsection{Example for Fine-grained Visual Reasoning}

%%%%%%%%%%%%%%%%%%%%%%%%%%%%%%%%%%%%%%%%%%%%%%%%%%%%%%%%%%%%%%%%
\begin{querybox}[The Query]
\begin{minipage}{0.3\linewidth}
    \includegraphics[width=1\linewidth]{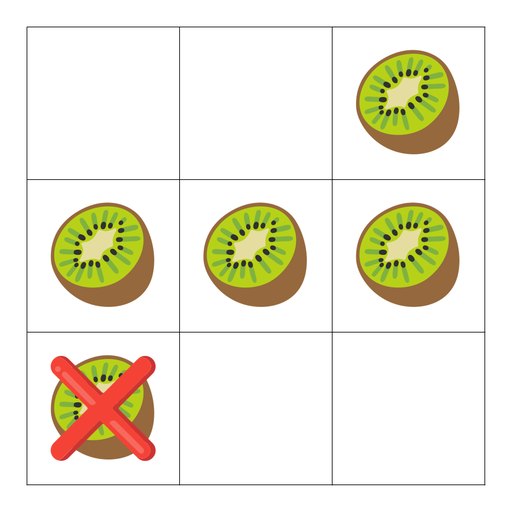}
\end{minipage}
\hspace{5pt}
\begin{minipage}{0.55\linewidth}
    \textbf{Image path:} \texttt{rotting\_kiwi\_0016.jpg}
    
    \textbf{Question:} You are given a 3 x 3 grid in which each cell can contain either no kiwi, one fresh kiwi, or one rotten kiwi. Every minute, any fresh kiwi that is 4-directionally adjacent to a rotten kiwi also becomes rotten. What is the minimum number of minutes that must elapse until no cell has a fresh kiwi?
    \\\\
    \textbf{Ground truth:} 4 minutes
    % \vspace{20pt}
\end{minipage}
\end{querybox}
%%%%%%%%%%%%%%%%%%%%%%%%%%%%%%%%%%%%%%%%%%%%%%%%%%%%%%%%%%%%%%%%

%%%%%%%%%%%%%%%%%%%%%%%%%%%%%%%%%%%%%%%%%%%%%%%%%%%%%%%%%%%%%%%%
\begin{querybox}[Wrong Prediction from \gpt]
To solve this, we need to track how the rot spreads from the rotten kiwi to the fresh ones:

\begin{enumerate}
    \item \textbf{Minute 1:} The rotten kiwi in the bottom left corner will rot the fresh kiwi directly above it (middle left cell).

    \item \textbf{Minute 2:} The newly rotten kiwi in the middle left cell will rot the fresh kiwi directly \red{above it (top left cell) and the fresh kiwi to its right (middle center cell)}.

    \item \red{\textbf{Minute 3:} The rotten kiwi in the middle center cell will rot the fresh kiwi directly above it (top center cell).}
\end{enumerate}

\red{After 3 minutes}, all fresh kiwis will be rotten.
\end{querybox}
%%%%%%%%%%%%%%%%%%%%%%%%%%%%%%%%%%%%%%%%%%%%%%%%%%%%%%%%%%%%%%%%

In contrast, \model addresses the query by first calling \texttt{Image\_Captioner\_Tool} to provide a general description of the associated image. This step is essential as the mistake from the \gpt is because of the wrong perception of the image, with the \texttt{Image\_Captioner\_Tool}, the reasoning over the rotting process is also based on the detailed description of the image, resulting in a correct answer.

%%%%%%%%%%%%%%%%%%%%%%%%%%%%%%%%%%%%%%%%%%%%%%%%%%%%%%%%%%%%%%%%
\begin{plannerbox}[Step 0: Initial Plan from Query Analyzer]
\textbf{Required skills:}

1. Image Analysis: To interpret the grid and identify the state of each kiwi (fresh or rotten).\\
2. Logical Reasoning: To simulate the process of fresh kiwis becoming rotten over time and determine the minimum time required.\\
3. Problem Solving: To apply the rules and constraints effectively to reach the correct answer.
\\\\
\textbf{Relevant tools:}

1. \texttt{Image\_Captioner\_Tool}: To generate a description of the image, identifying the state of each kiwi, which aids in understanding the initial setup.\\
2. \texttt{Generalist\_Solution\_Generator\_Tool}: To simulate the process of kiwis turning rotten and calculate the minimum time required, using logical reasoning based on the image description.
\\\\
\textbf{Additional considerations:}
Ensure the image is accurately interpreted to correctly identify the initial state of each kiwi. Logical reasoning should be applied carefully to simulate the process over time. Verify the final answer against the provided choices.
\end{plannerbox}
%%%%%%%%%%%%%%%%%%%%%%%%%%%%%%%%%%%%%%%%%%%%%%%%%%%%%%%%%%%%%%%%

%%%%%%%%%%%%%%%%%%%%%%%%%%%%%%%%%%%%%%%%%%%%%%%%%%%%%%%%%%%%%%%%
\begin{plannerbox}[Step 1: Action from Action Predictor]

\textbf{Justification:}
The \texttt{Image\_Captioner\_Tool} is the most appropriate choice for the next step because it can generate a detailed description of the image, which is crucial for understanding the initial state of the kiwis in the grid. This understanding is necessary to simulate the process of fresh kiwis becoming rotten and to determine the minimum time required. The tool's ability to describe the image will provide the foundational information needed for logical reasoning in subsequent steps.
\\\\
\textbf{Context:}
Image path: \texttt{rotting\_kiwi\_0016.jpg}
\\\\
\textbf{Sub-goal:}
Generate a detailed description of the image 'rotting\_kiwi\_0016.jpg' to identify the initial state of each kiwi in the 3x3 grid.
\\\\
\textbf{Tool name:} \texttt{Image\_Captioner\_Tool}
\end{plannerbox}
%%%%%%%%%%%%%%%%%%%%%%%%%%%%%%%%%%%%%%%%%%%%%%%%%%%%%%%%%%%%%%%%

%%%%%%%%%%%%%%%%%%%%%%%%%%%%%%%%%%%%%%%%%%%%%%%%%%%%%%%%%%%%%%%%
\begin{executorbox}[Step 1: Command from Command Generator]

\textbf{Analysis:}
The task requires generating a detailed description of an image to identify the initial state of each kiwi in a 3x3 grid. The selected tool, \texttt{Image\_Captioner\_Tool}, is designed to generate captions for images. The tool's input requires an image path and an optional prompt. The context provides the image path, which is 'rotting\_kiwi\_0016.jpg'. The sub-goal is to generate a detailed description of this image, which aligns with the tool's capability to describe images in detail. Therefore, the command will use the image path and a prompt to guide the captioning process.\\\\
\textbf{Explanation:}
The command uses the \texttt{Image\_Captioner\_Tool} to generate a detailed description of the image. The image path is provided directly from the context, and a prompt is used to ensure the tool focuses on describing the initial state of each kiwi in the grid. This setup aligns with the tool's input requirements and the task's sub-goal.
\\\\
\textbf{Command:}
\command{execution = tool.execute(image='rotting\_kiwi\_0016.jpg', prompt='Describe the initial state of each kiwi in the 3x3 grid in detail.')}
\end{executorbox}
%%%%%%%%%%%%%%%%%%%%%%%%%%%%%%%%%%%%%%%%%%%%%%%%%%%%%%%%%%%%%%%%

%%%%%%%%%%%%%%%%%%%%%%%%%%%%%%%%%%%%%%%%%%%%%%%%%%%%%%%%%%%%%%%%
\begin{executorbox}[Step 1: Result from Command Executor]
\textbf{Result:}
The 3x3 grid contains the following kiwis:

\begin{itemize}
    \item \textbf{Top Row:}
    \begin{itemize}
        \item Left: Empty
        \item Middle: Empty
        \item Right: A kiwi with a brown outer skin and a green interior with black seeds.
    \end{itemize}

    \item \textbf{Middle Row:}
    \begin{itemize}
        \item Left: A kiwi with a brown outer skin and a green interior with black seeds.
        \item Middle: A kiwi with a brown outer skin and a green interior with black seeds.
        \item Right: A kiwi with a brown outer skin and a green interior with black seeds.
    \end{itemize}

    \item \textbf{Bottom Row:}
    \begin{itemize}
        \item Left: A kiwi with a brown outer skin and a green interior with black seeds, marked with a red "X."
        \item Middle: Empty
        \item Right: Empty
    \end{itemize}
\end{itemize}

\end{executorbox}
%%%%%%%%%%%%%%%%%%%%%%%%%%%%%%%%%%%%%%%%%%%%%%%%%%%%%%%%%%%%%%%%

%%%%%%%%%%%%%%%%%%%%%%%%%%%%%%%%%%%%%%%%%%%%%%%%%%%%%%%%%%%%%%%%
\begin{plannerbox}[Step 1: Verification from Context Verifier]
The memory provides a detailed description of the initial state of each kiwi in the 3x3 grid. The top row has one fresh kiwi, the middle row has three fresh kiwis, and the bottom row has one rotten kiwi. The task is to determine how many minutes it will take for all fresh kiwis to become rotten.The memory is complete and accurate, and no additional tool usage is necessary to determine the answer.
\\\\
\textbf{Conclusion:} \command{\textbf{STOP}}
\end{plannerbox}
%%%%%%%%%%%%%%%%%%%%%%%%%%%%%%%%%%%%%%%%%%%%%%%%%%%%%%%%%%%%%%%%

%%%%%%%%%%%%%%%%%%%%%%%%%%%%%%%%%%%%%%%%%%%%%%%%%%%%%%%%%%%%%%%%

The final answer is summarized by the solution summarizer in the planner, given the full trajectory in the context described above. This two-step flow overcomes the base \gpt model’s limitation in visual perception.

%%%%%%%%%%%%%%%%%%%%%%%%%%%%%%%%%%%%%%%%%%%%%%%%%%%%%%%%%%%%%%%%
\begin{answerbox}[Final Answer]

\textbf{Detailed analysis:}
To solve the problem, we need to determine how many minutes it takes for all fresh kiwis to become rotten, given the initial state of the grid.
\\\\
\textbf{Initial Grid Analysis}
\begin{itemize}
    \item \textbf{Top Row:} 
    \begin{itemize}
        \item Left: Empty
        \item Middle: Empty
        \item Right: Fresh Kiwi
    \end{itemize}

    \item \textbf{Middle Row:}
    \begin{itemize}
        \item Left: Fresh Kiwi
        \item Middle: Fresh Kiwi
        \item Right: Fresh Kiwi
    \end{itemize}

    \item \textbf{Bottom Row:}
    \begin{itemize}
        \item Left: Rotten Kiwi
        \item Middle: Empty
        \item Right: Empty
    \end{itemize}
\end{itemize}

\textbf{Process}
\begin{enumerate}
    \item \textbf{Minute 1:}  
    The rotten kiwi at the bottom left will cause the fresh kiwi directly above it (middle row, left) to become rotten.

    \item \textbf{Minute 2:}  
    The newly rotten kiwi (middle row, left) will cause the adjacent fresh kiwis (middle row, middle) to become rotten.  
    The fresh kiwi in the top row, right, is not adjacent to any rotten kiwi yet.

    \item \textbf{Minute 3:}  
    The rotten kiwi (middle row, middle) will cause the adjacent fresh kiwi (middle row, right) to become rotten.  
    The fresh kiwi in the top row, right, is still not adjacent to any rotten kiwi.

    \item \textbf{Minute 4:}  
    The rotten kiwi (middle row, right) will cause the fresh kiwi in the top row, right, to become rotten.
\end{enumerate}

\textbf{Conclusion:}
It takes 4 minutes for all fresh kiwis to become rotten.

\end{answerbox}

\clearpage
\subsection{Exploration Example for Literature Review}
\label{app:example_exploration}

%%%%%%%%%%%%%%%%%%%%%%%%%%%%%%%%%%%%%%%%%%%%%%%%%%%%%%%%%%%%%%%%
\begin{querybox}[The Query]
\textbf{Question:}
What are the research trends in tool agents with large language models for scientific discovery? Please consider the latest literature from ArXiv, PubMed, Nature, and news sources.
\end{querybox}
%%%%%%%%%%%%%%%%%%%%%%%%%%%%%%%%%%%%%%%%%%%%%%%%%%%%%%%%%%%%%%%%

%%%%%%%%%%%%%%%%%%%%%%%%%%%%%%%%%%%%%%%%%%%%%%%%%%%%%%%%%%%%%%%%
\begin{answerbox}[Final Answer (Test date: Jan 30 2025)]
\textbf{1. Summary:}

The query sought to identify the latest research trends in tool agents with large language models (LLMs) for scientific discovery, drawing from sources like ArXiv, PubMed, Nature, and related news outlets. The analysis revealed a growing interest in integrating LLMs with various domains, emphasizing their role in enhancing scientific research, education, healthcare, and more.
\\\\
\textbf{2. Detailed Analysis:}\\

\quad \textbf{Step 1: ArXiv Paper Search}

\quad\quad - \textbf{Tool Used:} \texttt{ArXiv\_Paper\_Searcher\_Tool} 

\quad\quad - \textbf{Purpose:} To find recent academic papers on tool agents with LLMs.

\quad\quad- \textbf{Key Results:}

\quad\quad\quad  - Papers discussed the application of LLMs in multi-agent systems, UAVs, education, REST API testing, and more.
  
\quad\quad\quad - Notable papers include ``AI Agent for Education,'' ``UAVs Meet LLMs,'' and ``Reasoning Language Models: A Blueprint.''
  
\quad - \textbf{Contribution:} Provided a comprehensive view of how LLMs are being utilized across different fields, highlighting their versatility and potential for innovation. \\

\quad \textbf{Step 2: PubMed Article Search}

\quad\quad - \textbf{Tool Used:} \texttt{Pubmed\_Search\_Tool}

\quad\quad- \textbf{Purpose:} To identify recent articles on LLMs in scientific discovery.

\quad\quad- \textbf{Key Results:}

\quad\quad\quad - Articles focused on LLMs in healthcare, such as mental health support and vaccination information.

\quad\quad\quad - Highlighted the integration of LLMs in educational tools and their potential biases.

\quad\quad - \textbf{Contribution:} Offered insights into the application of LLMs in healthcare and education, emphasizing the need for accuracy and equity.\\

\quad \textbf{Step 3: Nature News Fetch}

\quad\quad - \textbf{Tool Used:} \texttt{Nature\_News\_Fetcher\_Tool}

\quad\quad - \textbf{Purpose:} To gather recent news articles on LLMs and tool agents.

\quad\quad - \textbf{Key Results:} 

\quad\quad\quad - Articles discussed the development of AI models like DeepSeek and their impact on scientific research.
  
\quad\quad\quad - Highlighted the role of AI in various scientific advancements and societal impacts.
  
\quad\quad - \textbf{Contribution:} Provided context on the broader implications of LLMs in scientific research and public perception.

\textbf{3. Key Findings:}

\quad - LLMs are increasingly being integrated into multi-agent systems, enhancing capabilities in fields like education, healthcare, and UAV operations.

\quad - There is a significant focus on improving the reasoning and problem-solving abilities of LLMs, as seen in the development of reasoning language models.
\end{answerbox}
%%%%%%%%%%%%%%%%%%%%%%%%%%%%%%%%%%%%%%%%%%%%%%%%%%%%%%%%%%%%%%%%

%%%%%%%%%%%%%%%%%%%%%%%%%%%%%%%%%%%%%%%%%%%%%%%%%%%%%%%%%%%%%%%%
\begin{answerbox}[Final Answer (Continued)]
\quad - The application of LLMs in healthcare is promising but requires careful consideration of biases and accuracy.

\quad - News articles highlight the rapid advancements in AI models and their potential to transform scientific research.
\\\\
\textbf{4. Answer to the Query:}

The latest research trends in tool agents with large language models for scientific discovery include:

\quad - The integration of LLMs into multi-agent systems to enhance collaborative problem-solving and decision-making.

\quad - The use of LLMs in healthcare for tasks like mental health support and vaccination information, with a focus on ensuring accuracy and equity.

\quad - The development of advanced reasoning models to improve the problem-solving capabilities of LLMs.

\quad - The exploration of LLMs in educational contexts to support personalized learning and cognitive development.
\\\\
\textbf{ 5. Additional Insights:}

\quad - While LLMs offer significant potential, there are challenges related to computational overhead, biases, and the need for domain-specific adaptations.

\quad - The ethical implications of LLMs, particularly in healthcare and education, require ongoing attention to ensure equitable access and outcomes.
\\\\
\textbf{6. Conclusion:}

The integration of large language models into tool agents is a rapidly evolving field with significant implications for scientific discovery. These models are enhancing capabilities across various domains, from healthcare to education, while also presenting challenges that need to be addressed. Future research should focus on optimizing these models for specific applications, ensuring ethical use, and exploring their full potential in scientific innovation.
\end{answerbox}
%%%%%%%%%%%%%%%%%%%%%%%%%%%%%%%%%%%%%%%%%%%%%%%%%%%%%%%%%%%%%%%%

\end{document}